\documentclass[10pt,journal,compsoc]{latexcls/IEEEtran}
\usepackage[T1]{fontenc}
\usepackage{enumitem}
\ifCLASSOPTIONcompsoc
  \usepackage[nocompress]{cite} 
\else
  \usepackage{cite}
\fi
\ifCLASSINFOpdf
  \usepackage[pdftex]{graphicx}
\else
  \usepackage[dvips]{graphicx}
\fi

\usepackage{pgfplots}
\pgfplotsset{compat=newest}
\usepackage{amsmath}
\ifCLASSOPTIONcompsoc
  \usepackage[caption=false,font=footnotesize,labelfont=sf,textfont=sf]{subfig}
\else
  \usepackage[caption=false,font=footnotesize]{subfig}
\fi
\usepackage{url}
\usepackage{import}
\newcommand{\etal}{\textit{et al}. }
\newcommand{\ie}{\textit{i}.\textit{e}., }
\newcommand{\eg}{\textit{e}.\textit{g}., }

\usepackage{listings}
\usepackage{example}

\begin{document}
%
% paper title
% Titles are generally capitalized except for words such as a, an, and, as,
% at, but, by, for, in, nor, of, on, or, the, to and up, which are usually
% not capitalized unless they are the first or last word of the title.
% Linebreaks \\ can be used within to get better formatting as desired.
% Do not put math or special symbols in the title.
\title{Bridging the Gap Between Computational Photography and Visual Recognition\thanks{* denotes equal contribution}}
%
%
% author names and IEEE memberships
% note positions of commas and nonbreaking spaces ( ~ ) LaTeX will not break
% a structure at a ~ so this keeps an author's name from being broken across
% two lines.
% use \thanks{} to gain access to the first footnote area
% a separate \thanks must be used for each paragraph as LaTeX2e's \thanks
% was not built to handle multiple paragraphs
%
%
%\IEEEcompsocitemizethanks is a special \thanks that produces the bulleted
% lists the Computer Society journals use for "first footnote" author
% affiliations. Use \IEEEcompsocthanksitem which works much like \item
% for each affiliation group. When not in compsoc mode,
% \IEEEcompsocitemizethanks becomes like \thanks and
% \IEEEcompsocthanksitem becomes a line break with idention. This
% facilitates dual compilation, although admittedly the differences in the
% desired content of \author between the different types of papers makes a
% one-size-fits-all approach a daunting prospect. For instance, compsoc 
% journal papers have the author affiliations above the "Manuscript
% received ..."  text while in non-compsoc journals this is reversed. Sigh. Rosaura G.~VidalMata,~\IEEEmembership{Student Member,~IEEE,}

\author{Rosaura G.~VidalMata*,
        Sreya~Banerjee*,
        Brandon~RichardWebster,
        Michael~Albright,
        Pedro~Davalos,
        Scott~McCloskey,
        Ben~Miller,
        Asong~Tambo,
        Sushobhan~Ghosh,
        Sudarshan~Nagesh,
        Ye~Yuan,
        Yueyu~Hu,
        Junru~Wu,
        Wenhan~Yang,
        Xiaoshuai~Zhang,
        Jiaying~Liu,
        Zhangyang~Wang,
        Hwann-Tzong~Chen,
        Tzu-Wei~Huang,
        Wen-Chi~Chin,
        Yi-Chun~Li,
        Mahmoud~Lababidi,
        Charles~Otto, 
        and~Walter~J. Scheirer% <-this % stops a space
\IEEEcompsocitemizethanks{

\IEEEcompsocthanksitem Rosaura~G.~VidalMata, Sreya~Banerjee, Brandon~RichardWebster, and Walter~J. Scheirer are with the University of Notre Dame %, Notre Dame, IN, 46556
% \protect\\
% note need leading \protect in front of \\ to get a newline within \thanks as
% \\ is fragile and will error, could use \hfil\break instead.
Corresponding Author's E-mail: walter.scheirer@nd.edu
\IEEEcompsocthanksitem Michael~Albright, Pedro~Davalos, Scott~McCloskey, Ben~Miller, and         Asong~Tambo are with Honeywell ACST.

\IEEEcompsocthanksitem Sushobhan~Ghosh is with  Northwestern University.

\IEEEcompsocthanksitem Sudarshan~Nagesh is with Zendar Co.

\IEEEcompsocthanksitem Ye~Yuan, Junru~Wu, and Zhangyang~Wang are with Texas A\&M University.

\IEEEcompsocthanksitem Yueyu~Hu, Xiaoshuai~Zhang, and Jiaying~Liu are with Peking University.

\IEEEcompsocthanksitem Wenhan~Yang is with the  National University of Singapore.

\IEEEcompsocthanksitem Hwann-Tzong~Chen, Tzu-Wei~Huang, Wen-Chi~Chin, and Yi-Chun~Li are with the National Tsing Hua University.

\IEEEcompsocthanksitem Mahmoud~Lababidi is with Johns Hopkins University.

\IEEEcompsocthanksitem Charles~Otto is with Noblis.

}% <-this % stops an unwanted space
%\thanks{Manuscript received April 19, 2005; revised August 26, 2015.}
}

% note the % following the last \IEEEmembership and also \thanks - 
% these prevent an unwanted space from occurring between the last author name
% and the end of the author line. \ie  if you had this:
% 
% \author{....lastname \thanks{...} \thanks{...} }
%                     ^------------^------------^----Do not want these spaces!
%
% a space would be appended to the last name and could cause every name on that
% line to be shifted left slightly. This is one of those "LaTeX things". For
% instance, "\textbf{A} \textbf{B}" will typeset as "A B" not "AB". To get
% "AB" then you have to do: "\textbf{A}\textbf{B}"
% \thanks is no different in this regard, so shield the last } of each \thanks
% that ends a line with a % and do not let a space in before the next \thanks.
% Spaces after \IEEEmembership other than the last one are OK (and needed) as
% you are supposed to have spaces between the names. For what it is worth,
% this is a minor point as most people would not even notice if the said evil
% space somehow managed to creep in.

% The paper headers
\markboth{IEEE TRANSACTIONS ON PATTERN ANALYSIS AND MACHINE INTELLIGENCE,~Vol.~X, No.~X, February~2020}%
{Shell \MakeLowercase{\textit{\etal }}: Bridging the Gap Between Computational Photography and Visual Recognition}
% The only time the second header will appear is for the odd numbered pages
% after the title page when using the twoside option.
% 
% *** Note that you probably will NOT want to include the author's ***
% *** name in the headers of peer review papers.                   ***
% You can use \ifCLASSOPTIONpeerreview for conditional compilation here if
% you desire.

% The publisher's ID mark at the bottom of the page is less important with
% Computer Society journal papers as those publications place the marks
% outside of the main text columns and, therefore, unlike regular IEEE
% journals, the available text space is not reduced by their presence.
% If you want to put a publisher's ID mark on the page you can do it like
% this:
%\IEEEpubid{0000--0000/00\$00.00~\copyright~2015 IEEE}
% or like this to get the Computer Society new two part style.
%\IEEEpubid{\makebox[\columnwidth]{\hfill 0000--0000/00/\$00.00~\copyright~2015 IEEE}%
%\hspace{\columnsep}\makebox[\columnwidth]{Published by the IEEE Computer Society\hfill}}
% Remember, if you use this you must call \IEEEpubidadjcol in the second
% column for its text to clear the IEEEpubid mark (Computer Society jorunal
% papers don't need this extra clearance.)

% use for special paper notices
%\IEEEspecialpapernotice{(Invited Paper)}

% for Computer Society papers, we must declare the abstract and index terms
% PRIOR to the title within the \IEEEtitleabstractindextext IEEEtran
% command as these need to go into the title area created by \maketitle.
% As a general rule, do not put math, special symbols or citations
% in the abstract or keywords.
\IEEEtitleabstractindextext{%
\begin{abstract}

What is the current state-of-the-art for image restoration and enhancement applied to degraded images acquired under less than ideal circumstances? Can the application of such algorithms as a pre-processing step improve image interpretability for manual analysis or automatic visual recognition to classify scene content? While there have been important advances in the area of computational photography to restore or enhance the visual quality of an image, the capabilities of such techniques have not always translated in a useful way to visual recognition tasks. Consequently, there is a pressing need for the development of algorithms that are designed for the joint problem of improving visual appearance and recognition, which will be an enabling factor for the deployment of visual recognition tools in many real-world scenarios. To address this, we introduce the UG$^2$ dataset as a large-scale benchmark composed of video imagery captured under challenging conditions, and two enhancement tasks designed to test algorithmic impact on visual quality and automatic object recognition. Furthermore, we propose a set of metrics to evaluate the joint improvement of such tasks as well as individual algorithmic advances, including a novel psychophysics-based evaluation regime for human assessment and a realistic set of quantitative measures for object recognition performance. We introduce six new algorithms for image restoration or enhancement, which were created as part of the IARPA sponsored UG$^2$ Challenge workshop held at CVPR 2018. Under the proposed evaluation regime, we present an in-depth analysis of these algorithms and a host of deep learning-based and classic baseline approaches. From the observed results, it is evident that we are in the early days of building a bridge between computational photography and visual recognition, leaving many opportunities for innovation in this area.
\end{abstract}

% Note that keywords are not normally used for peerreview papers.
}

% make the title area
\maketitle

% To allow for easy dual compilation without having to reenter the
% abstract/keywords data, the \IEEEtitleabstractindextext text will
% not be used in maketitle, but will appear (\ie  to be "transported")
% here as \IEEEdisplaynontitleabstractindextext when the compsoc 
% or transmag modes are not selected <OR> if conference mode is selected 
% - because all conference papers position the abstract like regular
% papers do.
\IEEEdisplaynontitleabstractindextext
% \IEEEdisplaynontitleabstractindextext has no effect when using
% compsoc or transmag under a non-conference mode.

% For peer review papers, you can put extra information on the cover
% page as needed:
% \ifCLASSOPTIONpeerreview
% \begin{center} \bfseries EDICS Category: 3-BBND \end{center}
% \fi
%
% For peerreview papers, this IEEEtran command inserts a page break and
% creates the second title. It will be ignored for other modes.
\IEEEpeerreviewmaketitle

\IEEEraisesectionheading{\section{Introduction}\label{sec:introduction}}

\IEEEPARstart{T}he advantages of collecting imagery from autonomous vehicle platforms such as small UAVs are clear. Man-portable systems can be launched from safe positions to penetrate difficult or dangerous terrain, acquiring hours of video without putting human lives at risk during search and rescue operations, disaster recovery, and other scenarios where some measure of danger has traditionally been a stumbling block. Similarly,  cars equipped with vision systems promise to improve road safety by more reliably reacting to hazards and other road users compared to humans. However, what remains unclear is how to automate the interpretation of what are inherently degraded images collected in such applications --- a necessary measure in the face of millions of frames from individual flights or road trips. A human-in-the-loop cannot manually sift through data of this scale for actionable information in real-time. Ideally, a computer vision system would be able to identify objects and events of interest or importance, surfacing valuable data out of a massive pool of largely uninteresting or irrelevant images, even when that data has been collected under less than ideal circumstances. To build such a system, one could turn to recent machine learning breakthroughs in visual recognition, which have been enabled by access to millions of training images from the Internet~\cite{russakovsky2015imagenet,lecun2015deep}. However, such approaches cannot be used as off-the-shelf components to assemble the system we desire, because they do not take into account artifacts unique to the operation of the sensor and optics configuration on an acquisition platform, nor are they strongly invariant to changes in weather, season, and time of day.

%\IEEEPARstart{M}{achine} learning based approaches are often %lauded for their state-of-the-art performance in a wide %range of vision problems. It is undeniable that the %introduction of learning models has been a turning point for %many vision tasks. The research and development of object %recognition techniques has experienced significant %advancements after the success of a deep learning based %method in the 2012 ILSVRC Single Object Classification %task~\cite{ILSVRC15}. Although these powerful learning %models have led to significant breakthroughs both in academy %and industry research, they are known to be vulnerable to $perturbations which can cause unforeseen classification %errors~\cite{DBLP:journals/corr/SzegedyZSBEGF13}. This %behaviour raises concerns to their suitability for %practical, real-world deployment where the environment %variables are often unconstrained and the images captured %contain a cocktail of distortions that can lead astray even %the best machine learning models. 

\begin{figure}[t]
\begin{center}
\includegraphics[width=0.37\textwidth]{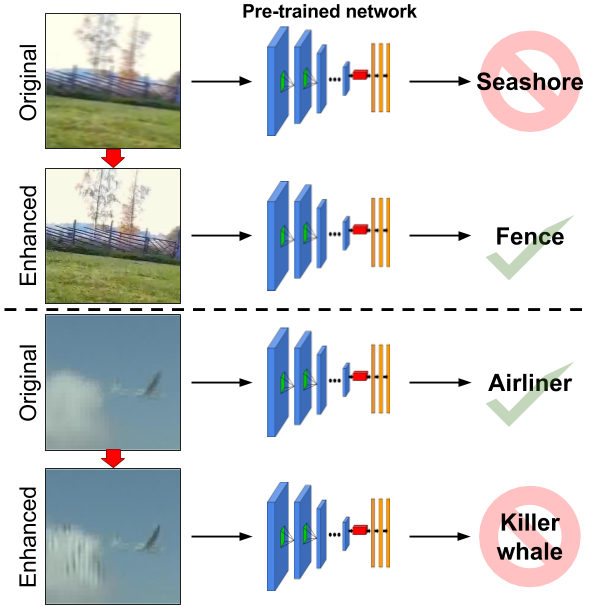}
\end{center}
   \caption{(Top) In principle, enhancement techniques like the Super-Resolution Convolutional Neural Network (SRCNN)~\cite{Chao:2014:SRCNN} should improve visual recognition performance by creating higher-quality inputs for recognition models. (Bottom) In practice, this is not always the case, especially when new artifacts are unintentionally introduced, such as in this application of Deep Video Deblurring~\cite{Su:2016:DBN}.}
   \vspace{-3mm}
\label{fig:Teaser}
\end{figure}

Whereas deep learning-based recognition algorithms can perform on par with humans on good quality images~\cite{He_2015_ICCV,Taigman_2014_CVPR}, their performance on distorted samples is degraded. It has been observed that the presence of imaging artifacts  can severely impact the recognition accuracy of state-of-the-art approaches~\cite{ug2, Dodge2016, Paranhos2016, 8038465, 8260620, DBLP:journals/corr/abs-1710-06805,RichardWebsterPsyPhy2018}. Having a real-world application such as a search and rescue drone or autonomous driving system fail in the presence of ambient perturbations such as rain, haze or even motion induced blur could have unfortunate aftereffects.  Consequently, developing and evaluating algorithms that can improve the object classification of images captured under less than ideal circumstances is fundamental for the implementation of visual recognition models that need to be reliable. And while one's first inclination would be to turn to the area of computational photography for algorithms that remove corruptions or gain resolution, one must ensure that they are compatible with the recognition process itself, and do not adversely affect the feature extraction or classification processes (Fig.~\ref{fig:Teaser}) before incorporating them into a processing pipeline that corrects and subsequently classifies images. 

The computer vision renaissance we are experiencing has yielded effective algorithms that can improve the visual appearance of an image~\cite{zeyde2010single,bevilacqua2012low,timofte2014,huang2015single,Su:2016:DBN}, but many of their enhancing capabilities do not translate well to recognition tasks as the training regime is often isolated from the visual recognition aspect of the pipeline. In fact, recent works~\cite{ug2, DBLP:journals/corr/DiamondSBWH17, DBLP:journals/corr/abs-1803-11316,DBLP:journals/corr/SajjadiSH16} have shown that approaches that obtain higher scores on classic quality estimation metrics (namely  Peak Signal to Noise Ratio), and thus, would be expected to produce high-quality images, do not necessarily perform well at improving or even maintaining the original image classification performance. Taking this into consideration, we propose to bridge the gap between traditional image enhancement approaches and visual recognition tasks as a way to jointly increase the abilities of enhancement techniques for both scenarios. 

In line with the above objective, in this work, we introduce UG$^2$: a large-scale video benchmark for assessing image restoration and enhancement for visual recognition. It consists of a publicly available dataset (\url{https://goo.gl/AjA6En}) composed of videos captured from three difficult real-world scenarios: uncontrolled videos taken by UAVs and manned gliders, as well as controlled videos taken on the ground. Over $200,000$ annotated frames for hundreds of ImageNet classes are available. From the base dataset, different enhancement tasks can be designed to evaluate improvement in visual quality and automatic object recognition, including supporting rules that can be followed to execute such evaluations in a precise and reproducible manner. This article describes the creation of the UG$^2$ dataset as well as the advances in visual enhancement and recognition that have been possible as a result. 

%This is a new challenge that does not have any %direct relationship with any past workshops held %at a major vision conference. However it is %similar in spirit to the PASCAL %VOC~\cite{pascal-voc-2012} and %ImageNet~\cite{ILSVRC15} workshops that have been %held over the years. 

Specifically, we summarize the results of the IARPA sponsored UG$^2$ Challenge workshop held at CVPR 2018. The challenge consisted of two specific tasks defined around the UG$^2$ dataset: (1) image restoration and enhancement to improve image quality for manual inspection, and (2) image restoration and enhancement to improve the automatic classification of objects found within individual images. The UG$^2$ dataset contains manually annotated video imagery (including object labels and bounding boxes) with an ample variety of imaging artifacts and optical aberrations; thus it allows for the development and quantitative evaluation of image enhancement algorithms. Participants in the challenge were able to use the provided imagery and as much out-of-dataset imagery as they liked for training and validation purposes. Enhancement algorithms were then submitted for evaluation and results were revealed at the end of the competition period.  

The competition resulted in six new algorithms, designed by different teams, for image restoration and enhancement in challenging image acquisition circumstances. These algorithms included strategies to dynamically estimate corruption and choose the appropriate response, the simultaneous targeting of multiple artifacts, the ability to leverage known image priors that match a candidate probe image, super-resolution techniques adapted from the area of remote sensing, and super-resolution via Generative Adversarial Networks. This was the largest concerted effort to-date to develop new approaches in computational photography supporting human preference and automatic recognition. We look at all of these algorithms in this article.

Having a good stable of existing and new restoration and enhancement algorithms is a nice start, but are any of them useful for the image analysis tasks at hand? Here we take a deeper look at the problem of scoring such algorithms. Specifically, the question of whether or not researchers have been doing the right thing when it comes to automated evaluation metrics for tasks like deconvolution, super-resolution and other forms of image artifact removal is explored. We introduce a visual psychophysics-inspired assessment regime, where human perception is the reference point, as an alternative to other forms of automatic and manual assessment that have been proposed in the literature. Using the methods and procedures of psychophysics that have been developed for the study of human vision in psychology, we can perform a more principled assessment of image improvement than just a simple A/B test, which is common in computer vision. We compare this human experiment with the recently introduced Learned Perceptual Image Patch Similarity (LPIPS) metric proposed by Zhang \etal \cite{DBLP:journals/corr/abs-1801-03924}. Further, when it comes to assessing the impact of restoration and enhancement algorithms on visual recognition, we suggest that the recognition performance numbers are the only metric that one should consider. As we will see from the results, much more work is needed before practical applications can be supported.

\begin{table*}[t]
\centering

\begin{tabular}{|l|l|l|l|l|}
\hline
\textbf{Dataset} & \textbf{Frames} & \textbf{Videos} & \textbf{Classes} & \textbf{Capture conditions}                                                         \\ \hline \hline

Inria-AILD~\cite{maggiori2017dataset}       & 360             & ---             & 2                & Ortho-rectified aerial imagery                                                       \\ \hline
iSAID~\cite{DBLP:journals/corr/abs-1905-12886}            & 2,806           & ---             & 15               & Satellite images (Earth Vision)                                                      \\ \hline
UAVDT~\cite{DBLP:journals/corr/abs-1804-00518}            & 80,000          & 100             & 3                & Mobile airborne videos (UAV)                                                        \\ \hline
UAV123~\cite{mueller2016benchmark}           & 112,578         & 123             &   ---               & Mobile airborne videos (UAV)                                                        \\ \hline
VisDrone~\cite{zhuvisdrone2018}         & 179,264         & 263             & 10               & Mobile airborne videos (UAV)                                                        \\ \hline
DOTA-v1.5~\cite{DBLP:journals/corr/abs-1711-10398}        & 400,000         & ---             & 16               & Satellite images (Earth Vision)                                                      \\ \hline
Campus~\cite{robicquet2016learning}           & 929,499         & ---             & 6                & Mobile airborne videos (UAV)                                                        \\ \hline \hline
UG$^2$              & \textbf{3,535,382}       & \textbf{629}             & \textbf{37}               & Mobile airborne videos (UAV, and Glider), ground-based videos \\ \hline
\end{tabular}

\vspace{2mm}
\caption{Comparison of the UG$^2$ dataset to related aerial datasets. Datasets with a missing video count contain only still images. The UAV123 dataset is designed for tracking the trajectories of cars, and as such, does not contain different object classes in its annotations.}
\vspace{-10mm}
\label{tab:datasetcomp}
\end{table*}

In summary, the contributions of this article are: 

\begin{itemize}
    \item A new video benchmark dataset representing both ideal conditions and common aerial image artifacts, which we make available to facilitate new research and to simplify the reproducibility of experimentation.
    \item A set of protocols for the study of image enhancement and restoration for image quality improvement, as well as visual recognition. This includes a novel psychophysics-based evaluation regime for human assessment and a realistic set of quantitative measures for object recognition performance.
    \item An evaluation of the influence of problematic conditions on object recognition models including VGG16 and VGG19~\cite{VGG:2014}, InceptionV3~\cite{Inception:2015},  ResNet50~\cite{ResNet50:2015}, MobileNet~\cite{howard2017mobilenets}, and NASNetMobile~\cite{Zoph_2018_CVPR}.
    \item The introduction of six new algorithms for image enhancement or restoration, which were created as part of the UG$^2$ Challenge workshop held at CVPR 2018. These algorithms are pitted against eight different classical and deep learning-based baseline algorithms from the literature on the same benchmark data. 
    \item A series of recommendations on specific aspects of the problem that the field should focus its attention on so that we have a better chance at enabling scene understanding with less than ideal acquisition circumstances. 
\end{itemize} 

%This paper may be of interest to researchers %working in the areas of computational photography %and image classification, and to anybody %interested in understanding the current state of %image processing techniques as part of a visual %recognition pipeline. The collected dataset and %additional information about the UG$^2$ Challenge %can be found at: \url{www.ug2challenge.org}.
\vspace{-3mm}
\section{Related Work} \label{Related-work}

% needed in second column of first page if using \IEEEpubid
%\IEEEpubidadjcol

% \subsubsection{Subsubsection Heading Here}
\textbf{Datasets.} The areas of image restoration and enhancement have a long history in computational photography, with associated benchmark datasets that are mainly used for the qualitative evaluation of image appearance. These include very small test image sets such as Set5~\cite{bevilacqua2012low} and Set14~\cite{zeyde2010single}, the set of blurred images introduced by Levin \etal\cite{Levin:2009}, and the DIVerse 2K resolution image
dataset (DIV2K)~\cite{Agustsson_2017_CVPR_Workshops} designed for super-resolution benchmarking. Datasets containing more diverse scene content have been proposed including Urban100~\cite{huang2015single} for enhancement comparisons and LIVE1~\cite{sheikh2006statistical} for image quality assessment. While not originally designed for computational photography, the Berkeley Segmentation Dataset has been used by itself~\cite{huang2015single} and in combination with LIVE1~\cite{yang2014single} for enhancement work. The popularity of deep learning methods has increased demand for training and testing data, which Su \etal  provide as video content for deblurring work~\cite{Su:2016:DBN}. Importantly, none of these datasets were designed to combine image restoration and enhancement with recognition for a unified benchmark.

Most similar to the dataset we employ in this paper are various large-scale video surveillance datasets, especially those which provide a ``fixed" overhead view of urban scenes~\cite{CAVIAR:Dataset:2004, grgic2011scface, CUHK:Dataset:2014, TISI:Dataset:2013}. However, these datasets are primarily meant for other research areas (\eg event/action understanding, video summarization, face recognition) and are ill-suited for object recognition tasks, even if they share some common imaging artifacts that impair recognition. 

With respect to data collected by aerial vehicles, the VIRAT Video Dataset~\cite{Virat:Dataset:2011} contains ``realistic, natural and challenging (in terms of its resolution, background clutter, diversity in scenes)'' imagery for event recognition, while the VisDrone2018 Dataset~\cite{zhuvisdrone2018} is designed for object detection and tracking. Other datasets including aerial imagery are the UCF Aerial Action Data Set~\cite{UCFAA}, UCF-ARG~\cite{UCFARG}, UAV123~\cite{mueller2016benchmark}, UAVDT~\cite{DBLP:journals/corr/abs-1804-00518}, Campus~\cite{robicquet2016learning}, and the multi-purpose dataset introduced by Yao \etal   \cite{LHI:Dataset:2007}. However, existing datasets in this area contain a limited number of frames and object categories. Table~\ref{tab:datasetcomp} provides a comparison of our UG$^2$ dataset to relevant aerial datasets. As with the computational photography datasets, none of these sets have protocols for image restoration and enhancement coupled with object recognition. 

\textbf{Visual Quality Enhancement.} There is a wide variety of enhancement methods dealing with different kinds of artifacts, such as deblurring (where the objective is to recover a sharp version $x'$ of a blurry image $y$ without knowledge of the blur parameters)~\cite{levin2011efficient, levin2007image, Levin:2009, Joshi:2009, Levin:2011, law2006lucky, matsushita2006full, cho2009fast}, denoising (where the goal is the restoration of an image $x$ from a corrupted observation $y = x + n$, where $n$ is assumed to be noise with variance $\sigma^2$)~\cite{Joshi:2009, Levin:2011, BM3D, NLM}, compression artifact reduction (which focuses on removing blocking artifacts, ringing effects or other lossy compression-induced degradation)~\cite{List2003, Reeve1984, Foi2007, FastARCNN}, reflection removal~\cite{990495, 7298939}, and super-resolution (which attempts to estimate a high-resolution image from one or more low-resolution images)~\cite{yang2008image, Yang:2010, freeman2002example, efrat2013accurate, freedman2011image, timofte2013anchored, Chao:2014:SRCNN,Kim:2016:VDSR, lai2017deep}. Other approaches designed to deal with atmospheric perturbations include dehazing (which attempts to recover the scene radiance $J$, the global atmospheric light $A$ and the medium transmission $t$ from a hazy image $I(x) = J(x)t(x) + A(1-t(x))$)~\cite{Fattal:2008, Schechner2001, 5567108, Tan2008, Tarel2009}, and rain removal techniques~\cite{4036636, Barnum2009, Jia2018RainRemoval, DBLP:journals/corr/abs-1803-10433}.  

Most of these approaches are tailored to address a particular kind of visual aberration, and the presence of multiple problematic conditions in a single image might lead to the introduction of artifacts by the chosen enhancement technique. Recent work has explored the possibility of handling multiple degradation types~\cite{ Guo2017, Zhang2017, Tai2017, Yu2018}.  

\begin{figure*}[t]
    \centering
    \includegraphics[width=0.85\textwidth]{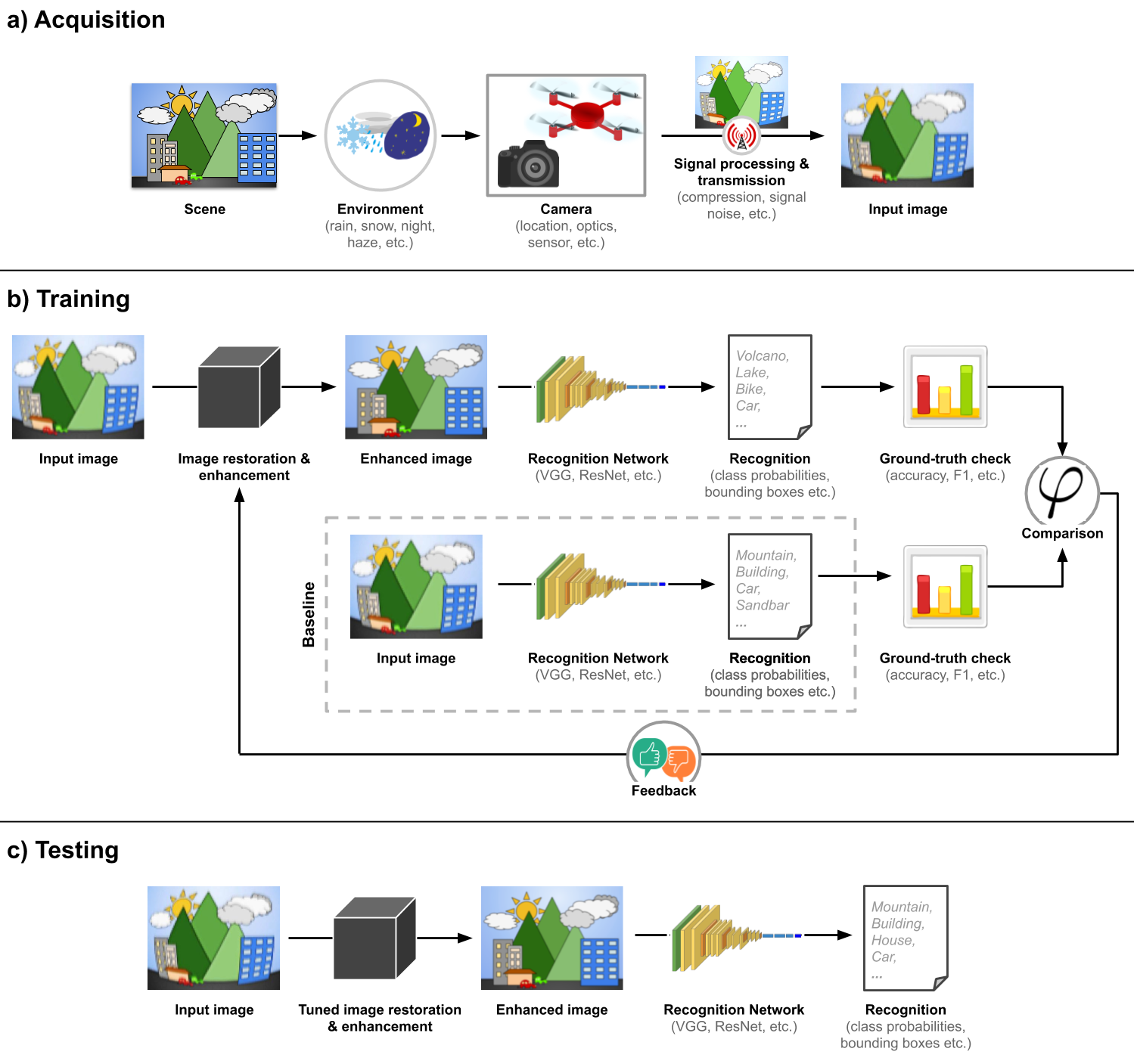}
    \vspace{-4mm}
    \caption{a) Sources of image degradation during acquisition. For a detailed discussion of how these artifacts occur, see Supp. Sec. 1. b) and c) The proposed framework for unifying image restoration and enhancement and object recognition, supporting machine learning training and testing.}
    \label{fig:framework}
\end{figure*}

\textbf{Visual Enhancement for Recognition.} Intuitively, if an image has been corrupted, then employing restoration techniques should improve the performance of recognizing objects in the image. An early attempt at unifying a high-level task like object recognition with a low-level task like deblurring was performed by Zeiler \etal through deconvolutional networks~\cite{zeiler2010deconvolutional, zeiler2011adaptive}. Similarly, Haris \etal \cite{DBLP:journals/corr/abs-1803-11316} proposed an end-to-end super-resolution training procedure that incorporated detection loss as a training objective, obtaining superior object detection results compared to traditional super-resolution methods for a variety of conditions (including additional perturbations on the low-resolution images such as the addition of Gaussian noise).

Sajjadi \etal~\cite{DBLP:journals/corr/SajjadiSH16} argue that the use of traditional metrics such as Peak Signal to Noise Ratio (PSNR), Structural Similarity Index (SSIM), or the Information Fidelity Criterion (IFC) might not reflect the performance of some models, and propose the use of object recognition performance as an evaluation metric. They observed that methods that produced images of higher perceptual quality obtained higher classification performance despite obtaining low PSNR scores. In agreement with this, Gondal \etal \cite{2018arXiv180800043W} observed the correlation of the perceptual quality of an image with its performance when processed by object recognition models. Similarly, Tahboub \etal \cite{8297071} evaluate the impact of degradation caused by video compression on pedestrian detection. Other approaches have used visual recognition as a way to evaluate the performance of visual enhancement algorithms for tasks such as text deblurring~\cite{hradivs2015convolutional, xiao2016learning}, image colorization~\cite{DBLP:journals/corr/ZhangIE16}, and single image super-resolution~\cite{6115959}. 

Also similar to our work, Li \etal\cite{Li_2019_CVPR} perform an in-depth analysis of diverse de-raining models where they compare them using a wide variety of metrics, including their impact on object detection. It is important to note that while similar in spirit to our work, the main purpose of their task-driven evaluation is to provide a complementary perspective on the performance of their visual enhancement method, rather than to improve machine learning-based object recognition.  %This is evidenced by their findings on the deterioration generated by the analyzed de-raining algorithms on the object recognition task (when compared to the performance of the original rainy images).

While the above approaches employ object recognition in addition to visual enhancement, there are approaches designed to overlook the visual appearance of the image and instead make use of enhancement techniques to exclusively improve the object recognition performance. Sharma \etal \cite{Sharma2017} make use of dynamic enhancement filters in an end-to-end processing and classification pipeline that incorporates two loss functions (enhancement and classification). The approach focuses on improving the performance of challenging high-quality images. In contrast to this, Yim \etal \cite{DBLP:journals/corr/abs-1710-06805} propose a classification architecture (comprised of a pre-processing module and a neural network model) to handle images degraded by noise. Li \etal \cite{8237773} introduced a dehazing method that is concatenated with Faster R-CNN and jointly optimized as a unified pipeline. It outperforms traditional Faster R-CNN and other non-joint approaches. Singh~\etal\cite{singh2019dual} propose the use of a dual directed capsule network with a dual high-resolution and targeted reconstruction loss to reconstruct very low-resolution images (16x16 pixels) in order to improve digit and face recognition.

Additional work has been undertaken in using visual enhancement techniques to improve high-level tasks such as face recognition~\cite{yao2008improving, nishiyama2009facial, zhang2011close, FOOKES201275, 4401949,DBLP:journals/corr/WuDXC16, lin2005face, lin2007super, hennings2008simultaneous, yu2011face, huang2011super, uiboupin2016facial, rasti2016convolutional} (through the incorporation of deblurring, super-resolution, and hallucination techniques) and person re-identification~\cite{jing2015super} algorithms for video surveillance data. 
\vspace{-3mm}

\section{A Framework for Improving Object Recognition in Degraded Imagery}

%For this year's challenge, we received 6 new %algorithms that are thoroughly described in %Section. 5. While the solutions were diverse in %terms of algorithms' usage (GANs,  %super-resolution, de-interlacing etc), atleast %two of them operated with similar underlying %theme of using the prior information about %collections the images came from for selective %enhancement/restoration. While it worked in %some case (Honeywell's CCRE), it did not %improve the result by a margin and introduced %some artifacts in the restored images (Texas %A\&M's CDRM). 
All images acquired in real environments are degraded in some way  (Fig.~\ref{fig:framework}.a). A detailed discussion of how this happens can be found in Supp. Sec.~1. 
The first step towards solving the problem at hand is to define a specific framework that algorithms will operate in. Ideally, the framework should be able to jointly optimize the dual tasks of image restoration and enhancement, and object recognition. The input to the system is a naturally corrupted image and the output is the corresponding improved image, optimized for classification performance. During training, a learning objective $\varphi$ can be defined that compares the reconstructed output from the system to that derived from baseline input. The result is then used to make further adjustments to the model, if needed, in order to improve   performance over the baseline. This process is shown in Fig.~\ref{fig:framework}.b. 
%The sources of corruption for the input image can be environment (weather, blooming, glare, occlusion, time of the day, scale), sensor (fish-eye, over/under exposure, noise), imaging algorithm within sensor (compression artifacts for saving digital image).} 

For example, if $m_{e}$ represents the evaluation metric for the enhanced image, and $m_{b}$ represents the baseline metric for the corresponding unaltered image, then a possible learning objective for this setup can be defined as:
\begin{equation}
    \varphi = m_{e} - m_{b} \geq  \varepsilon,  \varepsilon \in [-1,1] 
\label{eq:objective}
\end{equation}
%or, 
%\begin{equation}
%    1 - (m_{e} - m_{b}) \geq \varepsilon, \varepsilon \in (0,2) 
%\end{equation}
where $\varepsilon$ is a threshold applied over the difference between scores obtained for the enhanced and unaltered noisy images. 
%The metric in Eq. 2 can be either M1 and M2, we use throughout challenge 1.2. 
The metric $m$ in Eq.~\ref{eq:objective} can be the class probabilities obtained from the recognition model for the enhanced image and its corresponding unaltered image. Even when operating with probability scores, $\varepsilon$ lies between $[-1, 1]$ to account for situations when the enhanced image is worse than the unaltered image, leading to worse object recognition performance (\textit{e.g.}, accuracy is $0$ for the enhanced image and $1$ for the baseline). 

The learning objective can be used alone or in conjunction with other loss functions (\textit{e.g.}, reconstruction loss for restoration). It can be optimized using gradient descent. If ground-truth is available for both image artifacts and objects present in images, then the learning setting is fully supervised. If only ground-truth object annotations are available, then the setting is semi-supervised (this is what is considered for the UG$^2$ dataset). All of the novel algorithms we discuss in this paper operate within this framework.

\section{A New Evaluation Regime for Image Restoration and Enhancement} \label{Tasks}

To assess the interplay between restoration and enhancement and visual recognition, we designed two evaluation tasks: (1) \textit{enhancement to facilitate manual inspection}, where algorithms produce enhanced images to facilitate human assessment, and (2) \textit{enhancement to improve object recognition}, where algorithms produce enhanced images to improve object classification by state-of-the-art neural networks.
% \end{enumerate}

% This text feels like it's in the wrong place. This level of detail should probably go in Sec. 5
%The sequestered dataset of test images is %presented with no annotation or information %regarding their distortions. Algorithms have %to implement a well suited enhancement or %restoration method to improve the visual %recognition of the objects present in the %images. Images from this dataset are extracted %from video imagery captured by small UAVs, %pilot gliders, and ground video with induced %aberrations and is an extension to the %training data provided in~\cite{ug2}. Given %that the image quality of each of the %collections is quite diverse and for some of %them (particularly the UAV collection) it has %undergone multiple types of compression, we %allowed the participants to choose between %three image formats (PNG, JPG, and TIF) as %input for their proposed methods. This %distinction was important for algorithms %specialized in the removal of specific %compression artifacts. Section \ref{Dataset} %provides more details about the training and %testing datasets.

\vspace{-2mm}
\subsection{Enhancement to Facilitate Manual Inspection} \label{Tasks:1}

The first task is an evaluation of the qualitative enhancement of images. Through this task, we wish to answer two questions: Did the algorithm produce an enhancement that agrees with human perceptual judgment? And to what extent was the enhancement an improvement or deterioration? Widely used metrics such as SSIM have attempted to answer these questions by estimating human perception but have often failed to accurately imitate the nuances of human visual perception, and at times have caused algorithms to deteriorate perceptual quality~\cite{DBLP:journals/corr/abs-1801-03924}. 

While prior work has successfully adapted psychophysical methods from psychology as a means to directly study perceptual quality, these methods have been primarily posed as real-vs-fake tests~\cite{blau2018perception}. With respect to qualitative enhancement, these real-vs-fake methods can only indicate if an enhancement has caused enough alteration to cause humans to cross the threshold of perception, and provides little help in answering the two questions we are interested in for this task. Zhang \etal \cite{DBLP:journals/corr/abs-1801-03924} came close to answering these questions when they proposed LPIPS for evaluating perceptual similarity. However, this metric lacks the ability to measure whether the enhancement was an improvement or a deterioration (see the analysis in Sec.~\ref{ResAnalysis}).

\begin{figure}[t]
    \centering
    \includegraphics[width=0.25\textwidth]{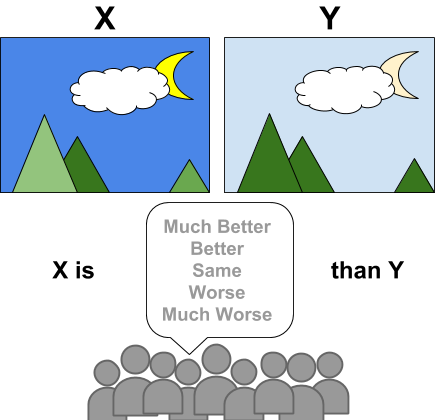}
    \caption{The visual enhancement task deployed on Amazon Mechanical Turk. An observer is presented with the original image $x$ and the enhanced image $y$. The observer is then asked to select which label they perceive is most applicable. The selected label is converted to an integer value $[1,5]$. The final rating for the enhanced image is the mean score from approximately $20$ observers. See Sect.~\ref{Tasks:1} for further details.}
    \vspace{-3mm}
    \label{fig:c1_task}
\end{figure}

In light of this, for our task we propose a new procedure, grounded in psychophysics, to evaluate the visual enhancement of images by answering both of our posed questions. The procedure for assessing image quality enhancement is a \textit{non-forced-choice} procedure that allows us to take measurements of both the threshold of perceived change and the suprathresholds, which represent the degree of perceived change~\cite{prins2016psychophysics}. Specifically, we employ a bipolar labeled Likert Scale to estimate the amount of improvement or deterioration the observer perceives, once the threshold has been crossed. The complete procedure is as follows. 

An observer is presented with an image $x$ positioned on the left-hand side of a screen and the output of the enhancement algorithm $y$ on the right. The observer is informed that $x$ is the original image and $y$ is the enhanced image. Below the image pair, five labels are provided and the observer is asked to select the label that most applies (see Fig.~\ref{fig:c1_task} for labels and layout). To capture as much of the underlying complexities in human judgment as possible, no criteria is provided for making a selection. To reduce any dependence on the subsystems in the visual cortex that specialize in memory, the pair of images is displayed until the observer selects their perceived label. An observer is given unlimited time and is informed that providing accurate labels is most important. For images larger than $480 \times 480$ pixels, the observer has the option to enlarge the images and examine them in finer detail.

The label that is selected by the observer is then converted to an assigned ordinal value $1-5$ where $[1,3)$ and $(3,5]$ are the suprathreshold measurements of improvement and deterioration, respectively. A rating of $3$ is the superficial threshold imposed on the observer, which indicates that the enhancement was imperceptible. In our proposed procedure, there is no notion of accuracy in the measurement of qualitative enhancement, as it is left entirely up to the observer's subjective discretion. However, when there are $\geq 2$ sampled observers, the perception of quality to the average observer can be estimated to provide a reliable metric for the evaluation of qualitative enhancement. We verified this holds true even when $x$ and $y$ are swapped (\textit{i.e.}, responses are symmetric).

% for measurements suprathresholdsemploys a bipolar Likert Scale to allow a rater to select their perceived perceptual enhancement. Unlike most forced-choice procedures, nonforced-choice procedures allow for measurements of suprathresholds

% Taking this into account, we propose a psychophysics-based evaluation procedure to analyze the perceived improvement of enhancement algorithms. For each image $x$, the submitted algorithms provided an enhanced version $y$ which was then evaluated by human raters in order to determine perceived improvement over $x$ in a likert-scale based task. For this task we employed a subset of the testing dataset consisting of 100 images per each collection (300 images in total). Each image pair was evaluated by approximately 20 workers on a likert-scale of $1$ to $5$ (as shown in Fig~\ref{fig:c1_task}, where 1 represented a strong improvement (the enhanced image was considered much better than the original), 3 represented no noticeable difference between the two, and 5 represented a strong deterioration (the enhanced image was considered much worse than the original), the enhanced image score was then calculated based on the average of the scores it received. Enhanced images with scores less than 3 were then considered an improvement over the original method, and scores equal or higher than 3 represented a lack of improvement. A participant's rank was then determined by the number of images their algorithm improved based on the visual perception of human raters.

To perform a large scale evaluation, we used Amazon's Mechanical Turk (AMT) service, which is widely deployed for many related tasks in computer vision \cite{DBLP:journals/corr/abs-1801-03924, Crump2013EvaluatingAM, russakovsky2015imagenet}. AMT allows a Requester (\ie researcher) to hire Workers to perform a task for payment. Our task was for a Worker to participate in the rating procedure for 100 image pairs. An additional three sentinel images pairs were given to ensure that the Worker was actively participating in the rating procedure. The ratings of Workers who failed to correctly respond to at least two of the three sentinel image pairs were discarded. In total, we had over $600$ Workers rating each image enhancement approximately $20$ times. Out of that pool, $\sim85.4\%$ successfully classified a majority of the sentinel image pairs (the ratings provided by the remaining Workers were discarded). See Sec.~\ref{ResAnalysis} for results and analysis.

% We hired over 600 workers to 

% While users were instructed to make an accurate evaluation of the two images, we needed to set up a quality control system to ensure we had mindful evaluations of each image pair. It is important to consider that human users make mistakes, and some of them might not follow the instructions or take time to appropriately evaluate the image pairs, moreover users do not always agree on their evaluations, specially when the perceived changes are subtle. To address these issues we had multiple users independently evaluate the image pairs for each algorithm, the score of each image being the user's mean rating. Furthermore to identify users who provided "low quality assessments" (product of not following the task's instructions or un-attentiveness while performing the task) we introduce sentinel image pairs with obvious improvements or degradations (e.g. an image with a large amount of Gaussian blur vs the original sharp image) in the evaluation tasks, the answers of users who fail to appropriately evaluate a majority of such sentinel image pairs were then discarded. The AMT workers were tasked to provide a judgment on the perceived improvement of enhanced images over the original frames, each worker evaluated 103 image pairs (100 valid image pairs and 3 sentinel pairs to check the validity of their answers) consisting of original un-altered frames and their enhanced counterparts. 
\vspace{-1mm}
\subsection{Enhancement to Improve Object  Recognition} \label{Tasks:2}

% \begin{figure}[t]
%     \centering
%     \includegraphics[width=0.45\textwidth]{images/Challenge2.png}
%     \caption{Work flow for classification improvement: after enhancing the input frame with the candidate algorithm, the annotated objects in the frame are extracted and sent as input to each of the classification networks, which classify the image. Each algorithm's score is then calculated as specified in Sec.~\ref{c2:EvalCrit}.}
%     \vspace{-3mm}
%     \label{fig:c2_workflow}
% \end{figure}

The second task is an evaluation of the performance improvement given by enhanced images when used as input to state-of-the-art image classification networks. When considering a fixed dataset for training (as in Fig.~\ref{fig:framework}.b), the evaluation protocol allows for the use of some within dataset training data (the training data provided by the UG$^2$ dataset, described below, contains frame-level annotations of the object classes of interest), and as much out of dataset data as needed for training and validation purposes. In order to establish good baselines for classification performance before and after the application of image enhancement and restoration algorithms, this task makes use of a selection of deep learning approaches to recognize annotated objects and then scores results based on the classification accuracy. The Keras~\cite{chollet2015keras} versions of the pre-trained networks VGG16 and VGG19~\cite{VGG:2014}, InceptionV3~\cite{Inception:2015}, and ResNet50~\cite{ResNet50:2015} are used for this purpose. We also look at two lightweight mobile networks: MobileNet~\cite{howard2017mobilenets} and NASNetMobile~\cite{Zoph_2018_CVPR}.

Each candidate algorithm is treated as an image pre-processing step to prepare sequestered test images to be submitted to all six networks. After pre-processing, the objects of interest are cropped out of the images based on verified ground-truth coordinates. The cropped images are then used as input to the networks. Algorithms are evaluated based on any improvement observed over the baseline classification result (\textit{i.e.}, the classification scores of the un-altered test images). The work flow of this evaluation pipeline is shown in Fig.~\ref{fig:framework}.c. To avoid introducing further artifacts due to down-sampling, algorithms are required to have consistent input and output frame sizes.

\subsubsection{Classification Metrics} \label{ClassifMetrics}
The networks used for the classification task return a list of the ImageNet synsets (ImageNet provides images for ``synsets" or ``synonym sets" of words or phrases that describe a concept in WordNet~\cite{fellbaum1998wordnet}) and the probability of the object belonging to each of the synset classes. However, (as will be discussed in Sec.~\ref{Dataset}), in many cases it is impossible to provide fine-grained labeling for the annotated objects. Consequently, most of the super-classes in our dataset are composed of more than one ImageNet synset. That is, each annotated image \(i\) has a single super-class label \(L_i\) which is defined by a set of ImageNet synsets \(L_i = \{s_1, ..., s_n\}\). 

To measure accuracy, we observe the number of correctly identified synsets in the top-five predictions made by each pre-trained network. A prediction is considered to be correct if its synset belongs to the set of synsets in the ground-truth super-class label. We use two metrics for this. The first measures the rate of achieving at least one correctly classified synset class (M1). In other words, for a super-class label \(L_i = \{s_1, ..., s_n\}\), a network is able to place at least one correctly classified synset in the top-five predictions. The second measures the rate of placing all the correct synset classes in the super-class label synset set (M2). For example, for a super-class label \(L_i = \{s_1, s_2, s_3\}\), a network is able to place three correct synsets in the top-five predictions.

\subsubsection{Scoring} \label{c2:EvalCrit}
Each image enhancement or restoration algorithm's performance on the classification task is then calculated by applying one of the two metrics defined above for each of the four networks and each collection within the UG$^2$ dataset. This results in up to $18$ scores for each metric (\textit{i.e.}, M1 or M2 scores from VGG16, VGG19, Inception, ResNet, MobileNet, and NASNetMobile for the UAV, Glider, and Ground collections).  For the image enhancement and restoration algorithms we consider in this article, each is ranked against all other algorithms based on these scores. A score for an enhancement algorithm is considered ``valid" if it was higher than that of the scores obtained by evaluating the classification performance of the un-altered images. In other words, we only consider a score valid if it improves upon the baseline classification task of classifying the original images. The number of valid scores in which an algorithm excels over the others being evaluated is then counted as its score in points for the task (for a maximum of $36$ points, achievable if an algorithm obtained the highest improvement --- compared to all other competitors --- in all possible configurations). 
\vspace{-1mm}
\section{The UG$^2$ Dataset} \label{Dataset}
As a basis for the evaluation regime described above in Sec.~\ref{Tasks}, we collected a new dataset called UG$^2$ (UAV, Glider, and Ground). The UG$^2$ dataset contains videos from challenging imaging scenarios containing mobile airborne cameras and ground-based captures. While we provide frame-level annotations for the purpose of object classification, the annotations can be re-purposed for other high-level tasks such as object detection and tracking~\cite{DBLP:journals/corr/abs-1907-11529}. The video files are provided to encourage further annotation for other vision tasks. The training and test datasets employed in the evaluation are composed of annotated frames from three different video collections. The annotations provide bounding boxes establishing object regions and classes, which were manually annotated using the VATIC tool for video annotation \cite{Vatic:2013}. For running classification experiments the objects were cropped from the frames in a square region of at least $224\times224$ pixels (a common input size for many deep learning-based recognition models), using the annotations as a guide. For details on annotation, see Supp. Sec. 2.

Each annotation in the dataset indicates the position, scale, visibility, and super-class of an object in a video. The need for high-level classes (super-classes) arises from the challenge of performing fine-grained object recognition using aerial collections, which have high variability in both object scale and rotation. These two factors make it difficult to differentiate some of the more fine-grained ImageNet categories. For example, while it may be easy to recognize a car from an aerial picture taken from hundreds (if not thousands) of feet above the ground, it might be impossible to determine whether that car is a taxi, a jeep or a sports car. Thus we defined super-classes that encompass multiple visually similar ImageNet synsets, as well as evaluation metrics that allow for a coarse-grained classification evaluation of such cases (see Sec.~\ref{ClassifMetrics}). The three different video collections consist of:

% \begin{figure}[!t]
% \centering
% \subfloat[UAV Collection]{
%     \includegraphics[width=0.18\textwidth]{images/YT_AV101_6852.png}
%     \includegraphics[width=0.18\textwidth]{images/YT_AV56_1104.png}
% } \vspace{-0.5em}
% \hfil
% \subfloat[Glider Collection]
% {
%     \includegraphics[width=0.18\textwidth]{images/K_KV35_9871.png}
%     \includegraphics[width=0.18\textwidth]{images/K_KV61_19797.png}
% } \vspace{-0.5em}
% \hfil
% \subfloat[Ground Collection]
% {
%     \includegraphics[width=0.18\textwidth]{images/G_car30140sungopro_398.png}
%     \includegraphics[width=0.18\textwidth]{images/G_lake50sunson_597.png}
% }
% \caption{Examples of images in the three UG$^2$ collections.}
% \vspace{-4mm}
% \label{fig:datasets-samples}
% \end{figure}

(1) \textbf{UAV Video Collection:} Composed of clips recorded from small UAVs in both rural and urban areas, the videos in this collection are open-source content tagged with a Creative Commons license, obtained from  YouTube (Supp. Fig. 3.a). Because of the source, they have different video resolutions (from $600\times400$ to $3840\times2026$), objects of interest sizes (cropped objects with sizes ranging from $224\times224$ to $800\times800$), and frame rates (from $12$ to $59$ FPS). This collection has distortions such as glare/lens flare, compression artifacts, occlusion, over/under exposure, camera shaking, sensor noise, motion blur, and fish-eye lens distortion. Videos with problematic scene/weather conditions such as night/low light video, fog, cloudy conditions and occlusion due to snowfall are also included. 

(2) \textbf{Glider Video Collection:} Consists of videos recorded by licensed pilots of fixed-wing gliders in both rural and urban areas (Supp. Fig. 3.b). The videos have frame rates ranging from $25$ to $50$ FPS, objects of interest sizes ranging from $224\times224$ to $900\times900$, and different types of compression such as MTS, MP4, and MOV. The videos mostly present imagery taken from thousands of feet above the ground, further increasing the difficulty of object recognition. Additionally, the scenes contain artifacts such as motion blur, camera shaking, noise, occlusion (which in some cases is pervasive throughout the videos, showcasing parts of the glider that partially occlude the objects of interest), glare/lens flare, over/under exposure, interlacing, and fish-eye lens distortion. Videos with problematic weather conditions such as fog, clouds, and rain are also present.

(3) \textbf{Ground Video Collection:} In order to provide some ground-truth with respect to problematic image conditions, this collection contains videos captured at ground level with intentionally induced artifacts (Supp. Fig. 3.c). These videos capture static objects  (\eg flower pots, buildings) at a wide range of distances ($30$, $40$, $50$, $60$, $70$, $100$, $150$, and $200$ft), and motion blur induced by an orbital shaker to generate horizontal movement at different rotations per minute ($120$rpm, $140$rpm, $160$rpm, and $180$rpm). Additionally, this collection includes videos under different weather conditions (sun, clouds, rain, snow) that can affect object recognition. We used a Sony Bloggie hand-held camera (with  $1280\times720$ resolution and a frame rate of 60 FPS) and a GoPro Hero 4 (with $1920\times1080$ resolution and a frame rate of 30 FPS), whose fish-eye lens introduced further distortion. Furthermore, we provide an additional class of videos (resolution-chart) showcasing a $9\times11$ inch $9\times9$ checkerboard grid exhibiting all the aforementioned distances at all intervals of rotation. The motivation for including this additional class is to provide a reference for camera calibration and to aid participants in finding the distortion measures of the cameras used. 

%\subsection{Training Dataset} %\label{Dataset:Training}

\begin{table}
\centering
\begin{center}
\begin{tabular}{|c|c|c|c|c|}
\hline
\textbf{Collection} & \textbf{UAV} & \textbf{Glider}  & \textbf{Ground}   & \textbf{Total} \\
\hline\hline

Total Videos & 231 & 120 & 278 & 629\\
Total Frames & 1,501,675 & 1,840,160 & 193,547 & 3,535,382\\ 
Annotated Videos & 30 & 30 & 136 & 196\\
Annotated Frames & 28,263 & 25,246 & 98,574 & 152,083\\ 
Extracted Objects & 29,826 & 31,760 & 98,574 & 160,160\\
Super-Classes~\cite{russakovsky2015imagenet} & 31 & 20 & 20 & 37\\

\hline
\end{tabular}
\end{center}

\caption{Summary of the UG$^2$ training dataset}
\label{tab:training_dataset_summary}
\vspace{-6mm}
\end{table}

\textbf{Training Dataset.} The training dataset is composed of $629$ videos with $3,535,382$ frames, representing $228$ ImageNet~\cite{russakovsky2015imagenet} classes extracted from annotated frames from the three different video collections. These classes are further categorized into $37$ super-classes encompassing visually similar ImageNet categories and two additional classes for pedestrian and resolution chart images. The dataset contains a subset of $152,083$ object-level annotated frames where $160,160$ objects are fully visible (out of $200,694$ total annotated frames) and the videos are tagged to indicate problematic conditions. Table~\ref{tab:training_dataset_summary} summarizes this dataset.
% and Fig.~\ref{fig:sharedClassDistrib} shows the class distribution of common classes between the three collections. Over 70\% of the UG$^2$ classes have more than 400 images and 58\% of the classes are present in the imagery of at least two collections. Around 20\% of the classes are present in all three collections.  

% \begin{figure}[t]
%     \centering
%     \includegraphics[width=0.45\textwidth]{figures/ClassDistrib_PieChart2.png}
%     \caption{Distribution of annotated images belonging to classes shared by at least two different UG$^2$ collections.}
%     \label{fig:sharedClassDistrib}
% \end{figure}

%\subsection{Testing Dataset} %\label{Dataset:Testing}

\textbf{Testing Dataset.} The testing dataset is composed of $55$ videos with frame-level annotations. Out of the annotated frames, $8,910$ disjoint frames were selected among the three different video collections, from which we extracted $9,187$ objects. These objects are further categorized into $42$ super-classes encompassing visually similar ImageNet categories. While most of the super-classes in the testing dataset overlap with those in the training dataset, there are some classes unique to each.  Table~\ref{tab:testing_dataset_summary} summarizes this dataset.

\begin{table}
    \centering
    \begin{center}
\begin{tabular}{|c|c|c|c|c|}
\hline
\textbf{Collection} & \textbf{UAV} & \textbf{Glider}  & \textbf{Ground} & \textbf{Total}  \\
\hline\hline

Total Videos & 19 & 15 & 21 & 55 \\
Annotated Frames & 2,814 & 2,909 & 3,187 & 8,910 \\ 
Extracted Objects & 3,000 & 3,000 & 3,187 & 9,187\\
Super-Classes~\cite{russakovsky2015imagenet} & 28 & 17 & 20 & 42\\

\hline
\end{tabular}
\end{center}
    \caption{Summary of the UG$^2$ testing dataset}
    \label{tab:testing_dataset_summary}
    \vspace{-6mm}
\end{table}

\section{Novel and Baseline Algorithms} \label{Methods}

% \begin{table*}
% \centering
% \subimport{./figures/}{ResultsTab.tex}
% \caption{Teams participating in UG$^2$ ordered alphabetically. Each method is identified with a codename used in the text. C1 and C2 list their overall score for the first and second challenges.}
% \label{tab:participants}
% \vspace{-6mm}
% \end{table*}

Six competitive teams participated in the 2018 UG$^2$ Workshop held at CVPR, each submitting a novel approach for image restoration and enhancement meant to address the evaluation tasks we described in Sec.~\ref{Tasks}. In addition, we assessed eight different classical and deep learning-based baseline algorithms from the literature. 
%accounting for 24 unique submissions for both %challenge tasks. 
%%The UG$^2$ Prize Challenge has allowed for %algorithmic advances in pre-processing techniques to %improve visual recognition, 
%This section highlights their innovative methods. We %provide an in-depth analysis of the scores for each %team in Section~\ref{ResAnalysis} %%discuss in detail %their scores for each challenge. 
% Table \ref{tab:participants} lists all the %participating teams and their overall scores for each %challenge.

\subsection{Challenge Workshop Entries} \label{Methods:Entries}

The six participating teams were Honeywell ACST, Northwestern University, Texas A\&M and Peking University, National Tsing Hua University, Johns Hopkins University, and Noblis. Each team had a unique take on the problem, with an approach designed for one or both of the tasks. 

\subsubsection{Camera and Conditions-Relevant Enhancements (CCRE)}

Honeywell ACST's algorithmic pipeline was motivated by a desire to closely target image enhancements based on image quality assessment. Of the wide range of image enhancement techniques, there is a smaller subset of enhancements which {\em may} be useful for a particular image. To find this subset, the CCRE pipeline considers the intersection of camera-relevant enhancements with conditions-relevant enhancements. Examples of camera-relevant enhancements include de-interlacing, rolling shutter removal (both depending on the sensor hardware), and de-vignetting (for fisheye lenses). Example conditions-relevant enhancements include de-hazing (when imaging distant objects outdoors) and raindrop removal. To choose among the enhancements relevant to various environmental conditions and the camera hardware, CCRE makes use of defect-specific detectors (Supp. Fig. 4) and takes $\sim12$ seconds to process each image.

This approach, however, requires a measure of manual tuning. For the evaluation task targeting human vision-based image quality assessment, manual inspection revealed severe interlacing in the glider set. Thus a simple interlacing detector was designed to separate each frame into two fields (comprised of the even and odd image rows, respectively) and compute the horizontal shift needed to register the two. If that horizontal shift was greater than $0.16$ pixels, then the image was deemed interlaced, and de-interlacing was performed by linearly interpolating the rows to restore the full resolution of one of the fields.

%% Should this be object detection?
For the evaluation task targeting automated object classification, de-interlacing is also performed with the expectation that the edge-type features learned by the VGG network will be impacted by jagged edges from interlacing artifacts. Beyond this, a camera and conditions assessment is partially automated using a file analysis heuristic to determine which of the collections a given video frame came from. While interlacing was the largest problem with the glider images, the ground and UAV collections were degraded by compression artifacts. Video frames from those collections were processed with the Fast Artifact Reduction CNN~\cite{FastARCNN}. 
%\textcolor{red}{Intuitively, their camera and %conditions assessment module can be considered %as an algorithm to estimate image quality since %it is capable of distinguishing known defects %from different collections (interlacing %artifacts from gliders, compression artifacts %from Ground, UAV) and thereby helping in %applying different enhancement to different %collections.}

% \begin{figure}[t]
%     \centering
%     \includegraphics[width=0.29\textwidth]{figures/HoneywellSystemFig.png}
%     \caption{The CCRE approach conditionally selects enhancements.}
%     \label{fig:HONsystem}
%     \vspace{-3mm}
% \end{figure}

\subsubsection{Multiple Artifact Removal CNN (MA-CNN)}
%Northwestern
The Northwestern team focused their attention on three major causes of artifacts in an image: (1) motion blur, (2) de-focus blur and (3) compression algorithms. They observed that in general, traditional algorithms address inverse problems in imaging via a two-step procedure: first by applying proximal algorithms to enforce measurement constraints and then by applying natural image priors (sometimes denoisers) on the resulting output~\cite{Heide:2014:FFC:2661229.2661260, pellizzari2017optically}. This process takes about 1 second per image. Recent trends in inverse imaging algorithms have focused on developing a single algorithm or network to address multiple imaging artifacts~\cite{chang2017one}.
% They observed that current enhancement techniques focus on developing one network that imposes measurement constraints on an image and another to impose natural image priors. 
These networks are alternately applied to denoise and deblur the image. Building on the  principle of using image quality as prior knowledge, the MA-CNN learning-based approach was developed to remove multiple artifacts in an image. A training dataset was created by introducing motion, de-focus and compression artifacts into $126,000$ images from ImageNet.
The motion-blur was introduced by using a kernel with a fixed length $l$ and random direction ${\theta}$ for each of the images in the training dataset. The defocus blur was introduced by using a Gaussian kernel with a fixed standard variance ${\sigma}$. The parameters \{$l,{\sigma}$\} were tuned to create a perceptually improved result.

MA-CNN is a fully convolutional network architecture with residual skip connections to generate the enhanced image (Supp. Fig. 5). To achieve better visual quality, a perceptual loss function that uses the first four convolutional layers of a pre-trained VGG-16 network is incorporated. %%The main idea of perceptual loss is that instead of penalizing for error per pixel loss, penalizing for visually relevant features leads to better visual quality.

% \begin{figure}[t]
%     \centering
%     \includegraphics[width=0.3\textwidth]{figures/Northwestern_net.png}
%     \caption{Network architecture used for MA-CNN image enhancement.}
%     \label{fig:NorthwesternNet}
%     \vspace{-6mm}
% \end{figure}

By default, the output of the MA-CNN contains checkerboard artifacts. Since these checkerboard artifacts are periodic, they can be removed by suppressing the corresponding frequencies in the Fourier-domain. Moreover, all images (of the same size) generated with the network have artifacts in a similar region in the Fourier domain. For images of differing sizes, the distance of the center of the artifact from the origin is proportional to its size. 
% Moreover, all the images generated with the network have the same spike in the frequency domain, thus, the centers of the these regions are identified and suppressed by a factor of $10$. This results in the removal of the artifacts. 

\subsubsection{Cascaded Degradation Removal Modules (CDRM)}
%TexasA\&M:
%%% Cutting for space --- WJS
%While most computational photography methods aim to %improve certain aspects of the image, in most cases %they only focus on removing a specific set of problems %from an otherwise good quality image. However the %assumption that images are afflicted by only one type %of aberrations does not necessarily hold for most %natural images. Such is the case for the UG$^2$ %dataset, where most images to suffer a mixture of low %resolution, noise, blur, compression artifacts, and %under-exposure.

The Texas A\&M team observed that independently removing any single type of degradation could, in fact, undermine performance in the object recognition evaluation task since other degradations were not simultaneously considered and those artifacts might be amplified during this process. Consequently, they proposed a pipeline that consists of sequentially cascaded degradation removal modules to improve recognition. Further, they observed that different collections within the UG$^2$ dataset had different degradation characteristics. As such, they proposed to first identify the incoming images as belonging to one of the three collections as a form of quality estimation, and then deploy a specific processing model for each collection. The entire pipeline process an image in $\sim14$ seconds. In their model, they adopted six different enhancement modules (Supp. Fig. 6).

\textbf{(1) Histogram Equalization} balances the distribution of pixel intensities and increases the global contrast of images. To do this, Contrast Limited Adaptive Histogram Equalization (CLAHE) is adopted~\cite{CLAHE}. The image is partitioned into regions and the histogram of the intensities in each is mapped to a more balanced distribution. As the method is applied at the region level, it is more robust to locally strong over-/under-exposures and can preserve edges better. \textbf{(2)} Given that removing blur effects is widely found to be helpful in fast-moving aerial cameras, and/or in low light filming conditions, \textbf{Deblur GAN}~\cite{DeblurGAN} is  employed as an enhancement module in which, with adversarial training, the generator in the network is able to transform a blurred image to a visually sharper one. \textbf{(3) Recurrent Residual Net for Super-Resolution} was previously proposed in~\cite{Yang2017}. Due to the large distance between objects and aerial cameras, low-resolution is a bottleneck for recognizing most objects from UAV photos. This model is a recurrent residual convolutional neural network consisting of six layers and skip-connections. \textbf{(4) Deblocking Net}~\cite{DeblockingNet} is an auto-encoder-based neural network with dilation convolutions to remove blocking effects in videos, which was fine-tuned using the VGG-19 perceptual loss function, after training using JPEG-compressed images. Since lossy video coding for on-board sensors introduced blocking effects in many frames, the adoption of the deblocking net was found to suppress visual artifacts. \textbf{(5) RED-Net}~\cite{REDNet} is trained to restore multiple mixed degradations, including noise and low-resolution together. Images with various noise levels and scale levels are used for training. The network can improve the overall quality of images. \textbf{(6) HDR-Net}~\cite{HDRNet} can further enhance the contrast of images to improve the quality for machine and human analysis. This network learns to produce a set of affine transformations in bilateral space to enhance the image while preserving sharp edges.

%%% Move to supplemental material
%\begin{figure}[!t]
%\centering
%\subfloat[Checkerboard artifacts in an image and %high-intensity spots in the Fourier domain.]{
%    \includegraphics[width=0.30\textwidth]{images/Nort%hwestern_a.png}\label{fig:NorthwesternOut_a}
%}
%\hfil
%\subfloat[Checkerboard artifacts removed in the image %and suppressed spots in the Fourier domain]
%{
%    \includegraphics[width=0.30\textwidth]{images/Nort%hwestern_b.png}\label{fig:NorthwesternOut_b}
%}
%\caption{Northwestern's checkerboard artifact %removal.}
%\label{fig:NorthwesternOut}
%\end{figure}

% \begin{figure}[t]
%     \centering
%     %left bottom right top
%     \includegraphics[clip, trim=4cm 4.3cm 4cm 4.3cm, width=0.45\textwidth]{figures/TAM.pdf}
%     \caption{The CDRM enhancement pipleline. If the glider set is detected, no action is taken (recognition is deemed to be good enough by default).}
%     \label{fig:TAM}
%     \vspace{-2em}
% \end{figure}

%%% Cut for space -- WJS
%Several interesting facts emerged from CDRM. First, as %previously observed, human visibility and machine %perception goals are not always aligned, the same model %with hyperparameters to achieve best classification %accuracy does not necessarily bring the best human %evaluation performance. Second, sequential models can help %while single enhancement model does not necessarily improve %classification. Surprisingly, even a single model that %hurts performance when applied independently, can help a %lot when run together with other models in the right %sequential order.

\subsubsection{Tone Mapping Deep Image Prior (TM-DIP)}
%TsingHua:
The main idea of the National Tsing Hua University team's approach was to derive deep image priors for enhancing images that are captured from a specific scene with certain poor imaging conditions, such as the UG$^2$ collections. They consider the setting that the high-quality counterparts of the poor-quality input images are unavailable, and hence it is not possible to collect pairwise input/output data for end-to-end supervised training to learn how to recover the sharp images from blurry ones.
%% SREYA: No new lines added. Just highlighting the priors used in the algorithm for eaasy identification
The method of Deep Image Prior presented by Ulyanov \etal \cite{DeepImagePrior} can reconstruct images without using information from ground-truth sharp images. However, it usually takes several minutes to produce a prior image by training an individual network for each image. Thus a new method was designed to replace the per-image prior model of \cite{DeepImagePrior} by a generic prior network. This idea is feasible since images taken in the same setting, \eg the UG$^2$ videos, often share similar features. It is not necessary to have a distinct prior model for each image. One can learn a generic prior network that takes every image as input and generates its corresponding prior as output.

At training time, the method from~\cite{DeepImagePrior} is used to generate image pairs $\{(I, V)\}$ for training a generic prior network, where $I$ is an original input image and $V$ is its corresponding prior image. The generic prior network adopts an encoder-decoder architecture with skip connections as in \cite{RonnebergerFB15}. 
At inference time, given a new image, its corresponding prior image is efficiently obtained from the learned generic prior network, with tone mapping then applied to enhance the details.

It was observed that the prior images obtained by the learned generic prior network usually preserve the significant structure of the input images but exhibit fewer details.
This observation, therefore, led to a different line of thought on the image enhancement problem.
By comparing the prior image with the original input image, details for enhancement may be extracted.
Thus, the tone mapping technique presented in \cite{ChenLC05} was used to enhance the details:
\begin{equation}
\label{toneMapping}
\widetilde{I} = \left(\frac{I}{V}\right) ^ \gamma (V) \,,
\end{equation}
\noindent
where the ratio $I/V$ can be considered the details, and $\gamma$ is a factor for adjusting the degree of detail-enhancement. 
With the tone-mapping function in Eq.~\eqref{toneMapping}, the local details are detached from the input image, and the factor $\gamma$ is subsequently adjusted to obtain an enhanced image $\widetilde{I}$. 

\subsubsection{Satellite Images Super-Resolution (SSR)}
%Hopkins:
The team from Johns Hopkins University proposed a neural network-based approach to apply super-resolution on images. They trained their model on satellite imagery, which has an abundance of detailed features. Their network is fully convolutional, and takes as input an image of any resolution and outputs an image that is exactly double the original input in width and height in  $\sim3$ seconds.

% \begin{table}
% \centering
% \subimport{./figures/}{HopkinsLayers.tex}
% \caption{The SSR network layers.}
% \label{tab:HopkinsNetwork}
% \vspace{-6mm}
% \end{table}

The network is constrained to $32 \times 32$ pixel patches of the image with an ``apron" of $2$ pixels for an overlap. This results in a $64 \times 64$ output where the outer $4$ pixels are ignored, as they are the apron --- they mirror the edge to ``pad" the image. These segments are then stitched together to form the final image. The network consists of five convolutional layers (Supp. Table 1). Most of the network's layers contain $1 \times 1$ kernels, and hence are just convolutionalized fully connected layers. This network structure is appropriate for a super-resolution task because it can be equated to a regression problem where the input is a $25$ dimension ($5 \times 5$) vector leading to a $4$ dimensional ($2 \times 2$) vector. The first convolutional layer is necessary to maintain the spatial relationships of the visual features through the $5 \times 5$ kernel.

The SpaceNet dataset\cite{SpaceNet} is used to train this network and is derived from satellite-based images. Images were downsampled and paired with the originals. Training took place for $20$ epochs using an L2 + L1 combined loss and the Adam optimizer in Keras/Tensorflow~\cite{chollet2015keras}.

\subsubsection{Style-Transfer Enhancement Using GANs (ST-GAN)}
%Noblis: 
Noblis attempted a style-transfer approach for improving the quality of the UG\textsuperscript{2} imagery. Since the classification networks used in the UG\textsuperscript{2} evaluation protocol were all trained on ImageNet, a CycleGAN~\cite{cyclegan} variant was trained to translate between the UAV and drone collections and ImageNet, using LSGAN~\cite{lsgan} losses. The architecture was based on the original CycleGAN paper, with modified generators adding skip connections between same spatial resolution convolutional layers on both sides of the residual blocks (in essence a U-Net~\cite{RonnebergerFB15} style network), which appeared to improve retention of details in the output images. This algorithm was able to process each image in less than $4$ seconds. The UG\textsuperscript{2} to ImageNet generator was also made to perform $4\times$ upscaling (by adding two $0.5$ strided convolutional layers after the first convolutional layer), and it was also made to perform $4\times$ downscaling (by adding two stride $2$ convolutional layers after the first convolutional layer). The discriminators were left unmodified. 
The networks were trained using $128\times128$ patches selected from the UG\textsuperscript{2} images, and ImageNet images cropped and resized to $512\times512$. UG\textsuperscript{2} patches were selected by randomly sampling regions around the ground-truth annotation bounding boxes to avoid accidentally sampling flat-colored patches.

However, several problems were initially encountered when optimizing the network. Optimization would fail outright, unless it employed some form of normalization.
%However, it was not possible to employ the %typical batch-norm~\cite{batchnorm}, due to %small batch sizes. Instead, instance %normalization~\cite{instancenorm} was initially %used, which proved reasonably effective. %However, even after avoiding immediate %optimization failure, training a network with %just the cycle-consistency and GAN losses can %lead to mode failures such as both generators %performing color inversion within a few %thousand iterations. 
Adding the identity mapping losses (\textit{i.e.}, loss terms for $G(X)-X$, and $F(Y)-Y$) discussed in the original CycleGAN paper proved effective in avoiding these kinds of failures.
Since the UG$^2$ evaluation protocol specifies the enhancement of full video frames, either a larger input to the generator must be used (which seemed feasible considering a fully-convolutional architecture), or the input image must be divided into tiles. To address this, an operation that performed normalization independently down the channels of each pixel was used. This stabilized convergence and did not cause problems when tiling out very large images.

\subsection{Baseline Algorithms} \label{Methods:Baselines}

%%% This text is irrelevant --- WJS
%Upon the release of the UG$^2$ training dataset, we %provided an analysis of the effect of classic and %state-of-the-art approaches for two kinds of enhancement %techniques on the image classification: (1) Resolution %enhancement, and (2) Deblurring algorithms. We again %evaluate our testing dataset on these algorithms, whose %only focus is to improve the visual quality rather than the %classification accuracy, and compare them to the %participant's submissions for the second image enhancement %task.

%%The algorithms we tested focused on image interpolation~\cite{Keys:1981:Interpolation} (namely Nearest neighbor, Bilinear, and Bicubic interpolation), super-resolution (SRCNN~\cite{Chao:2014:SRCNN}, and VDSR~\cite{Kim:2016:VDSR}), and deblurring (Blind deconvolution~\cite{Pan:2016:DCPD}, and Dynamic Deep Deblurring~\cite{Su:2016:DBN}). We had originally evaluated a third video deblurring algorithm proposed by Na \etal \cite{Nah:2016:DSDD}, however this algorithm employs information across multiple consecutive frames to perform its deblurring operation, given that the participant's algorithms were provided only with disjoint video frames we omitted this method to provide a fair comparison.
The following algorithms serve as canonical references or baselines against which the  algorithms in Sec.~\ref{Methods:Entries} were tested. We used both classical methods and state-of-the-art deep learning-based methods for image interpolation~\cite{Keys:1981:Interpolation}, super-resolution~\cite{Chao:2014:SRCNN, Kim:2016:VDSR}, and deblurring~\cite{Pan:2016:DCPD,Nah:2016:DSDD}

\textbf{Classical Methods.} For image enhancement, we used three different interpolation methods (bilinear, bicubic and nearest neighbor)~\cite{Keys:1981:Interpolation} and a single restoration algorithm (blind deconvolution~\cite{Pan:2016:DCPD}). The interpolation algorithms attempt to obtain a high-resolution image by up-sampling the source low-resolution image and by providing the best approximation of a pixel's color and intensity values depending on the nearby pixels. Since they do not need any prior training, they can be directly applied to any image. Nearest neighbor interpolation uses a weighted average of the nearby translated pixel values in order to calculate the output pixel value. Bilinear interpolation increases the number of translated pixel values to two and bicubic interpolation increases it to four. Different from image enhancement, in image restoration, the degradation, which is the product of motion or depth variation from the object or the camera, is modeled. The blind deconvolution algorithm can be used effectively when no information about the degradation (blur and noise) is known~\cite{kundur1996blind}. The algorithm restores the image and the point-spread function (PSF) simultaneously. We used Matlab's blind deconvolution algorithm, which deconvolves the image using the maximum likelihood algorithm, with a $3\times3$ array of $1$s as the initial PSF.

\textbf{Deep Learning-Based Methods.} With respect to state-of-the-art deep learning-based super-resolution algorithms, we tested the Super-Resolution Convolutional Neural Network (SRCNN)~\cite{Chao:2014:SRCNN} and Very Deep Super Resolution (VDSR)~\cite{Kim:2016:VDSR}. The SRCNN method employs a feed-forward deep CNN to learn an end-to-end mapping between low-resolution and high-resolution images.
The network was trained on $5$ million ``sub-images" generated from $395,909$ images of the ILSVRC 2013 ImageNet detection training partition~\cite{russakovsky2015imagenet}. 
The VDSR algorithm~\cite{Kim:2016:VDSR} outperforms SRCNN by employing a deeper CNN inspired by the VGG architecture~\cite{VGG:2014} and decreases training iterations and time by employing residual learning with a very high learning rate for faster convergence.
Unlike SRCNN, the network is capable of handling different scale factors. 

With respect to deep learning-based image restoration algorithms, we tested Deep Dynamic Scene Deblurring~\cite{Nah:2016:DSDD}, which was designed to address camera shake blur. However, the results presented in~\cite{Nah:2016:DSDD} indicated that this method can obtain good results for other types of blur. The algorithm employs a CNN that was trained with video frames containing synthesized motion blur such that it receives a stack of neighboring frames and returns a deblurred frame. The algorithm allows for three types of frame-to-frame alignment: no alignment, optical flow alignment, and homography alignment. For our experiments, we used optical flow alignment, which was reported to have the best performance with this algorithm. We had originally evaluated an additional video deblurring algorithm proposed by Su \etal \cite{Su:2016:DBN}. However, this algorithm employs information across multiple consecutive frames to perform its deblurring operation. Given that the training and testing partitions of the UG$^2$ dataset consist of disjoint video frames, we omitted this method to provide a fair comparison.

With respect to image enhancement specifically to improve classification, we tested the recently released algorithm by Sharma \etal\cite{Sharma2017}. This approach learns a dynamic image enhancement network with the overall goal to improve classification, but not necessarily the human perception of the image. 
%In their work, Sharma \etal develop a method to jointly optimize a convolutional neural network for enhancement and classification tasks in an end-to-end manner. 
The proposed architecture enhances image features selectively, in such a way that the enhanced features provide a valuable improvement for a classification task. High quality (\textit{i.e.}, free of visual artifacts) images are used to train a CNN to learn a configuration of enhancement filters that can be applied to an input image to yield an enhanced version that provides better classification performance. 

\begin{figure*}[!ht]
\centering

\subfloat[UAV Collection]
{
    \label{fig:LPIPS_UAV}
    \includegraphics[width=0.32\linewidth]{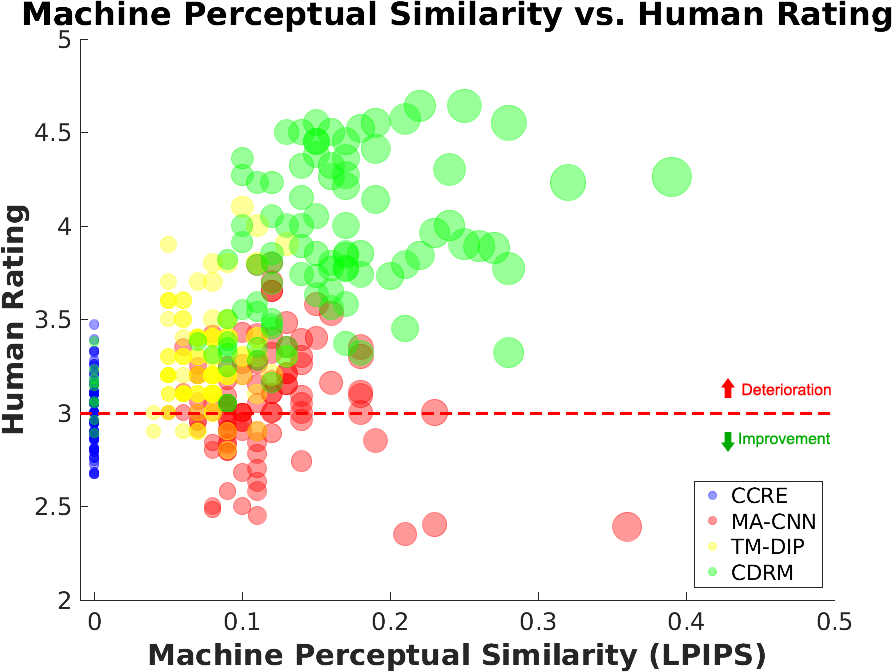}
}
\subfloat[Glider Collection]
{
    \label{fig:LPIPS_Glider}
    \includegraphics[width=0.32\linewidth]{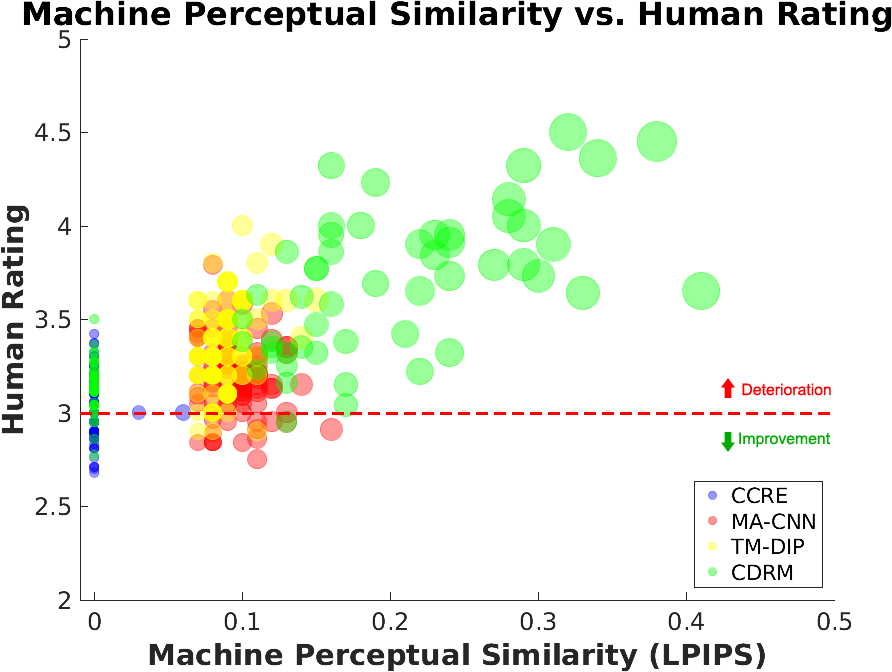}
}
\subfloat[Ground Collection]
{
    \label{fig:LPIPS_Ground}
    \includegraphics[width=0.32\linewidth]{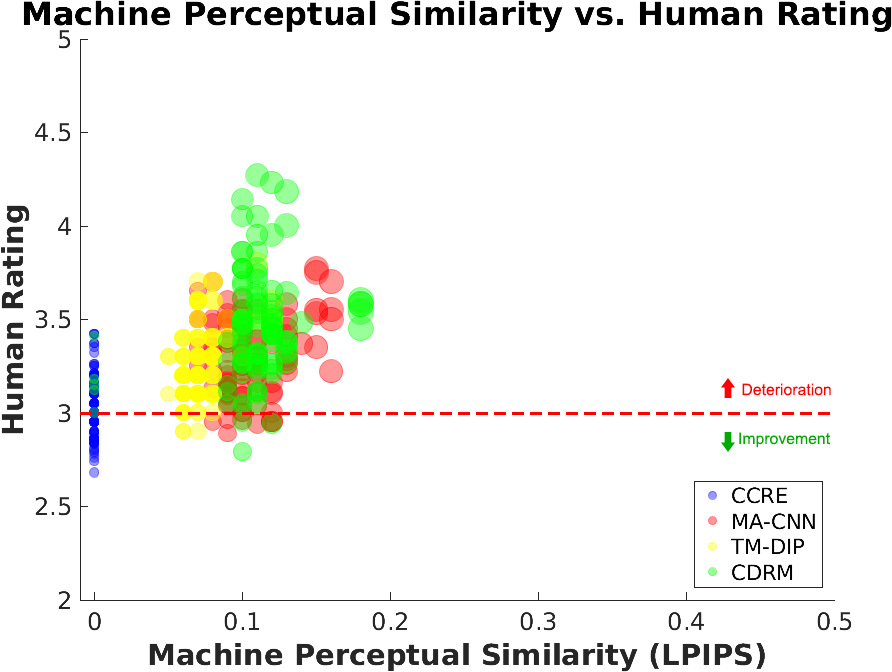}
}
\vfill \vspace{0.5em}
\caption{Distribution of LPIPS similarity between original and enhanced image pairs from four different approaches, and the human perceived improvement / deterioration for each of the collections within the UG$^2$ dataset. Images human raters considered as having a  high-level of improvement tended to also have low LPIPS scores, while images with higher LPIPS scores tended to be rated poorly by human observers.}
\label{fig:LPIPS}
\vspace{-3mm}
\end{figure*}   

\section{Results \& Analysis} \label{ResAnalysis}

In the following analysis, we review the results that came out of the UG$^2$ Workshop held at CVPR 2018, and discuss additional results from the slate of baseline algorithms.

%The undisputed winner for the Image Enhancement to %Facilitate Manual Inspection task was the Honeywell %team, with an overall score of 126 points. The second %place in qualitative visual improvement went to the %Northwestern team, with a total score of 59 points. %The winning classification entry for the Image %Enhancement to Improve Automatic Object Recognition %task was submitted by the Honeywell team, which %excelled in 4 classification categories. The second %place in quantitative visual improvement went to the %TexasA\&M team, which was the best in 3 of the %classification categories.
\vspace{-3mm}
\subsection{Enhancement to Facilitate Manual Inspection (UG$^2$ Evaluation Task 1)}

% \begin{figure*}[!ht]
% \centering

% \subfloat[Each participant's top algorithm. \label{fig:C1_ParticipantScores}]
% {
%     \includegraphics[width=0.3\textwidth]{figures/C1_ParticipantScores.png}
% }
% \subfloat[Classic and state-of-the-art enhancement methods. \label{fig:C1_Baselines}]
% {
%     \includegraphics[width=0.3\textwidth]{figures/C1_BSum.png}
% }
% \subfloat[Comparison of top participant's and other state-of-the-art-methods. \label{fig:C1_PVsB}]
% {
%     \includegraphics[width=0.3\textwidth]{figures/C1_PvsB.png}
% }
% \vfill 

% \caption{Comparison of perceived visual improvement for all collections.}
% \label{fig:graph:C1}
% \end{figure*}  

%Most enhancement algorithms evaluate qualitative %improvement of images through evaluation metrics %such as PSNR or SSIM. However, optimizing PSNR %is primarily minimizing the mean squared %pixel-wise error that often results in blurry %reconstruction and does not capture perceptually %relevant differences, such as high texture %details ~\cite{ledig2017photo}. 

There is a recent trend to use deep features, as measured by the difference in activations from the higher convolutional layers of a pre-trained network for the original and reconstructed images, as a perceptual metric --- the motivation being deep features somewhat mimic human perception. Zhang \etal \cite{DBLP:journals/corr/abs-1801-03924}  evaluate the usability of these deep features as a measure of human perception. Their goal is to find a perceptual distance metric that resembles human judgment. The outcome of their work is the LPIPS metric, which measures the perceptual similarity between two images ranging from $0$, meaning exactly similar, to $1$, which is equivalent to two completely dissimilar images. Here we compare this metric (\textit{i.e.}, similarity between the original image and the output of the evaluated enhancement algorithms) directly to human perception, which we argue is the better reference point for such assessments.  

\begin{figure}[b!]
\centering
% \vspace{-6mm}
\subfloat[Baseline Algs.]
{
    \includegraphics[clip,width=0.68\columnwidth]{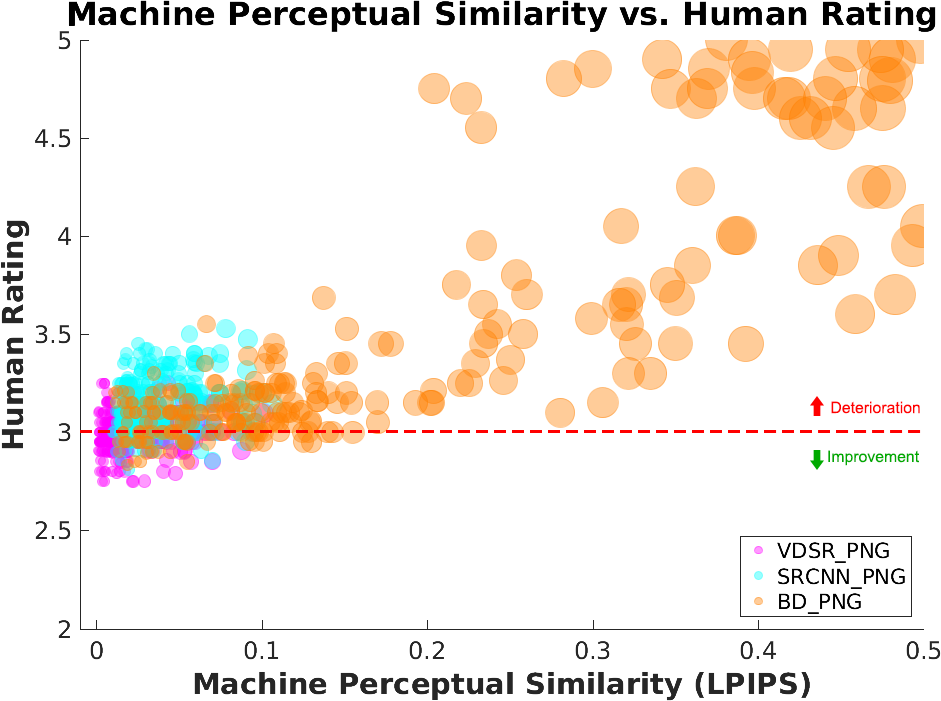}
    \label{fig:LPIPS_Bas}
}

\subfloat[Top Participant vs. Baseline Algs.]
{
    \includegraphics[clip,width=0.68\columnwidth]{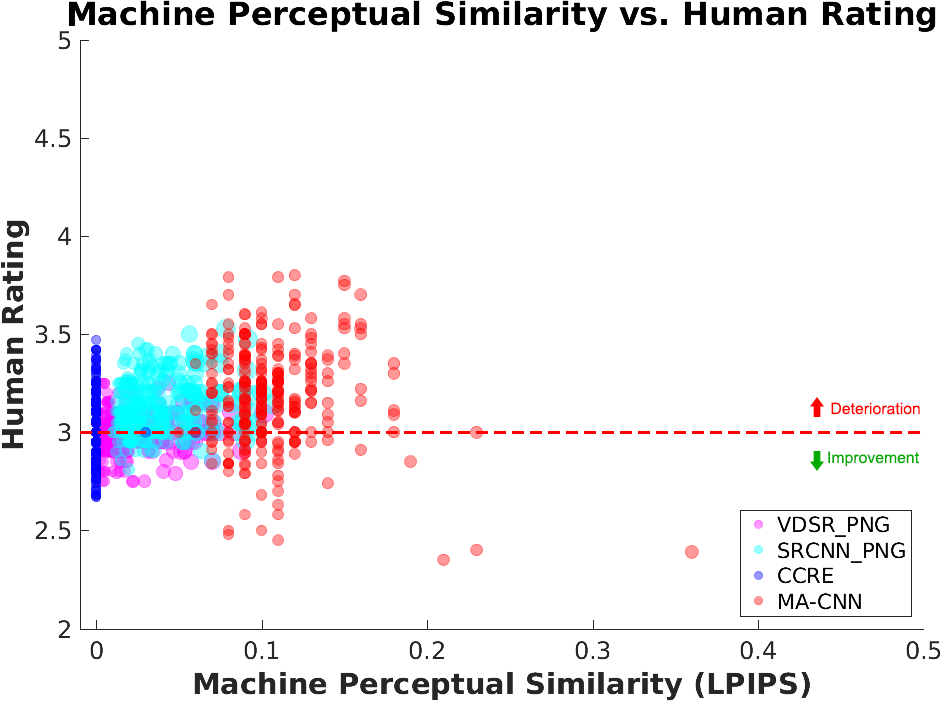}
    \label{fig:LPIPS_PartVsBas}
}
\caption{LPIPS similarity vs. human ratings for the baseline algorithms, over all of the collections within the UG$^2$ dataset.}
\label{fig:LPIPS:Comp}
% \vspace{-6mm}
\end{figure}

We used the most current version (v.0.1) of the LPIPS metric with a pre-trained, linearly calibrated AlexNet model. As can be observed in Fig.~\ref{fig:LPIPS}, the four novel algorithms that were submitted by participants for the first UG$^2$ evaluation task have very heterogeneous effects on different images of the dataset, with LPIPS scores ranging all the way from $0$ (no perceptual dissimilarity) to $0.4$ (moderate dissimilarity). This effect is accentuated for the images in the UAV Collection (Fig.~\ref{fig:LPIPS_UAV}), which yields more variance in LPIPS scores, whereas LPIPS scores for the remaining two collections (Fig.~\ref{fig:LPIPS_Glider}, \ref{fig:LPIPS_Ground}) remained between $0$ and $0.15$ for most of these algorithms. We observed a similar effect for some of our baseline algorithms (Fig.~\ref{fig:LPIPS_Bas}), particularly for Blind Deconvolution (BD), which sharpened images, but also amplified artifacts that were already present. 

The images human raters considered as having a high-level of improvement tended to also have low LPIPS scores (usually between $0$ and $0.15$), while images with higher LPIPS scores tended to be rated negatively by human observers (see Figs.~\ref{fig:LPIPS} and~\ref{fig:LPIPS_Bas}). Similar behavior was observed by Zu \etal \cite{NIPS2017_6650}. They suggest that high LPIPS scores might indicate the presence of unnatural looking images. Contrasting the results of the two best participant and baseline algorithms (Fig.~\ref{fig:LPIPS_PartVsBas}), we observed that for the baselines and CCRE the LPIPS and human rating distributions were more tightly grouped than MA-CNN. Nevertheless, their changes to images tended to be considered small by both human raters and the LPIPS metric, with a user rating closer to $3.0$ (no change) and LPIPS scores between $0$ and $0.1$. In contrast, changes induced by the MA-CNN method reached extremes of $0.36$ and $2.35$ for LPIPS score and human rating respectively, flagging the presence of very noticeable, but in some cases detrimental, changes. This is a further constraint of the LPIPS metric.

\begin{figure*}[!ht]
\centering

\subfloat[Participant Algs.]
{
    \includegraphics[width=0.32\textwidth]{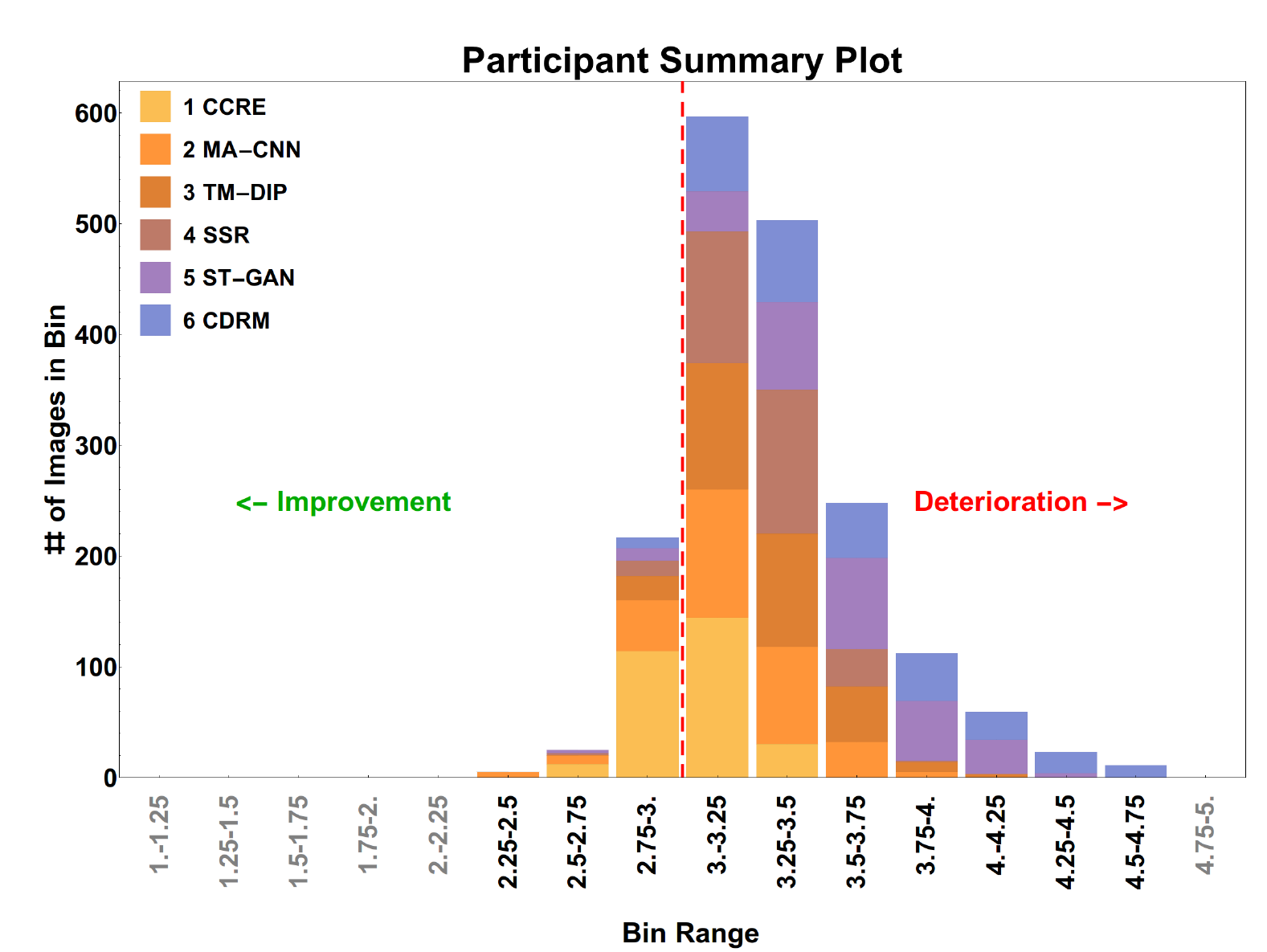}
    \label{fig:C1_ParticipantScores}
}
\subfloat[Baseline Algs.]
{
    \includegraphics[width=0.32\textwidth]{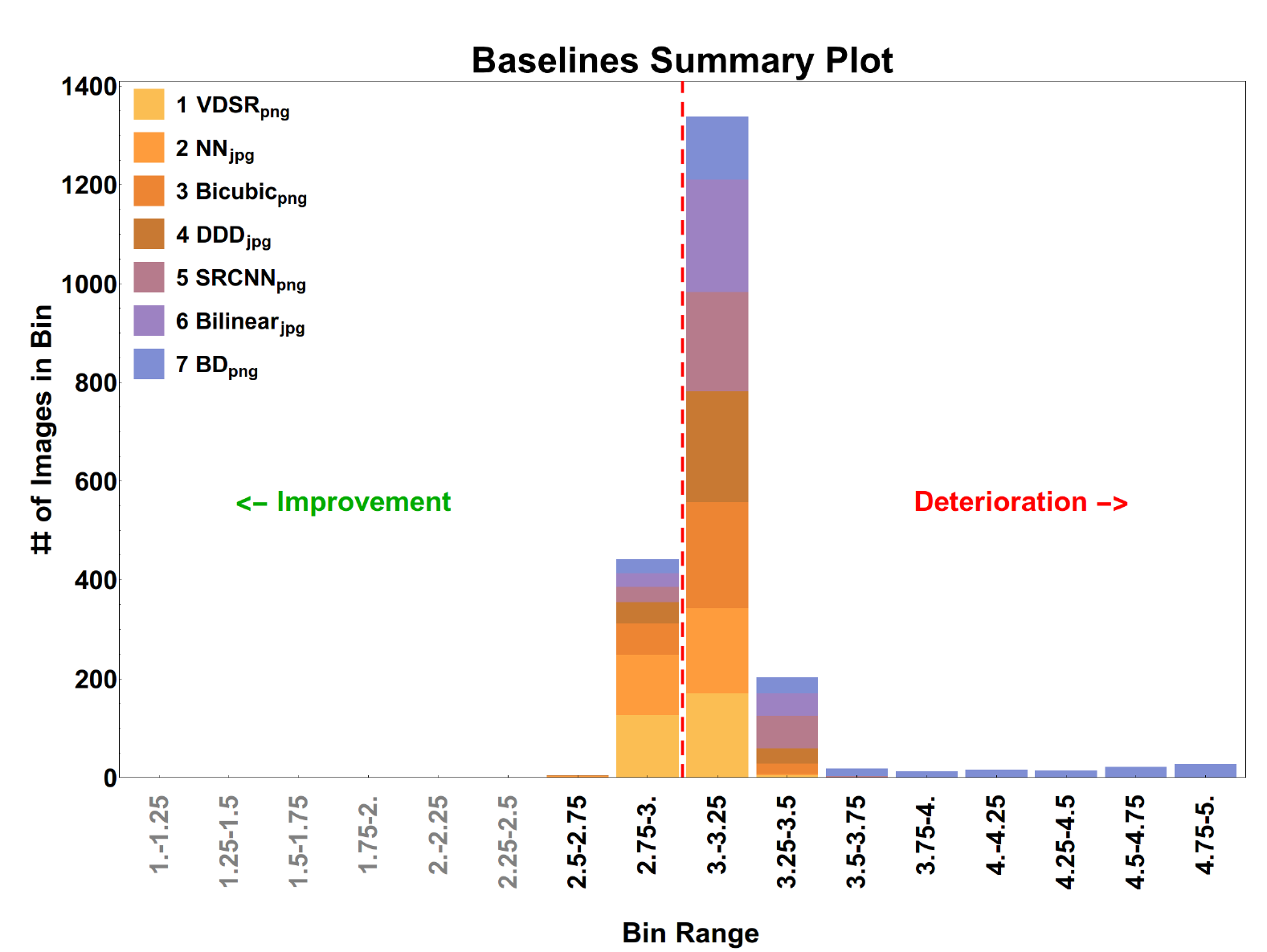}
    \label{fig:C1_Baselines}
}
\subfloat[Top Participant vs. Top Baseline Algs.]
{
    \includegraphics[width=0.32\textwidth]{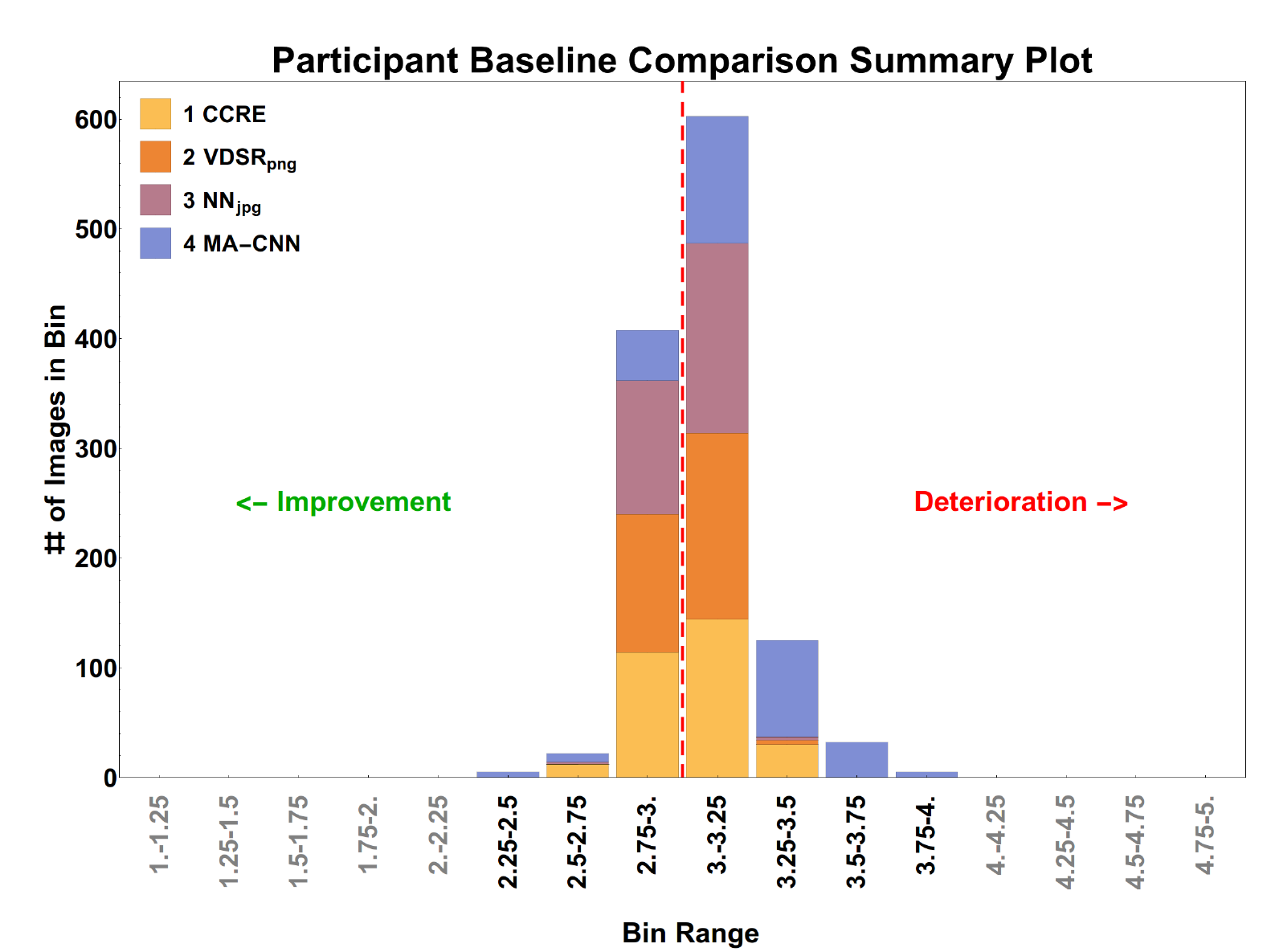}
    \label{fig:C1_PVsB}
}
\vfill \vspace{0.5em}
\caption{Comparison of perceived visual improvement for all collections after applying restoration and enhancement algorithms.}
\label{fig:C1}
\end{figure*} 

It is important to note that while we calculated the mean user rating of all the workshop participant submissions and baseline algorithms, it was not possible to obtain the LPIPS scores for any of the super-resolution approaches. This would have required us to down-sample enhanced images to be of the same size as that of the original images, which would have negated the improvement of such methods.

Focusing just on the image improvement / deterioration as perceived by human raters, we can turn to Fig.~\ref{fig:C1_ParticipantScores} for the performance of all algorithms, including the super-resolution approaches, submitted by each team. It is important to note that while most of the algorithms tended to improve the visual quality of the images they were presented with, a large fraction of the images they enhanced tended to have an average score between $3$ and $3.25$. In such cases, humans could not detect any meaningful change.

The best performing algorithm submitted for this task, CCRE, was able to improve the visual quality of $126$ images, even though the algorithm did not appear to perform any significant changes to most of the images ($144$ images had a rating between $3$ and $3.25$). The enhancement applied was considered a subtle improvement in most scenarios: $114$ images had a score between $2.75$ and $3$, with the remaining $12$ having a higher improvement score between $2.5$ and $2.75$. However, only $10\%$ of the modified images were considered to degrade the image quality, and even then they had a rating of between $3.25$ and $3.5$, which means that the degradation was very small. 

As mentioned previously, the visual changes generated by the runner-up enhancement algorithm MA-CNN were more explicit than those present in CCRE. While the number of images that were considered to be improved was smaller ($59$ improved images), $13$ of them were between the range of $2.25$ and $2.75$, indicating a good measure of higher visual quality. Nevertheless, the sharp changes introduced by this algorithm also seemed to increase the perceived degradation on a larger portion of the images, with almost $25\%$ having a score between $3.5$ and $3.75$ (thus indicating a significant deterioration of the image quality).

% \begin{figure}[h]
%     \centering
%     \includegraphics[width=0.45\textwidth]{figures/C1_ParticipantScores.pdf}
%     \caption{Comparison of perceived visual improvement for all collections after applying each team's top algorithm.}
%     \label{fig:C1_ParticipantScores}
% \end{figure}

% \begin{figure}[t]
%     \centering
%     \includegraphics[width=0.45\textwidth]{figures/C1_BSumf.pdf}
%     \caption{Perceived visual improvement of classic and state-of-the-art enhancement methods.}
%     \label{fig:C1_Baselines}
% \end{figure}

% \begin{figure}[t]
%     \centering
%     \includegraphics[width=0.45\textwidth]{figures/C1_PvsB.pdf}
%     \caption{Comparison of perceived visual improvement between the top algorithms and the evaluated methods.}
%     \label{fig:C1_PVsB}
% \end{figure}

With respect to the baseline algorithms, we observed a less dramatic perception of quality degradation. Given that most of the algorithms tested were focused on enhancing the image resolution (by performing image interpolation or super-resolution), they were more prone to perform very subtle changes in the structure of the image. This is reflected in Fig.~\ref{fig:C1_Baselines}. Fig.~\ref{fig:C1_PVsB} shows a side by side comparison of the two best baselines (VDSR and Nearest Neighbor interpolation) and the two best performing participant submissions.

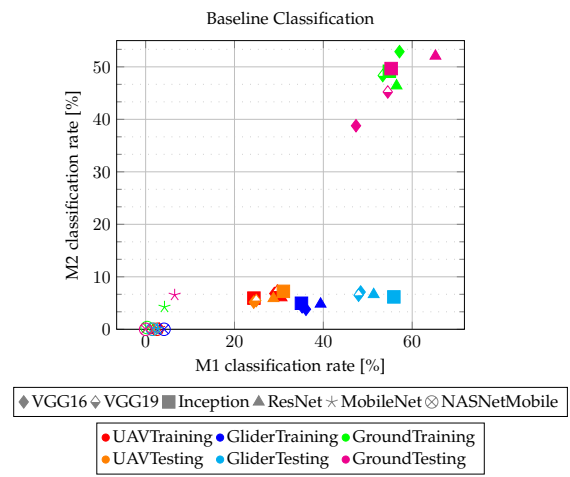
\begin{figure}[b!]
\centering
\resizebox{.85\linewidth}{!}{\pgfplotscreateplotcyclelist{my colors}{
        % color for the legend
        black!50\\
        % color for the "real" plots
        red\\
        blue\\
        green\\
        orange\\
        cyan\\
        magenta\\
    }
    \pgfplotsset{
        compat=1.3,
        cycle list name=my colors,
        legend cell align=left,
    }

\begin{tikzpicture}
    \begin{axis}[
        title={Baseline Classification},
        xlabel={M1 classification rate [\%]},
        ylabel={M2 classification rate [\%]},
        ymin=0.0,ymax=55,
        minor y tick num=2,
        ymajorgrids=true,
        xmajorgrids=true,
        yminorgrids=true,
        minor grid style=loosely dotted,
        only marks,
        scatter,
        mark size=3.5pt,
        scatter src=explicit symbolic,
        table/meta=Method,
        scatter/classes={
            VGG16={mark=diamond*},
            VGG19={mark=halfdiamond*},
            Inception={mark=square*},
            ResNet={mark=triangle*},
            MobileNet={mark=star},
            NASNetMobile={mark=otimes}
        },
        % this legend only shows the defined scatter classes
        % (as it is stated in the manual)
        legend entries={
            VGG16,
            VGG19,
            Inception,
            ResNet,
            MobileNet,
            NASNetMobile%
        },
        %legend pos= north west,
        %legend pos=outer north east,
        legend style={  at={(0.5,-0.2)},
                        anchor=north,legend columns=-1}
    ]

        % ---------------------------------------------------------------------
        % dummy plot for the legend
        % (make sure the expression values are outside the visible axis limits
        %  if this `\addplot wouldn't be present. This requires at least
        %  setting one of the limits explicitly, i.e. in this case `xmin')
        \addplot table [
            x expr=-10,
            y expr=-10,
        ] {figures/Baselines.dat};
        % ---------------------------------------------------------------------

        % simplified the call for the data
        \foreach \i in {
            UAVTraining,
            GliderTraining,
            GroundTraining,
            UAVTesting,
            GliderTesting,
            GroundTesting%
        }{
            \addplot table [
                x=\i-M1,
                y=\i-M2,
            ] {figures/Baselines.dat};
        }

    \end{axis}

    % this is a dummy `axis' environment only to create the second legend
    \begin{axis}[
        % set some axis limits and plot the coordinates outside that box
        % so they don't show up
        xmin=1,
        xmax=2,
        ymin=1,
        ymax=2,
        % of course we also don't want to show this axis
        hide axis,
        % we need only marks
        only marks,
        % state the legend entries for the second legend
        % (here we don't have scatter classes, so each `\addplot' gets its
        %  own entry in the legend)
        legend entries={
            ,       % the dummy plot should not show up in the legend
            UAVTraining,
            GliderTraining,
            GroundTraining,
            UAVTesting,
            GliderTesting,
            GroundTesting%
        },
        % place it below the other legend
        % therefore we have to shift it down (manually)
        %legend pos=outer north east,
        legend style={  at={(0.5,-0.2)},
                        anchor=north,
                        legend columns=3,
                        yshift=-20pt}
        %legend style={
         %   yshift=-60pt,
        %},
    ]
        % just add some dummy plots to create the legend
        \foreach \i in {0,...,6} {
            \addplot+ [mark=*] coordinates { (0,0) };
        }
    \end{axis}
\end{tikzpicture}}
\caption{Classification rates at rank 5 for the original, un-processed, frames for each collection in the training and testing datasets.}
\label{fig:graph:baselines}
\end{figure}

\subsection{Evaluation of Object Recognition Performance (UG$^2$ Evaluation Task 2)}

The results for this evaluation task fill a gap in our knowledge on the effects certain restoration and enhancement algorithms have on visual recognition. In the following section, we introduce results for various algorithms ranging from image interpolation to deep learning-based approaches that are designed solely for the purpose of improving the ``perceived'' visual quality of an image. We then compare the performance of such techniques to the enhancement approaches intentionally designed to improve object classification.

For this evaluation task, the participants were expected to provide enhancement techniques catering to machine vision rather than human perception. Supp. Table~2 and  Fig.~\ref{fig:graph:baselines} depict the baseline classification results for the UG$^2$ training and testing datasets, without any restoration or enhancement algorithm applied, at rank 5. 

Given the very poor quality of its videos, the UAV Collection proved to be the most challenging for all networks in terms of object classification, leading to the lowest classification performance out of the three collections. While the Glider Collection shares similar problematic conditions with the UAV Collection, the images in this collection lead to a higher classification rate than those in the UAV Collection in terms of identifying at least one correctly classified synset class (metric M1). This improvement might be caused by the limited degree of movement of the gliders, since it ensures that the displacement between frames was kept more stable over time, as well as a higher recording quality (taking into consideration the camera weight limitations present in small UAVs are no longer a limiting factor for this collection's videos). The controlled Ground Collection yielded the highest classification rates, which, in an absolute sense, are still low (the highest classification rate being $65.23\%$ for metric M1 and $52.06\%$ for metric M2 for the testing dataset; see Supp. Table~2). The participants of the UG$^2$ workshop were expected to develop algorithms to improve upon the baseline scores.

Additionally, we were interested in assessing the performance of the participants' enhancement methods when applied to light-weight classification models. As such, we employed two additional networks designed for mobile and embedded vision applications: MobileNet (v2) and NASNetMobile. The performance follows similar patterns to the ones obtained from larger classification models: (1) the UAV Collection remains the most challenging for both networks, while the Ground Collection obtained the best results out of the three collections; (2) the sharp difference in the performance on the two metrics remained. However, both mobile networks obtained poor results  (when compared to the larger networks) when classifying the UG$^2$ data. It is likely that these models are specialized towards ImageNet-like images, and have more trouble generalizing to new data without further fine-tuning.

\begin{figure*}[!ht]
\centering

\subfloat[\label{fig:graph:C2Sharma:UAV}]
{
    \pgfplotscreateplotcyclelist{my colors}{
        % color for the legend
        black!50\\
        % color for the "real" plots
        red\\
        blue\\
        green\\
        magenta\\
        yellow\\
        cyan\\
    }
    \pgfplotsset{
        compat=newest,
        cycle list name=my colors,
        legend cell align=left,
        width=0.33\textwidth,
    }

\begin{tikzpicture}
    \begin{axis}
        [
            title={UAV Collection},
            xlabel={M1 classification rate [\%]},
            ylabel={M2 classification rate [\%]},
            ymin=0.0,ymax=8,
            minor y tick num=2,
            ymajorgrids=true,
            xmajorgrids=true,
            yminorgrids=true,
            minor grid style=loosely dotted,
            only marks,
            scatter,
            mark size=2.5pt,
            scatter src=explicit symbolic,
            table/meta=Method,
            scatter/classes={
                VGG16={mark=diamond*},
                VGG19={mark=halfdiamond*},
                Inception={mark=square*},
                ResNet={mark=triangle*},
                MobileNet={mark=star},
                NASNetMobile={mark=otimes}
            },
            % this legend only shows the defined scatter classes
            % (as it is stated in the manual)
            % legend entries={
            %     VGG16,
            %     VGG19,
            %     Inception,
            %     ResNet%
            % },
            % %legend pos= north west,
            % %legend pos=outer north east,
            % legend style={  at={(0.5,-0.2)},
            %                 anchor=north,legend columns=-1}
        ]

        % ---------------------------------------------------------------------
        % dummy plot for the legend
        % (make sure the expression values are outside the visible axis limits
        %  if this `\addplot wouldn't be present. This requires at least
        %  setting one of the limits explicitly, i.e. in this case `xmin')
        \addplot table [
            x expr=-10,
            y expr=-10,
        ] {figures/C2_Sharma_UAV.dat};
        % ---------------------------------------------------------------------

        % simplified the call for the data
        \foreach \i in {
            Baseline,
            BF,
            GF,
            HE,
            IMSHARP, 
            WLS%
        }{
            \addplot table [
                x=\i-M1,
                y=\i-M2,
            ] {figures/C2_Sharma_UAV.dat};
        }

    \end{axis}

    % this is a dummy `axis' environment only to create the second legend
    % \begin{axis}[
    %     % set some axis limits and plot the coordinates outside that box
    %     % so they don't show up
    %     xmin=1,
    %     xmax=2,
    %     ymin=1,
    %     ymax=2,
    %     % of course we also don't want to show this axis
    %     hide axis,
    %     % we need only marks
    %     only marks,
    %     % state the legend entries for the second legend
    %     % (here we don't have scatter classes, so each `\addplot' gets its
    %     %  own entry in the legend)
    %     legend entries={
    %         ,       % the dummy plot should not show up in the legend
    %         Baseline,
    %         BF,
    %         GF,
    %         HE,
    %         IMSHARP,
    %         WLS,       
    %
    %     },
    %     % place it below the other legend
    %     % therefore we have to shift it down (manually)
    %     %legend pos=outer north east,
    %     legend style={  at={(0.5,-0.2)},
    %                     anchor=north,
    %                     legend columns=3,
    %                     yshift=-20pt}
    %     %legend style={
    %      %   yshift=-60pt,
    %     %},
    % ]
    %     % just add some dummy plots to create the legend
    %     \foreach \i in {0,...,5} {
    %         \addplot+ [mark=*] coordinates { (0,0) };
    %     }
    % \end{axis}
\end{tikzpicture}
}
\subfloat[\label{fig:graph:C2Sharma:Glider}]
{
    \pgfplotscreateplotcyclelist{my colors}{
        % color for the legend
        black!50\\
        % color for the "real" plots
        red\\
        blue\\
        green\\
        magenta\\
        yellow\\
        cyan\\
    }
    \pgfplotsset{
        compat=newest,
        cycle list name=my colors,
        legend cell align=left,
        width=0.33\textwidth,
    }

\begin{tikzpicture}
    \begin{axis}
        [
            title={Glider Collection},
            xlabel={M1 classification rate [\%]},
            ylabel={M2 classification rate [\%]},
            ymin=0.0,ymax=8,
            minor y tick num=2,
            ymajorgrids=true,
            xmajorgrids=true,
            yminorgrids=true,
            minor grid style=loosely dotted,
            only marks,
            scatter,
            mark size=2.5pt,
            scatter src=explicit symbolic,
            table/meta=Method,
            scatter/classes={
                VGG16={mark=diamond*},
                VGG19={mark=halfdiamond*},
                Inception={mark=square*},
                ResNet={mark=triangle*},
                MobileNet={mark=star},
                NASNetMobile={mark=otimes}
            },
            % this legend only shows the defined scatter classes
            % (as it is stated in the manual)
            % legend entries={
            %     VGG16,
            %     VGG19,
            %     Inception,
            %     ResNet%
            % },
            % %legend pos= north west,
            % %legend pos=outer north east,
            % legend style={  at={(0.5,-0.2)},
            %                 anchor=north,legend columns=-1}
        ]

        % ---------------------------------------------------------------------
        % dummy plot for the legend
        % (make sure the expression values are outside the visible axis limits
        %  if this `\addplot wouldn't be present. This requires at least
        %  setting one of the limits explicitly, i.e. in this case `xmin')
        \addplot table [
            x expr=-10,
            y expr=-10,
        ] {figures/C2_Sharma_Glider.dat};
        % ---------------------------------------------------------------------

        % simplified the call for the data
        \foreach \i in {
            Baseline,
            BF,
            GF,
            HE,
            IMSHARP, 
            WLS%
        }{
            \addplot table [
                x=\i-M1,
                y=\i-M2,
            ] {figures/C2_Sharma_Glider.dat};
        }

    \end{axis}

    % this is a dummy `axis' environment only to create the second legend
    % \begin{axis}[
    %     % set some axis limits and plot the coordinates outside that box
    %     % so they don't show up
    %     xmin=1,
    %     xmax=2,
    %     ymin=1,
    %     ymax=2,
    %     % of course we also don't want to show this axis
    %     hide axis,
    %     % we need only marks
    %     only marks,
    %     % state the legend entries for the second legend
    %     % (here we don't have scatter classes, so each `\addplot' gets its
    %     %  own entry in the legend)
    %     legend entries={
    %         ,       % the dummy plot should not show up in the legend
    %         Baseline,
    %         BF,
    %         GF,
    %         HE,
    %         IMSHARP,
    %         WLS,       
    %
    %     },
    %     % place it below the other legend
    %     % therefore we have to shift it down (manually)
    %     %legend pos=outer north east,
    %     legend style={  at={(0.5,-0.2)},
    %                     anchor=north,
    %                     legend columns=3,
    %                     yshift=-20pt}
    %     %legend style={
    %      %   yshift=-60pt,
    %     %},
    % ]
    %     % just add some dummy plots to create the legend
    %     \foreach \i in {0,...,5} {
    %         \addplot+ [mark=*] coordinates { (0,0) };
    %     }
    % \end{axis}
\end{tikzpicture}
}
\subfloat[\label{fig:graph:C2Sharma:Ground}]
{
    \pgfplotscreateplotcyclelist{my colors}{
        % color for the legend
        black!50\\
        % color for the "real" plots
        red\\
        blue\\
        green\\
        magenta\\
        yellow\\
        cyan\\
    }
    \pgfplotsset{
        compat=newest,
        cycle list name=my colors,
        legend cell align=left,
        width=0.33\textwidth,
    }

\begin{tikzpicture}
    \begin{axis}
        [
            title={Ground Collection},
            xlabel={M1 classification rate [\%]},
            ylabel={M2 classification rate [\%]},
            ymin=0.0,ymax=55,
            minor y tick num=2,
            ymajorgrids=true,
            xmajorgrids=true,
            yminorgrids=true,
            minor grid style=loosely dotted,
            only marks,
            scatter,
            mark size=2.5pt,
            scatter src=explicit symbolic,
            table/meta=Method,
            scatter/classes={
                VGG16={mark=diamond*},
                VGG19={mark=halfdiamond*},
                Inception={mark=square*},
                ResNet={mark=triangle*},
                MobileNet={mark=star},
                NASNetMobile={mark=otimes}
            },
            % this legend only shows the defined scatter classes
            % (as it is stated in the manual)
            % legend entries={
            %     VGG16,
            %     VGG19,
            %     Inception,
            %     ResNet%
            % },
            % %legend pos= north west,
            % %legend pos=outer north east,
            % legend style={  at={(0.5,-0.2)},
            %                 anchor=north,legend columns=-1}
        ]

        % ---------------------------------------------------------------------
        % dummy plot for the legend
        % (make sure the expression values are outside the visible axis limits
        %  if this `\addplot wouldn't be present. This requires at least
        %  setting one of the limits explicitly, i.e. in this case `xmin')
        \addplot table [
            x expr=-10,
            y expr=-10,
        ] {figures/C2_Sharma_Ground.dat};
        % ---------------------------------------------------------------------

        % simplified the call for the data
        \foreach \i in {
            Baseline,
            BF,
            GF,
            HE,
            IMSHARP, 
            WLS%
        }{
            \addplot table [
                x=\i-M1,
                y=\i-M2,
            ] {figures/C2_Sharma_Ground.dat};
        }

    \end{axis}

    % this is a dummy `axis' environment only to create the second legend
    % \begin{axis}[
    %     % set some axis limits and plot the coordinates outside that box
    %     % so they don't show up
    %     xmin=1,
    %     xmax=2,
    %     ymin=1,
    %     ymax=2,
    %     % of course we also don't want to show this axis
    %     hide axis,
    %     % we need only marks
    %     only marks,
    %     % state the legend entries for the second legend
    %     % (here we don't have scatter classes, so each `\addplot' gets its
    %     %  own entry in the legend)
    %     legend entries={
    %         ,       % the dummy plot should not show up in the legend
    %         Baseline,
    %         BF,
    %         GF,
    %         HE,
    %         IMSHARP,
    %         WLS,       
    %
    %     },
    %     % place it below the other legend
    %     % therefore we have to shift it down (manually)
    %     %legend pos=outer north east,
    %     legend style={  at={(0.5,-0.2)},
    %                     anchor=north,
    %                     legend columns=3,
    %                     yshift=-20pt}
    %     %legend style={
    %      %   yshift=-60pt,
    %     %},
    % ]
    %     % just add some dummy plots to create the legend
    %     \foreach \i in {0,...,5} {
    %         \addplot+ [mark=*] coordinates { (0,0) };
    %     }
    % \end{axis}
\end{tikzpicture}
}
\vfill \vspace{0.5em}
\begin{tikzpicture}
    \begin{axis}[%
        hide axis,
        xmin=10,
        xmax=50,
        ymin=0,
        ymax=0.4,
        only marks,
        mark size=3.5pt,
        legend style={
            draw=white!15!black,
            legend cell align=left, 
            legend columns=6,
        }
    ]
    \addlegendimage{black!50,mark=diamond*}
    \addlegendentry{VGG16};
    \addlegendimage{black!50,mark=halfdiamond*}
    \addlegendentry{VGG19};
    \addlegendimage{black!50,mark=square*}
    \addlegendentry{Inception};
    \addlegendimage{black!50,mark=triangle*}
    \addlegendentry{ResNet};
    \addlegendimage{black!50,mark=star}
    \addlegendentry{MobileNet};
    \addlegendimage{black!50,mark=otimes}
    \addlegendentry{NASNetMobile};
    % \addlegendimage{empty legend}
    % \addlegendentry{};
    
    \addlegendimage{red,mark=*}
    \addlegendentry{Baseline};
    
    \addlegendimage{blue,mark=*}
    \addlegendentry{Bilateral Filtering};
    
    \addlegendimage{green,mark=*}
    \addlegendentry{Guided Filters};
    
    \addlegendimage{magenta,mark=*}
    \addlegendentry{Histogram Equalization};
    
    \addlegendimage{yellow,mark=*}
    \addlegendentry{Image sharpening};
    
    \addlegendimage{cyan,mark=*}
    \addlegendentry{Weighted Least Squares};

    \end{axis}
\end{tikzpicture}\vspace{-2em}
\caption{Comparison of classification rates at rank 5 for each collection after applying classification driven image enhancement algorithms by Sharma \etal \cite{Sharma2017}. Markers in red indicate results on original images.}
\label{fig:graph:C2Sharma}
\end{figure*}
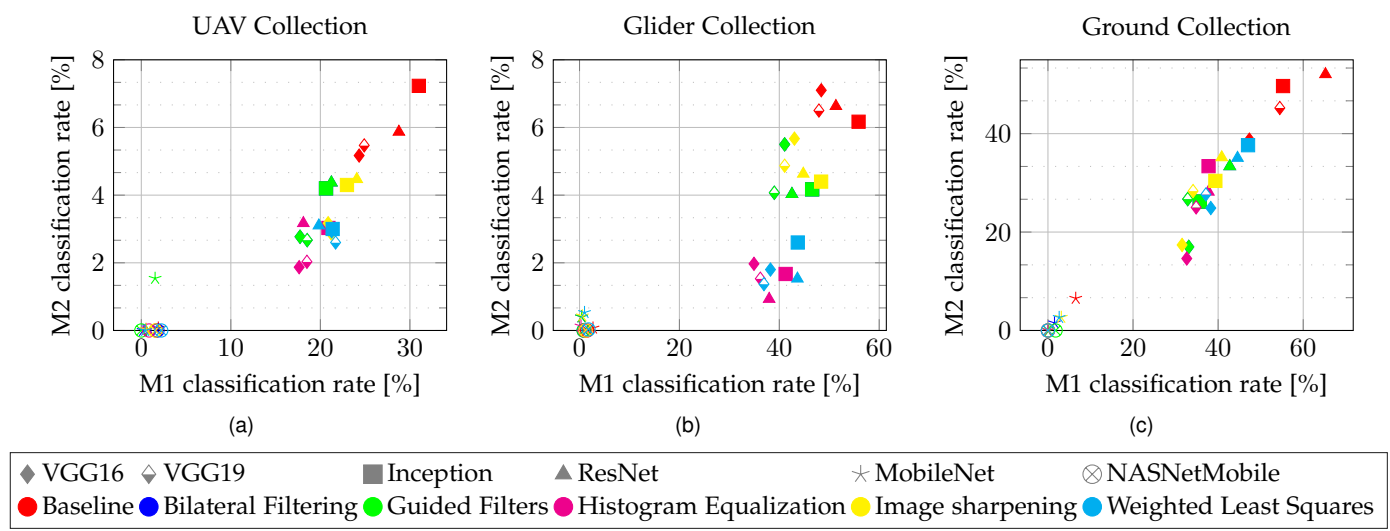

While there are some correlations between the improvement of visual quality as perceived by humans and high-level tasks such as object classification performed by networks, research by Sharma~\etal\cite{Sharma2017} suggests that image enhancement focused on improving image characteristics valuable for object recognition can lead to an increase in classification performance. Thus we tried such a technique as an initial experiment and compared it to the baseline results.  Fig.~\ref{fig:graph:C2Sharma} shows the performance of the five enhancement filters proposed by Sharma \etal on the UG$^2$ dataset. It is important to note that said filters were the un-altered filters Sharma \etal trained making use of good quality images. This is because their focus was on improving the classification performance of images with few existing perturbations. As such, the effect on improving highly corrupted images is much different from that obtained on a standard dataset of images crawled from the web. The results in Fig.~\ref{fig:graph:C2Sharma} establish that even existing deep learning networks designed for this task cannot achieve good classification rates for UG$^2$ due to the domain shift in training.
%fig:graph:C2Sharma

\begin{figure*}[!ht]
\centering
\subfloat[\label{fig:graph:C2Participants:UAV}]
{
    \pgfplotscreateplotcyclelist{my colors}{
        % color for the legend
        black!50\\
        % color for the "real" plots
        red\\
        blue\\
        green\\
        magenta\\
        yellow\\
        cyan\\
    }
    \pgfplotsset{
        compat=newest,
        cycle list name=my colors,
        legend cell align=left,
        width=0.33\textwidth,
    }

\begin{tikzpicture}
    \begin{axis}
        [
            title={UAV Collection},
            xlabel={M1 classification rate [\%]},
            ylabel={M2 classification rate [\%]},
            ymin=0.0,ymax=10,
            minor y tick num=2,
            ymajorgrids=true,
            xmajorgrids=true,
            yminorgrids=true,
            minor grid style=loosely dotted,
            only marks,
            scatter,
            mark size=2.5pt,
            scatter src=explicit symbolic,
            table/meta=Method,
            scatter/classes={
                VGG16={mark=diamond*},
                VGG19={mark=halfdiamond*},
                Inception={mark=square*},
                ResNet={mark=triangle*},
                MobileNet={mark=star},
                NASNetMobile={mark=otimes}
            },
            % this legend only shows the defined scatter classes
            % (as it is stated in the manual)
            % legend entries={
            %     VGG16,
            %     VGG19,
            %     Inception,
            %     ResNet%
            % },
            % %legend pos= north west,
            % %legend pos=outer north east,
            % legend style={  at={(0.5,-0.2)},
            %                 anchor=north,legend columns=-1}
        ]

        % ---------------------------------------------------------------------
        % dummy plot for the legend
        % (make sure the expression values are outside the visible axis limits
        %  if this `\addplot wouldn't be present. This requires at least
        %  setting one of the limits explicitly, i.e. in this case `xmin')
        \addplot table [
            x expr=-10,
            y expr=-10,
        ] {figures/C2_PUAV.dat};
        % ---------------------------------------------------------------------

        % simplified the call for the data
        \foreach \i in {
            Baseline,
            Honeywell,
            TexasAMPeking,
            Northwestern,
            TsingHua%
        }{
            \addplot table [
                x=\i-M1,
                y=\i-M2,
            ] {figures/C2_PUAV.dat};
        }

    \end{axis}

    % this is a dummy `axis' environment only to create the second legend
    % \begin{axis}[
    %     % set some axis limits and plot the coordinates outside that box
    %     % so they don't show up
    %     xmin=1,
    %     xmax=2,
    %     ymin=1,
    %     ymax=2,
    %     % of course we also don't want to show this axis
    %     hide axis,
    %     % we need only marks
    %     only marks,
    %     % state the legend entries for the second legend
    %     % (here we don't have scatter classes, so each `\addplot' gets its
    %     %  own entry in the legend)
    %     legend entries={
    %         ,       % the dummy plot should not show up in the legend
    %         Baseline,
    %         Honeywell,
    %         TexasAMPeking,
    %         Northwestern,
    %         TsingHua%
    %     },
    %     % place it below the other legend
    %     % therefore we have to shift it down (manually)
    %     %legend pos=outer north east,
    %     legend style={  at={(0.5,-0.2)},
    %                     anchor=north,
    %                     legend columns=3,
    %                     yshift=-20pt}
    %     %legend style={
    %      %   yshift=-60pt,
    %     %},
    % ]
    %     % just add some dummy plots to create the legend
    %     \foreach \i in {0,...,5} {
    %         \addplot+ [mark=*] coordinates { (0,0) };
    %     }
    % \end{axis}
\end{tikzpicture}
}
\subfloat[\label{fig:graph:C2Participants:Glider}]
{
    \pgfplotscreateplotcyclelist{my colors}{
        % color for the legend
        black!50\\
        % color for the "real" plots
        red\\
        blue\\
        green\\
        magenta\\
        yellow\\
        cyan\\
    }
    \pgfplotsset{
        compat=newest,
        cycle list name=my colors,
        legend cell align=left,
        width=0.33\textwidth,
    }

\begin{tikzpicture}
    \begin{axis}[
        title={Glider Collection},
        xlabel={M1 classification rate [\%]},
        ylabel={M2 classification rate [\%]},
        ymin=0.0,ymax=10,
        xmin=0,
        minor y tick num=2,
        ymajorgrids=true,
        xmajorgrids=true,
        yminorgrids=true,
        minor grid style=loosely dotted,
        only marks,
        scatter,
        mark size=2.5pt,
        scatter src=explicit symbolic,
        table/meta=Method,
        scatter/classes={
            VGG16={mark=diamond*},
            VGG19={mark=halfdiamond*},
            Inception={mark=square*},
            ResNet={mark=triangle*},
            MobileNet={mark=star},
            NASNetMobile={mark=otimes}
        },
        % this legend only shows the defined scatter classes
        % (as it is stated in the manual)
        % legend entries={
        %     VGG16,
        %     VGG19,
        %     Inception,
        %     ResNet%
        % },
        % %legend pos= north west,
        % %legend pos=outer north east,
        % legend style={  at={(0.5,-0.2)},
        %                 anchor=north,legend columns=-1}
    ]

        % ---------------------------------------------------------------------
        % dummy plot for the legend
        % (make sure the expression values are outside the visible axis limits
        %  if this `\addplot wouldn't be present. This requires at least
        %  setting one of the limits explicitly, i.e. in this case `xmin')
        \addplot table [
            x expr=-10,
            y expr=-10,
        ] {figures/C2_PGlider.dat};
        % ---------------------------------------------------------------------

        % simplified the call for the data
        \foreach \i in {
            Baseline,
            Honeywell,
            TexasAMPeking,
            Northwestern,
            TsingHua%
        }{
            \addplot table [
                x=\i-M1,
                y=\i-M2,
            ] {figures/C2_PGlider.dat};
        }

    \end{axis}

    % this is a dummy `axis' environment only to create the second legend
    % \begin{axis}[
    %     % set some axis limits and plot the coordinates outside that box
    %     % so they don't show up
    %     xmin=1,
    %     xmax=2,
    %     ymin=1,
    %     ymax=2,
    %     % of course we also don't want to show this axis
    %     hide axis,
    %     % we need only marks
    %     only marks,
    %     % state the legend entries for the second legend
    %     % (here we don't have scatter classes, so each `\addplot' gets its
    %     %  own entry in the legend)
    %     legend entries={
    %         ,       % the dummy plot should not show up in the legend
    %         Baseline,
    %         Honeywell,
    %         TexasAMPeking,
    %         Northwestern,
    %         TsingHua%
    %     },
    %     % place it below the other legend
    %     % therefore we have to shift it down (manually)
    %     %legend pos=outer north east,
    %     legend style={  at={(0.5,-0.2)},
    %                     anchor=north,
    %                     legend columns=3,
    %                     yshift=-20pt}
    %     %legend style={
    %      %   yshift=-60pt,
    %     %},
    % ]
    %     % just add some dummy plots to create the legend
    %     \foreach \i in {0,...,5} {
    %         \addplot+ [mark=*] coordinates { (0,0) };
    %     }
    % \end{axis}
\end{tikzpicture}
}
\subfloat[\label{fig:graph:C2Participants:Ground}]
{
    \pgfplotscreateplotcyclelist{my colors}{
        % color for the legend
        black!50\\
        % color for the "real" plots
        red\\
        blue\\
        green\\
        magenta\\
        yellow\\
        cyan\\
    }
    \pgfplotsset{
        compat=newest,
        cycle list name=my colors,
        legend cell align=left,
        width=0.33\textwidth,
    }

\begin{tikzpicture}
    \begin{axis}[
        title={Ground Collection},
        xlabel={M1 classification rate [\%]},
        ylabel={M2 classification rate [\%]},
        ymin=0,ymax=60,
        minor y tick num=2,
        ymajorgrids=true,
        xmajorgrids=true,
        yminorgrids=true,
        minor grid style=loosely dotted,
        only marks,
        scatter,
        mark size=2.5pt,
        scatter src=explicit symbolic,
        table/meta=Method,
        scatter/classes={
            VGG16={mark=diamond*},
            VGG19={mark=halfdiamond*},
            Inception={mark=square*},
            ResNet={mark=triangle*},
            MobileNet={mark=star},
            NASNetMobile={mark=otimes}
        },
        % this legend only shows the defined scatter classes
        % (as it is stated in the manual)
        % legend entries={
        %     VGG16,
        %     VGG19,
        %     Inception,
        %     ResNet%
        % },
        % %legend pos= north west,
        % %legend pos=outer north east,
        % legend style={  at={(0.5,-0.3)},
        %                 anchor=north,legend columns=-1}
    ]

        % ---------------------------------------------------------------------
        % dummy plot for the legend
        % (make sure the expression values are outside the visible axis limits
        %  if this `\addplot wouldn't be present. This requires at least
        %  setting one of the limits explicitly, i.e. in this case `xmin')
        \addplot table [
            x expr=-10,
            y expr=-10,
        ] {figures/C2_PGround.dat};
        % ---------------------------------------------------------------------

        % simplified the call for the data
        \foreach \i in {
            Baseline,
            Honeywell,
            TexasAMPeking,
            Northwestern,
            TsingHua%
        }{
            \addplot table [
                x=\i-M1,
                y=\i-M2,
            ] {figures/C2_PGround.dat};
        }

    \end{axis}

    % % this is a dummy `axis' environment only to create the second legend
    % \begin{axis}[
    %     % set some axis limits and plot the coordinates outside that box
    %     % so they don't show up
    %     xmin=1,
    %     xmax=2,
    %     ymin=1,
    %     ymax=2,
    %     % of course we also don't want to show this axis
    %     hide axis,
    %     % we need only marks
    %     only marks,
    %     % state the legend entries for the second legend
    %     % (here we don't have scatter classes, so each `\addplot' gets its
    %     %  own entry in the legend)
    %     legend entries={
    %         ,       % the dummy plot should not show up in the legend
    %         Baseline,
    %         Honeywell,
    %         TexasAMPeking,
    %         Northwestern,
    %         TsingHua%
    %     },
    %     % place it below the other legend
    %     % therefore we have to shift it down (manually)
    %     %legend pos=outer north east,
    %     legend style={  at={(0.5,-0.2)},
    %                     anchor=north,
    %                     legend columns=3,
    %                     yshift=-20pt}
    %     %legend style={
    %      %   yshift=-60pt,
    %     %},
    % ]
    %     % just add some dummy plots to create the legend
    %     \foreach \i in {0,...,5} {
    %         \addplot+ [mark=*] coordinates { (0,0) };
    %     }
    % \end{axis}
\end{tikzpicture}
}
\vfill \vspace{0.5em}
\begin{tikzpicture}
    \begin{axis}[%
        hide axis,
        xmin=0,
        xmax=50,
        ymin=0,
        ymax=0.4,
        only marks,
        mark size=3.5pt,
        legend style={
            draw=white!15!black,
            legend cell align=left, 
            legend columns=6,
        }
    ]
    \addlegendimage{black!50,mark=diamond*}
    \addlegendentry{VGG16};
    \addlegendimage{black!50,mark=halfdiamond*}
    \addlegendentry{VGG19};
    \addlegendimage{black!50,mark=square*}
    \addlegendentry{Inception};
    \addlegendimage{black!50,mark=triangle*}
    \addlegendentry{ResNet};
    \addlegendimage{black!50,mark=star}
    \addlegendentry{MobileNet};
    \addlegendimage{black!50,mark=otimes}
    \addlegendentry{NASNetMobile};
    
    \addlegendimage{red,mark=*}
    \addlegendentry{Baseline};
    
    \addlegendimage{blue,mark=*}
    \addlegendentry{CCRE};
    
    \addlegendimage{green,mark=*}
    \addlegendentry{CDRM};
    
    \addlegendimage{magenta,mark=*}
    \addlegendentry{MA-CNN};
    
    \addlegendimage{yellow,mark=*}
    \addlegendentry{TM-DIP};

    \end{axis}
\end{tikzpicture}
\caption{Comparison of classification rates at rank 5 for each collection after applying the four algorithms submitted by teams for this task.\vspace{-2em}}
\label{fig:graph:C2Participants}
\end{figure*}
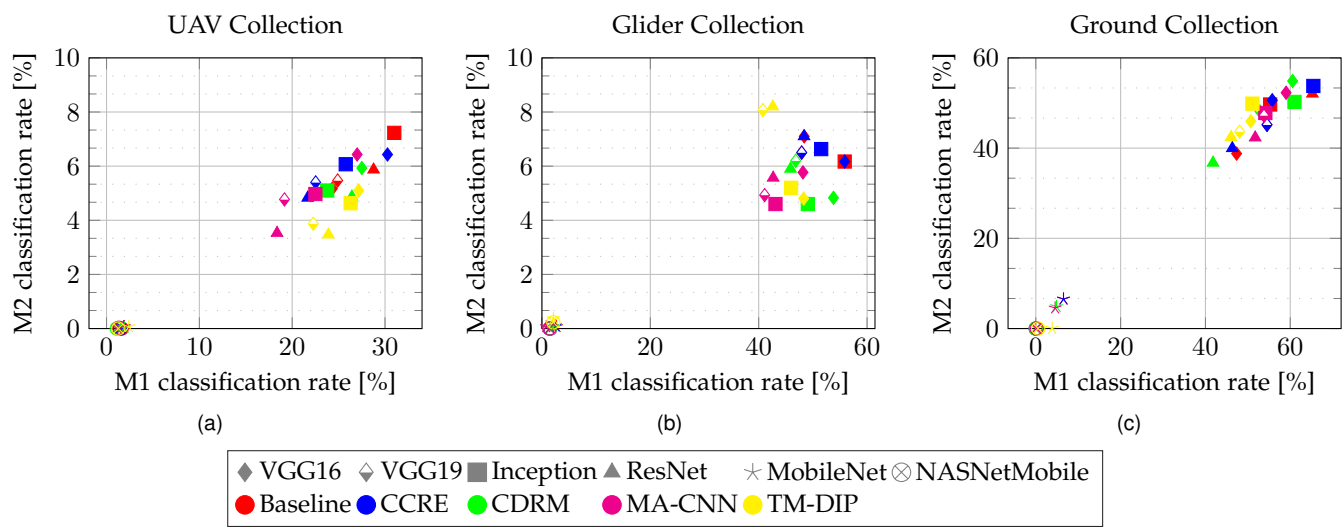

As can be observed in Supp. Table~3 and Fig.~\ref{fig:graph:C2Participants}, the four novel algorithms submitted by participants for the second UG$^2$ evaluation task excelled in the processing of certain collections while falling short in others. Most of the submitted algorithms were able to improve the classification performance of the images in the Ground Collection, but they struggled in improving the classification for the aerial collections, whose scenes tend to have a higher degree of variability than those present in the Ground Collection. Only the CCRE algorithm was able to improve the performance of one of the metrics for the UAV Collection (the M2 metric for the ResNet network, with a $0.20\%$ improvement over the baseline). The MA-CNN algorithm was able to improve two of the Glider Collection metrics (the M2 metric for the VGG16 and VGG19 networks with $2.23\%$ and $3.10\%$ improvement respectively over the baselines). For the most part, the algorithms tended to improve the metrics for the Ground Collection, with the highest classification improvement being provided by the CDRM method, with an improvement of $5.30\%$ and $5.21\%$ over the baselines for the Inception M1 and M2 metrics.

Further along these lines, while both metrics saw some improvement, the M2 metric benefited the most from these enhancement algorithms. This behavior is more pronounced when examined in the context of classic vs. state-of-the-art algorithms (Supp. Fig. 7). While the baseline enhancement algorithms had moderate improvements in both metrics, the participant algorithms seemed to favor the M2 metric over the M1 metric. For example, while the highest improvement (on the Ground Collection) for the VGG19 network in M1 was $0.03\%$, the improvement for the same network in  M2 was $2.95\%$. This leads us to conclude that the effect these algorithms have on automatic object recognition would be that of increasing within-class classification certainty. In other words, they would make an object belonging to a super-class \(L_i = \{s_1, ..., s_n\}\) become a better representation of such class features, such that the networks are able to detect more members of that class in their top 5 predictions.

% if have a single appendix:
%\appendix[Proof of the Zonklar Equations]
% or
%\appendix  % for no appendix heading
% do not use \section anymore after \appendix, only \section*
% is possibly needed

% use appendices with more than one appendix
% then use \section to start each appendix
% you must declare a \section before using any
% \subsection or using \label (\appendices by itself
% starts a section numbered zero.)
%

% \begin{figure}[t]
%     \centering
%     %left bottom right top
%     \includegraphics[clip, trim=4cm 4.3cm 4cm 4.3cm, width=0.45\textwidth]{figures/TAM.pdf}
%     \caption{The CDRM enhancement pipleline. If the glider set is detected, no action is taken (recognition is deemed to be good enough by default).}
%     \label{fig:TAM}
%     \vspace{-2em}
% \end{figure}

\section{Discussion}

The area of computational photography has a long-standing history of generating algorithms to improve image quality. However, the metrics those algorithms tend to optimize do not correspond to human perception. And as we found out, this makes them unreliable for the task of object recognition. 
%Given this, how do we design an enhancement \& %recognition pipeline that simultaneously corrects %imaging artifacts and recognizes objects of interests? %Skeptics may argue that re-training or fine-tuning %existing recognition networks can solve the problem. %As promising as this may sound, training a network %involves a large cost in terms of time and resources. %And if we were to train these networks each time for %a new task or dataset, we would be ignoring other %paths to solving this problem. 
The UG\textsuperscript{2} dataset paves the way for this new research through the introduction of a benchmark that is well matched to the problem area. %The challenge is to jointly optimize two seemingly %divergent but relevant tasks of (1) image enhancement %and restoration and (2) object recognition. 
The first iteration of the challenge workshop making use of this dataset saw the participation of teams from around the globe and introduced six unique algorithms to bridge the gap between computational photography and recognition, which we have described in this article. As noted by some participants and in accordance with our initial results, the problem is still not solved --- improving image quality using existing techniques does not necessarily solve the recognition problem. 

The results of our experiments led to some surprises. Even though the restoration and enhancement algorithms tended to improve the classification results for the diverse imagery included in our dataset, no approach was able to uniformly improve the results for all of the candidate networks. 
%%As a matter of fact, it has been noted by Palacio \etal~\cite{DBLP:journals/corr/abs-1803-08337} that, while classifiers tend to extract features based on a common portion of an input image, this common portion is not necessarily comprehensive, being as small as $10\%$ for networks such as ResNet50, and as large as $80\%$ for VGG networks. Thus, enhancement algorithms that were focused on improving visual features valuable for a network such as ResNet50 might have not have the same impact when the enhanced image is evaluated by a VGG network.
Moreover, in some cases, performance degraded after image processing, particularly for frames with higher amounts of image aberrations. This highlights the often single focus nature of image enhancement algorithms, which tend to specialize in removing a specific kind of artifact from an otherwise good quality image, which might not always be the present. Some of the algorithmic advancements (\textit{e.g.}, MA-CNN and CDRM) developed as a product of this challenge seek to address the problem by incorporating techniques such as deblurring, denoising, deblocking, and super-resolution into a single pre-processing pipeline. This practice pointed out the fact that while the individual implementation of some of these techniques might be detrimental to the visual quality or visual recognition task, when applied in conjunction with other enhancement techniques their effect turned out to be beneficial for both of these objectives. 

We also found out that image quality is a subjective assessment and better left to humans who are physiologically tuned to notice higher variations and artifacts in images as a result of evolution. Based on this observation, we developed a psychophysics-based evaluation regime for human assessment and a realistic set of quantitative measures for object recognition performance. The code for conducting such studies will be made publicly available following the publication of this article.

Inspired by the success of the UG$^2$ challenge workshop held at CVPR 2018, we intend to hold subsequent iterations with associated competitions based on the UG$^2$ dataset. These workshops will be similar in spirit to the PASCAL VOC and ImageNet workshops that have been held over the years and will feature new tasks, extending the reach of UG\textsuperscript{2} beyond the realm of image quality assessment and object classification.

% \appendices
% \section{Proof of the First Zonklar Equation}
% Appendix one text goes here.

% % you can choose not to have a title for an appendix
% % if you want by leaving the argument blank
% \section{}
% Appendix two text goes here.

% use section* for acknowledgment
\ifCLASSOPTIONcompsoc
  % The Computer Society usually uses the plural form
  \section*{Acknowledgments}
\else
  % regular IEEE prefers the singular form
  \section*{Acknowledgment}
\fi

Funding for this work was provided under IARPA contract \#2016-16070500002, and NSF DGE \#1313583. This workshop is supported in part by the Office of the Director of National Intelligence (ODNI), Intelligence Advanced Research Projects Activity (IARPA). The views and conclusions contained herein are those of the organizers and should not be interpreted as necessarily representing the official policies, either expressed or implied, of ODNI, IARPA, or the U.S. Government. The U.S. Government is authorized to reproduce and distribute reprints for governmental purposes notwithstanding any copyright annotation therein. 

Hardware support was generously provided by the NVIDIA Corporation, and made available by the National Science Foundation (NSF) through grant \#CNS-1629914. We thank Drs. Adam Czajka, and Christopher Boehnen for conducting an impartial judgment for the challenge tracks to determine the winners, Mr. Vivek Sharma for executing his code on our data and providing us with the result, Kelly Malecki for her tireless effort in annotating the test dataset for the UAV and Glider collections, and Sandipan Banerjee for assistance with data collection.   \vspace{-1em}

%NSF funding support: NSF DGE #1313583

% The authors would like to thank...

% Can use something like this to put references on a page
% by themselves when using endfloat and the captionsoff option.
\ifCLASSOPTIONcaptionsoff
  \newpage
\fi

% trigger a \newpage just before the given reference
% number - used to balance the columns on the last page
% adjust value as needed - may need to be readjusted if
% the document is modified later
%\IEEEtriggeratref{8}
% The "triggered" command can be changed if desired:
%\IEEEtriggercmd{\enlargethispage{-5in}}

% references section

% can use a bibliography generated by BibTeX as a .bbl file
% BibTeX documentation can be easily obtained at:
% http://mirror.ctan.org/biblio/bibtex/contrib/doc/
% The IEEEtran BibTeX style support page is at:
% http://www.michaelshell.org/tex/ieeetran/bibtex/
\bibliographystyle{IEEEtran}
% argument is your BibTeX string definitions and bibliography database(s)
\bibliography{main}
%
% <OR> manually copy in the resultant .bbl file
% set second argument of \begin to the number of references
% (used to reserve space for the reference number labels box)
% \begin{thebibliography}{1}

% \bibitem{IEEEhowto:kopka}
% H.~Kopka and P.~W. Daly, \emph{A Guide to \LaTeX}, 3rd~ed.\hskip 1em plus
%   0.5em minus 0.4em\relax Harlow, England: Addison-Wesley, 1999.

% \end{thebibliography}

% biography section
% 
% If you have an EPS/PDF photo (graphicx package needed) extra braces are
% needed around the contents of the optional argument to biography to prevent
% the LaTeX parser from getting confused when it sees the complicated
% \includegraphics command within an optional argument. (You could create
% your own custom macro containing the \includegraphics command to make things
% simpler here.)
%\begin{IEEEbiography}[{\includegraphics[width=0.9in,height=1.1in,clip,keepaspectratio]{mshell}}]{Michael Shell}
% or if you just want to reserve a space for a photo:
\vspace{-5em}
\begin{IEEEbiography}[{\includegraphics[width=0.9in,height=1.1in,clip,keepaspectratio]{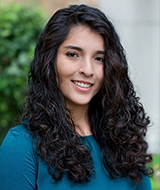}}]{Rosaura G. VidalMata} is a Ph.D. student in the Computer Science and Engineering Department at the University of Notre Dame. She received the B.S. degree in Computer Science at the Tecnologico de Monterrey (ITESM) in 2015, were she graduated with an Honorable Mention for Excellence. Her research interests include computer vision, machine learning and biometrics.
\end{IEEEbiography}
\vspace{-5em}
\begin{IEEEbiography}[{\includegraphics[width=0.9in,height=1.1in,clip,keepaspectratio]{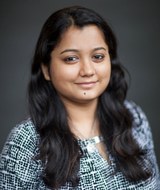}}]{Sreya Banerjee} is a Ph.D. student in the Computer Science and Engineering department at the University of Notre Dame. She received the B.S. degree in Information Technology from West Bengal University of Technology, Kolkata, India in 2010. Her research interests include computer vision, machine learning, computational neuroscience. 
\end{IEEEbiography}
\vspace{-5em}
\begin{IEEEbiography}[{\includegraphics[width=0.9in,height=1.1in,clip,keepaspectratio]{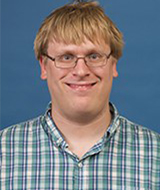}}]{Brandon RichardWebster} is a National Science Foundation (NSF) Graduate Research Fellow and Ph.D. student in the Department of Computer Science and Engineering at the University of Notre Dame. He received the B.S. degree in computer science from Bethel University, Minnesota, USA, in 2015. His research interests include computer vision, machine learning, and visual psychophysics.
\end{IEEEbiography}
% \begin{IEEEbiography}[{\includegraphics[width=0.9in,height=1.1in,clip,keepaspectratio]{authors/kmalecki.jpeg}}]{Kelly Malecki}
% Biography text here.
% \end{IEEEbiography}
\vspace{-5em}
\begin{IEEEbiography}[{\includegraphics[width=0.9in,height=1.1in,clip,keepaspectratio]{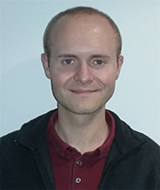}}]{Michael Albright} is a Data Scientist at Honeywell ACST, which he joined in 2015.  He received his B.S. degree in Physics from St. Cloud State University in 2009 and his Ph.D. in Theoretical Physics from the University of Minnesota in 2015.  His current research interests include computer vision, deep learning, and data science applied to the internet of things.
\end{IEEEbiography}
\vspace{-5em}
\begin{IEEEbiography}[{\includegraphics[width=0.9in,height=1.1in,clip,keepaspectratio]{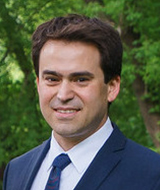}}]{Pedro Davalos} is a Principal Data Scientist at Honeywell ACST, joined in 2010. He obtained the B.S. degree in Computer Engineering, and the M.S. degree in Computer Science, both from Texas A\&M University. His research interests include machine learning and computer vision.
\end{IEEEbiography}
\vspace{-5em}
\begin{IEEEbiography}[{\includegraphics[width=0.9in,height=1.1in,clip,keepaspectratio]{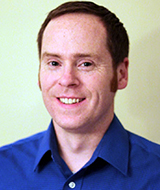}}]{Scott McCloskey} is a Technical Fellow at Honeywell ACST, which he joined in 2007.  He obtained his Ph.D. from McGill University, a MS from the Rochester Institute of Technology, and a BS from the University of Wisconsin – Madison.  His research interests include computer vision, computational photography, and industrial applications of imaging and recognition
\end{IEEEbiography}
\vspace{-5em}
\begin{IEEEbiography}[{\includegraphics[width=0.9in,height=1.1in,clip,keepaspectratio]{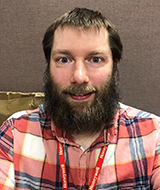}}]{Ben Miller} is a Software Engineer at Honeywell ACST, he joined in 2005.  He received his MS in Computer Science from the University of Minnesota in 2006. His research interests include cybersecurity, distributed computing, machine learning frameworks and applications in computer vision.
\end{IEEEbiography}
\vspace{-4.5em}
\begin{IEEEbiography}[{\includegraphics[width=0.9in,height=1.1in,clip,keepaspectratio]{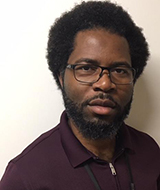}}]{Asongu Tambo}  is a Research Scientist as Honeywell ACST, joined in 2017. He obtained his Ph.D. in Electrical Engineering from the University of California at Riverside in 2016.  Dr. Tambo's research interests include image processing, tracking, learning, and the application of computer vision techniques to new frontiers such as Biology.
\end{IEEEbiography}
\vspace{-5em}
\begin{IEEEbiography}[{\includegraphics[width=0.9in,height=1.1in,clip,keepaspectratio]{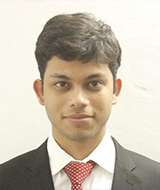}}]{Sushobhan Ghosh} is a Ph.D. student in the Computer Science Department working in the Computational Photography Lab at Northwestern University, Evanston. He completed his undergrad at Indian Institute of Technology Delhi (IIT Delhi), India in 2016. Sushobhan is interested in deep learning/machine learning applications in computer vision and computational photography problems. 
\end{IEEEbiography}
\vspace{-5em}
\begin{IEEEbiography}[{\includegraphics[width=0.9in,height=1.1in,clip,keepaspectratio]{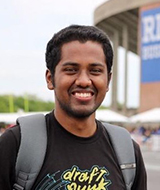}}]{Sudarshan Nagesh} is a Research Engineer working at Zendar Co., Berkeley. He completed his undergraduation in electrical engineering at National Institute of Technology Karnataka (NITK Surathkal), India in 2012. He received an MSc. Engg degree from the electrical engineering department at the Indian Institute of Science, Bangalore in 2015. In December 2017 he completed his MS in computational imaging from Rice University. 
\end{IEEEbiography}
\vspace{-5em}
\begin{IEEEbiography}[{\includegraphics[width=0.9in,height=1.1in,clip,keepaspectratio]{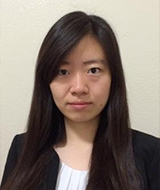}}]{Ye Yuan} is a Ph.D. student in Computer Science at the Texas A\&M University. She obtained the B.S. degree at the University of Science and Technology of China (USTC) in 2015. During 2014-2015, she was a visiting student intern and completed her undergraduate thesis at Rice University. Ye Yuan focuses her research on machine learning, deep learning, computer vision techniques and applications especially in person re-identification and healthcare.
\end{IEEEbiography}
\vspace{-5.0em}
\begin{IEEEbiography}[{\includegraphics[width=0.9in,height=1.1in,clip,keepaspectratio]{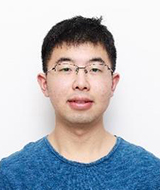}}]{Yueyu Hu} received the B.S. degree in computer science from Peking University, Beijing, China, in 2018, where he is currently pursuing the master's degree with the Institute of Computer Science and Technology. His current research interests include video and image compression, understanding and machine intelligence.
\end{IEEEbiography}
\vspace{-5em}
\begin{IEEEbiography}[{\includegraphics[width=0.9in,height=1.1in,clip,keepaspectratio]{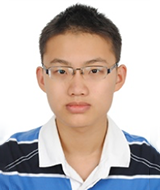}}]{Junru Wu} received the BE degree in Electrical Engineering from Tongji University in 2016. From 2016 to 2017, he was a research assistant at ShanghaiTech University. He is currently a Ph.D. student at the Visual Informatics Group at Texas A\&M University.His research interests lie in the broad areas of deep learning and computer vision, specifically including image restoration, saliency detection, and deep network compression. 
\end{IEEEbiography}
\vspace{-5em}
\begin{IEEEbiography}[{\includegraphics[width=0.9in,height=1.1in,clip,keepaspectratio]{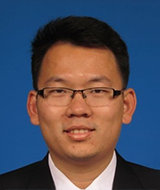}}]{Wenhan Yang} is a postdoctoral researcher at Learning and Vision Lab, Department of Electrical and Computer Engineering, National University of Singapore. He received the Ph.D. degree (Hons.) in computer science from Peking University, Beijing, China 2018.  His current research interests include image/video processing, compression, and computer vision. 
\end{IEEEbiography}
\vspace{-5.0em}
\begin{IEEEbiography}[{\includegraphics[width=0.9in,height=1.1in,clip,keepaspectratio]{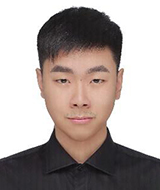}}]{Xiaoshuai Zhang} is an undergraduate student at Peking University, where he is majoring in Machine Intelligence and Computer Science. He is currently working as research intern at the Institute of Computer Science and Technology, Peking University. His current research interests include image processing, computer vision, machine learning and security.
\end{IEEEbiography}
\vspace{-5em}
\begin{IEEEbiography}[{\includegraphics[width=0.9in,height=1.1in,clip,keepaspectratio]{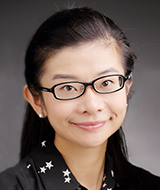}}]{Jiaying Liu} is currently an Associate Professor with the Institute of Computer Science and Technology, Peking University. She received the Ph.D. degree (Hons.) in computer science from Peking University, Beijing, China 2010. She has authored over 100 technical articles in refereed journals and proceedings and holds 28 patents. Her  research interests include image/video processing, compression, and computer vision.
\end{IEEEbiography}
\vspace{-5em}
\begin{IEEEbiography}[{\includegraphics[width=0.9in,height=1.1in,clip,keepaspectratio]{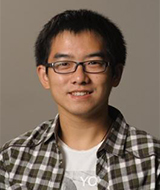}}]{Zhangyang Wang} is an Assistant Professor of Computer Science and Engineering, at the Texas A\&M University. During 2012-2016, he was a Ph.D. student in the Electrical and Computer Engineering  Department, at the University of Illinois at Urbana-Champaign. Prior to that, he obtained the B.E. degree at the University of Science and Technology of China (USTC), in 2012. 
\end{IEEEbiography}
\vspace{-5em}
\begin{IEEEbiography}[{\includegraphics[width=0.9in,height=1.1in,clip,keepaspectratio]{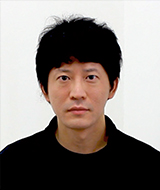}}]{Hwann-Tzong Chen} joined the Department of Computer Science, National Tsing Hua University, Taiwan, as a faculty member in 2006. He is currently an associate professor at the department. He received his PhD degree in computer science from National Taiwan University. Dr. Chen’s research interests include computer vision, image processing, and pattern recognition.
\end{IEEEbiography}
\vspace{-5em}
\begin{IEEEbiography}[{\includegraphics[width=0.9in,height=1.1in,clip,keepaspectratio]{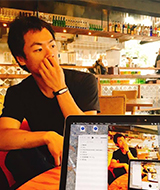}}]{Tzu-Wei Huang} is a PhD student and an adjunct lecturer in National Tsing-Hua University. He loves open source projects and contributes code to the community. His research interests include computer vision and deep learning.
\end{IEEEbiography}
\vspace{-5em}
\begin{IEEEbiography}[{\includegraphics[width=0.9in,height=1.1in,clip,keepaspectratio]{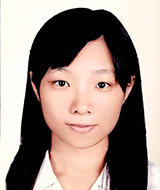}}]{Wen-Chi Chin} is a graduate student at National Tsing Hua University in Taiwan, she majored in Computer Science. Her focus is on the field of computer vision, such as video stabilization, SLAM and vehicle localization, and unsupervised depth and ego-motion estimation. Also, she participated in the project of Industrial Technology Research Institute. They developed the Scrabble-Playing Robot with Resnet-34 and won the CES 2018 Innovation Awards.
\end{IEEEbiography}
\vspace{-5em}
\begin{IEEEbiography}[{\includegraphics[width=0.9in,height=1.1in,clip,keepaspectratio]{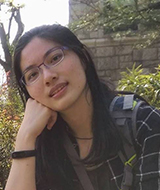}}]{Yi-Chun Li} is a MS student in National Tsing Hua University. Her research interests include computer vision, deep learning and deep reinforcement learning. Her focus is on topics such as video stabilization, object detection, and planning network with deep reinforcement learning for PCB auto-routing. 
\end{IEEEbiography}
\vspace{-5em}
\begin{IEEEbiography}[{\includegraphics[width=0.9in,height=1.1in,clip,keepaspectratio]{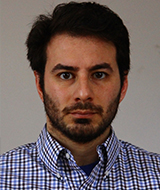}}]{Mahmoud Lababidi} (PhD Physics George Mason University, BS Computer Engineering University of Florida) has studied Topological Insulators (such as Graphene) along with their interplay with superconductors during his PhD. He navigated to Machine Learning by way of the Remote Sensing world by applying and modifying neural network based computer vision techniques on satellite based images. 
\end{IEEEbiography}
\vspace{-5em}
\begin{IEEEbiography}[{\includegraphics[width=0.9in,height=1.1in,clip,keepaspectratio]{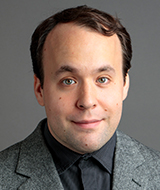}}]{Charles Otto} received the B.S. degree in Computer Science at Michigan State University in 2008, and the Ph.D. degree in Computer Science at Michigan State University in 2016. He currently works at Noblis, in Reston Virginia. His research interests include face recognition, computer vision, and pattern recognition.
\end{IEEEbiography}
\vspace{-5em}
\begin{IEEEbiography}[{\includegraphics[width=0.9in,height=1.1in,clip,keepaspectratio]{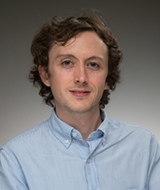}}]{Walter J. Scheirer} received the M.S. degree in computer science from Lehigh University, in 2006, and the Ph.D. degree in engineering from the University of Colorado, Colorado Springs, CO, USA, in 2009. He is an Assistant Professor with the Department of Computer Science and Engineering, University of Notre Dame. Prior to joining the University of Notre Dame, he was a Postdoctoral Fellow with Harvard University from 2012 to 2015, and the Director of Research and Development with Securics, Inc., from 2007 to 2012. His research interests include computer vision, machine learning, biometrics, and digital humanities. 
\end{IEEEbiography}

% insert where needed to balance the two columns on the last page with
% biographies
%\newpage
% if you will not have a photo at all:
% \begin{IEEEbiographynophoto}{Jane Doe}
% Biography text here.
% \end{IEEEbiographynophoto}

% You can push biographies down or up by placing
% a \vfill before or after them. The appropriate
% use of \vfill depends on what kind of text is
% on the last page and whether or not the columns
% are being equalized.

%\vfill

% Can be used to pull up biographies so that the bottom of the last one
% is flush with the other column.
%\enlargethispage{-5in}

% that's all folks
\end{document}

\title{Bridging the Gap Between Computational Photography and Visual Recognition: Supplemental Material \vspace{-15mm}}

\markboth{IEEE TRANSACTIONS ON PATTERN ANALYSIS AND MACHINE INTELLIGENCE,~Vol.~X, No.~X, February~2020}%
{Vidal \MakeLowercase{\textit{\etal }}: Bridging the Gap Between Computational Photography and Visual Recognition}

% use for special paper notices
%\IEEEspecialpapernotice{(Invited Paper)}

% for Computer Society papers, we must declare the abstract and index terms
% PRIOR to the title within the \IEEEtitleabstractindextext IEEEtran
% command as these need to go into the title area created by \maketitle.
% As a general rule, do not put math, special symbols or citations
% in the abstract or keywords.

% make the title area
\maketitle

% For peer review papers, you can put extra information on the cover
% page as needed:
% \ifCLASSOPTIONpeerreview
% \begin{center} \bfseries EDICS Category: 3-BBND \end{center}
% \fi
%
% For peerreview papers, this IEEEtran command inserts a page break and
% creates the second title. It will be ignored for other modes.
\IEEEpeerreviewmaketitle

\IEEEraisesectionheading{\section{Image Acquisition Pipeline}}
% There should be a subsection that reviews imaging mechanism. From the perspective of imaging mechanism, we can know the reason of low quality image and develop algorithms based on these mechanisms. However, this paper simply presents several deep learning algorithms and lacks discussions with imaging mechanism. We expect to see how to improve the high-level tasks by taking the imaging mechanism into account.

%%The sources of corruption for the input image can be environment (weather, blooming, glare, occlusion, time of the day, scale), sensor (fish eye, over/under exposure, noise), imaging algorithm within sensor (compression artifacts for saving digital image).
From the selection of a scene in the world to capture with a camera to the formation of the digital image itself, the sources of corruption in digital images or videos are diverse. In this section, we review the image acquisition pipeline and describe these sources of problems in order to facilitate development of algorithms that specifically target them.

\textbf{Environment:} When capturing a scene, the first source of corruption that comes to mind is the environment. Weather, occlusion (by objects, such as people, suddenly appearing in the scene disrupting the view), and time of day change the scene periodically and introduce artifacts naturally within the scene. For example, in our previous work describing the UG$^2$ Challenge dataset~\cite{ug2}, we describe in detail the effect of weather on  digital imagery, specifically in reference to the Ground collection. Without any pre-processing applied to the dataset, the  classification performance varies widely across different weather conditions --- objects on a cloudy day are easier to recognize than on a sunny, snowy or rainy day respectively. Our observations on UG$^2$~\cite{ug2} directly follow from previous insights on the impact of weather on face recognition in outdoor acquisition settings~\cite{Boult_2009_HRB}. Similarly, objects in images taken during the day are easier to recognize than those taken at night due to the availability of visible features under sufficient light levels and quality. Too much light, however, can cause lens glare (\textit{i.e.}, a naturally induced artifact due to light bouncing back from the camera lens) and corrupt the image by partially saturating a portion of the image, thereby losing important features from that region.

\begin{figure}[!t]
\centering
\subfloat[UAV Collection]{
    \includegraphics[width=0.20\textwidth]{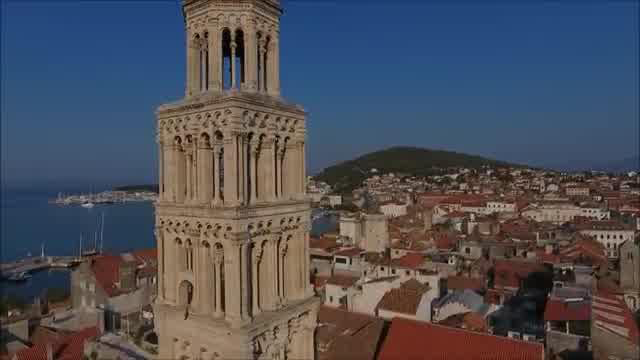}
    \includegraphics[width=0.20\textwidth]{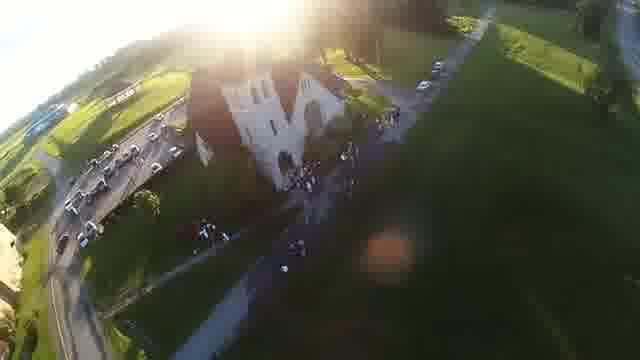}
} \vspace{-0.5em}
\hfil
\subfloat[Glider Collection]
{
    \includegraphics[width=0.20\textwidth]{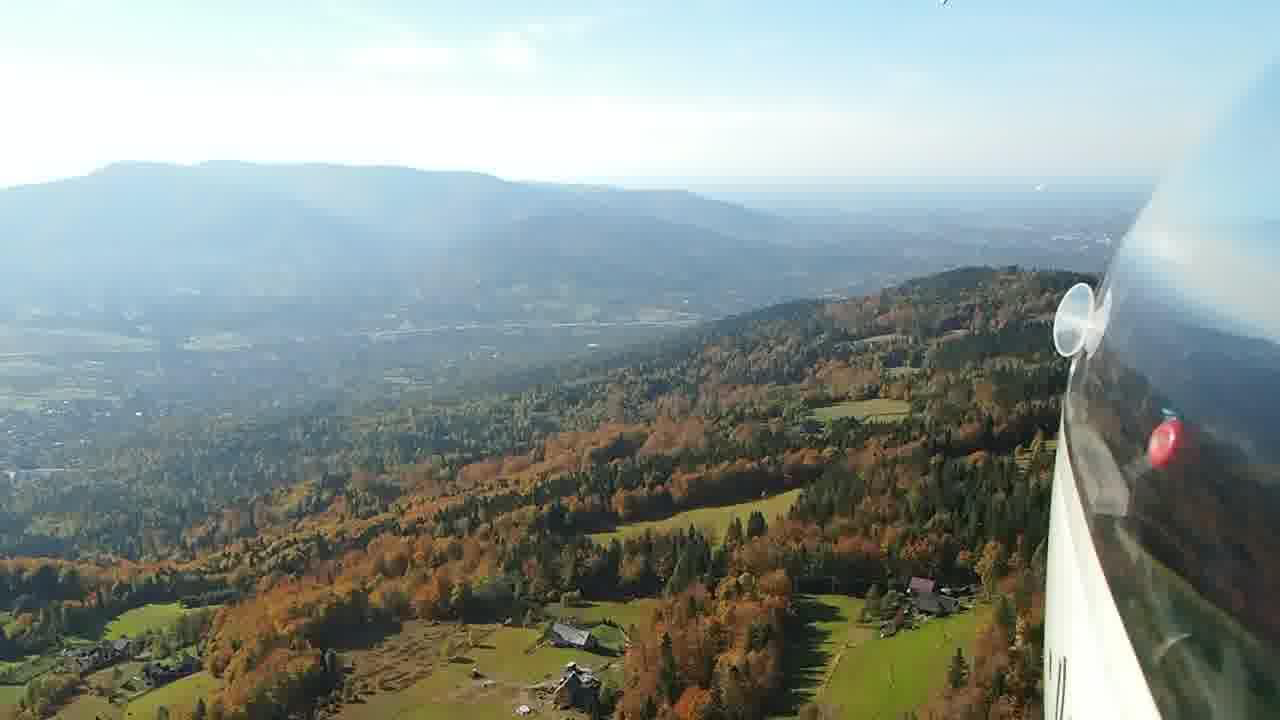}
    \includegraphics[width=0.20\textwidth]{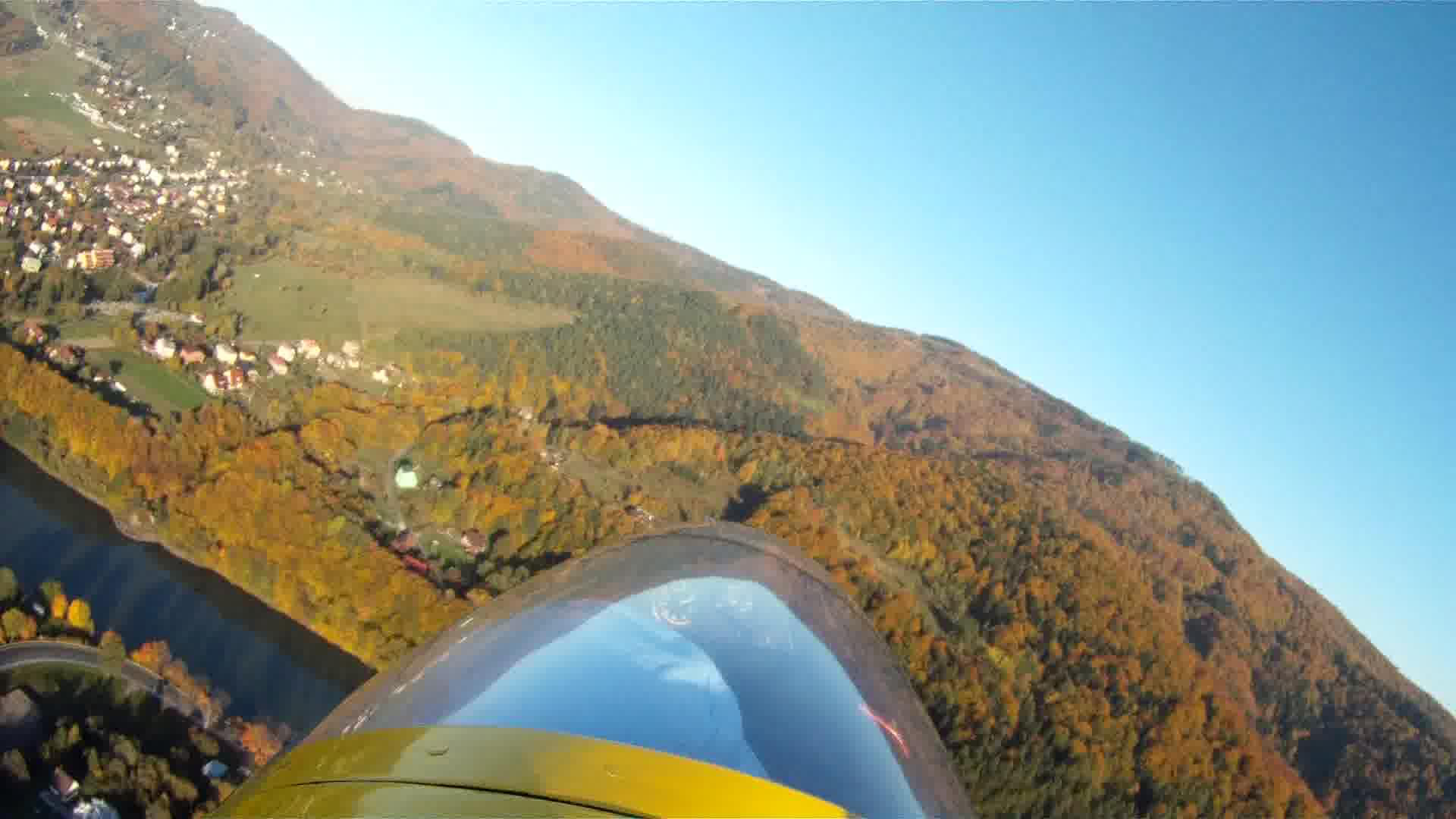}
} \vspace{-0.5em}
\hfil
\subfloat[Ground Collection]
{
    \includegraphics[width=0.20\textwidth]{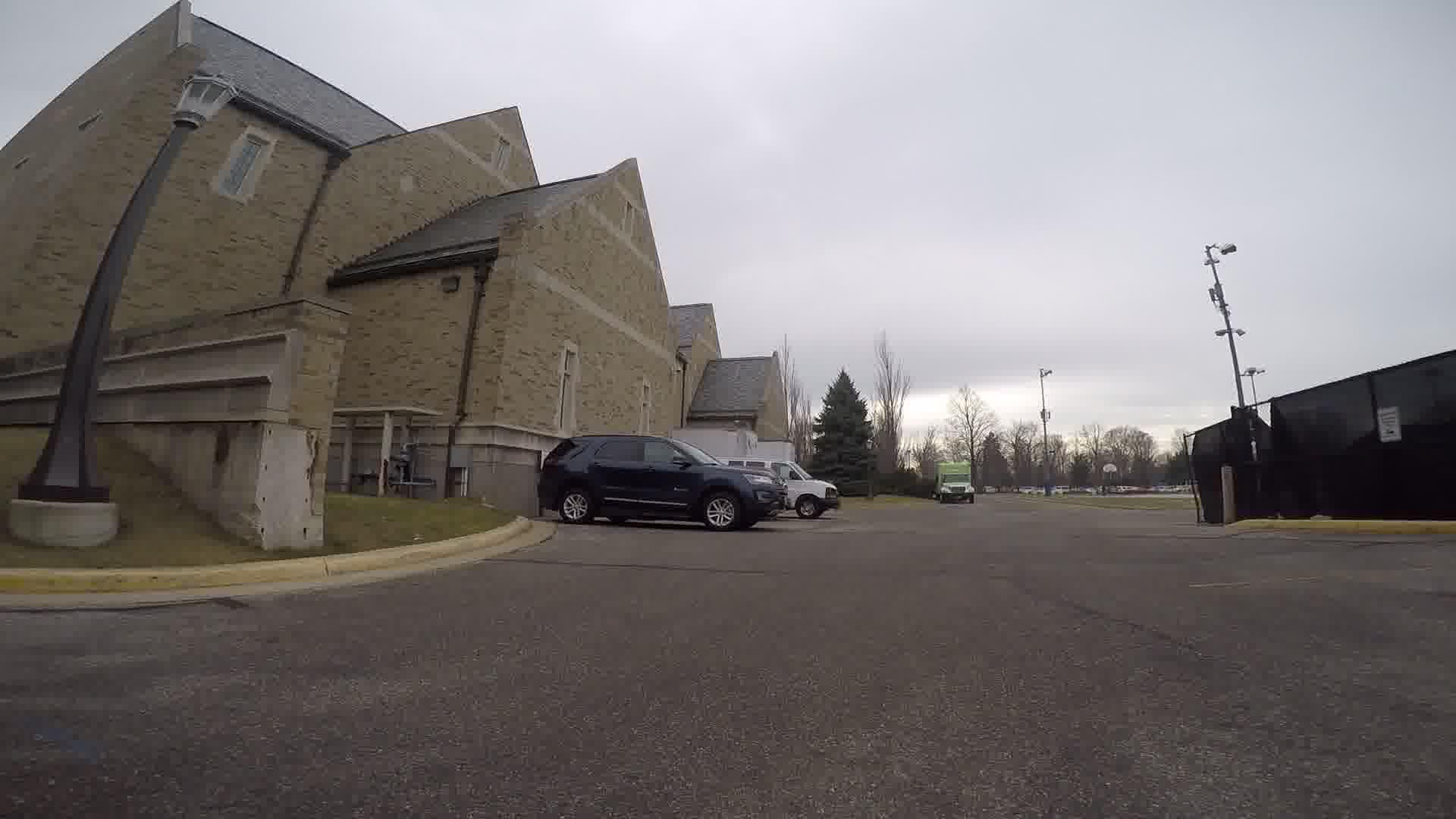}
    \includegraphics[width=0.20\textwidth]{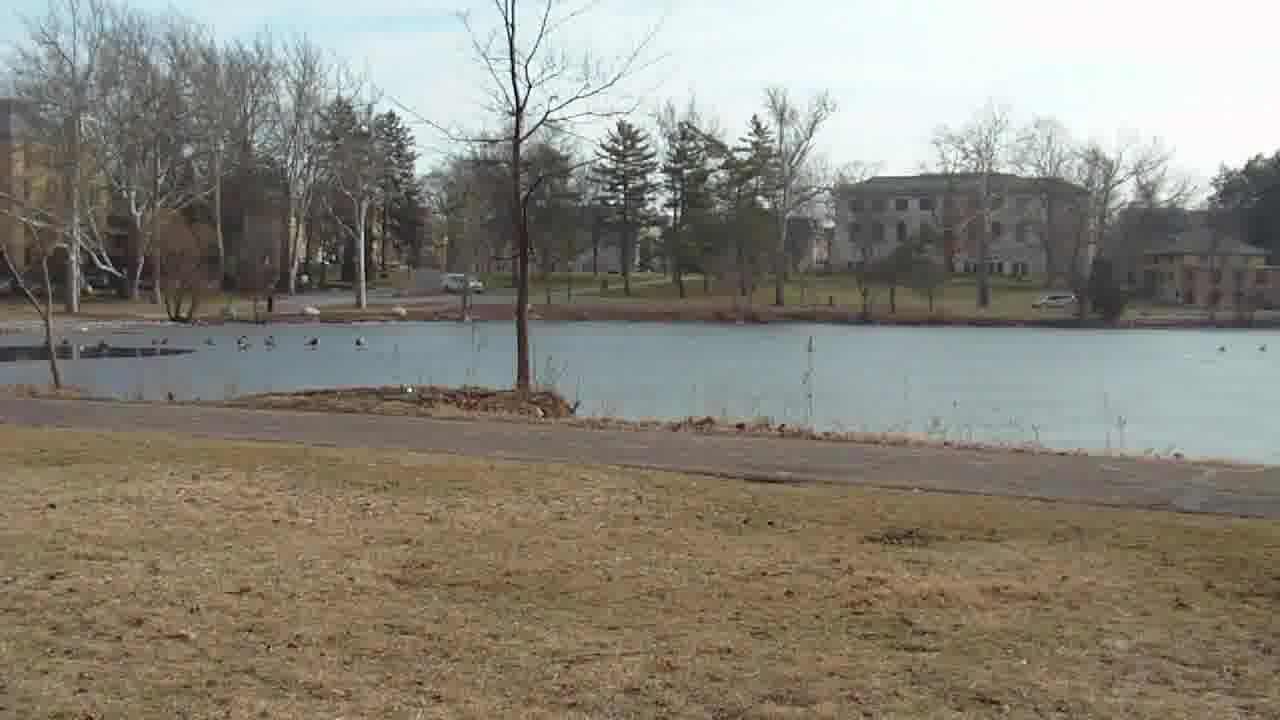}
}
\caption{Examples of images in the three UG$^2$ collections.}
\label{fig:datasets-samples}
\end{figure}

\textbf{Camera:} When capturing media using digital cameras, the light reflected from one or more surfaces passes through the camera's lenses and is picked up by the active sensing area in the imaging sensor. 
The type of lens the digital camera uses can affect the quality of the image tremendously. Camera lenses can be broadly classified into the following categories: a standard lens with mid-range focal length and roughly the same angle of view as the photographer to produce natural looking images, a macro lens designed for close-up photography, a telephoto lens with long focal length and high-level magnification allowing for photos of objects taken at a great distance, and a wide angle lens with a short focal length and an angle of view beyond that of a standard lens for the capture of a scene in its entirety. Wide angle lenses are useful for photographing landscapes, cramped interiors, and other objects that do not fit into a normal lens's field of view. A special category of ultra-wide angle lens is the fisheye lens, which when used creates a distorted, spherical view of the world, most evident in the curved outer corners of the photo, known as the ``fisheye effect''. Some examples of the fisheye effect can be found in the Ground Collection of UG$^2$ (see the left-hand image in Fig.~\ref{fig:datasets-samples}.c below).

The imaging sensor can either be a charge-coupled device (CCD) or a complementary metal oxide on silicon (CMOS). While CCD sensors (used in digital single-lens reflex cameras) usually outperform CMOS sensors in quality, CMOS tends to be better in low-power applications and, as such, those sensors are used in most digital cameras. 

One of the main differences between both sensor types is in the way they capture each frame, using a global (CCD) or rolling (CMOS) shutter. With a global shutter the entire frame is captured at once, while a rolling shutter captures a frame by quickly scanning though the scene either vertically or horizontally. In a static, well-lit environment the difference between both would be unnoticeable. However, with flickering illumination or fast moving objects, the difference becomes more apparent. With flickering lights a rolling shutter will generate an image in which the frame appears to be split (partial exposure), each section displaying different levels of illumination. With a moving object (or during a fast-moving panning shot) the CMOS sensor will seem to skew the image (as the camera or object moves, it captures different parts of the frame at different times making the image ``bend'' in one direction) or introduce a more severe wobbling effect. On the other hand, a global shutter will result in a blurry image (a product of its swish-pan operation). Algorithms, such as the CCRE method introduced in this paper, can take into account this effect (which is prevalent in high-speed scenes captured by aerial vehicles) and use it to improve the overall quality of the observed scenes. 

In addition to this, the performance of a digital imaging sensor can be affected by a variety of factors, such as  sensor noise (fixed-pattern noise, amplifier noise, dark current noise, etc.), shutter speed (the exposure time controls the amount of light and motion blur in the resulting picture), analog gain (the signal is usually boosted by a sense amplifier that is controlled by the automatic gain control logic in the camera, which serves as an automated exposure control, however some settings can amplify the sensor noise), sampling pitch (a smaller sampling pitch can provide a higher resolution but would mean that the sensor does not have enough area to be able to accumulate as many photons, resulting in poor light sensitivity and a higher likelihood of noise), and chip size (having a larger chip size is ideal, since this allows for a better photo-sensitivity, however this would increase the size --- and weight --- of the camera, which might not be possible for cameras located on small UAVs).

\textbf{Signal Processing and Transmission.} Once the observed light has arrived at the sensor and has been converted to bits, cameras tend to perform their own pre-processing of the captured signal to enhance the image before compressing and storing it. A problem one runs into while transforming the natural analog signal to digitized format is over/under-exposure, which is characterised by having a portion of the image completely saturated or darkened due to abundance or absence of light and the limited dynamic range of some cameras. Most cameras use 8 to 32 bits for representing the brightness of a single color channel. To avoid this problem, most cameras nowadays have built-in high dynamic range (HDR) processing. Other processing operations include color balancing. This operation moves the brightest point in the image closer to pure white, which might lead to overly significant compensation when said point has a high value in one of the RGB channels.

\section{Annotation format}
\begin{figure}[h]
    \centering
    \includegraphics[width=0.75\textwidth]{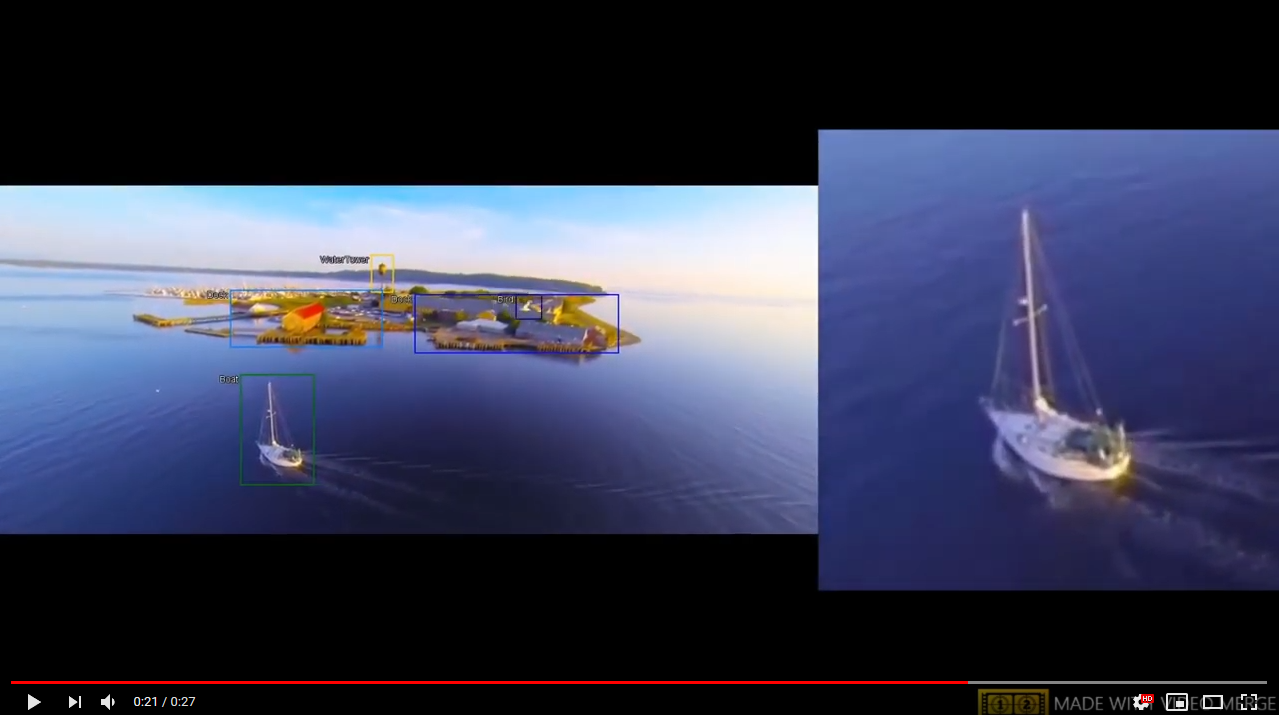}
    \caption{Screenshot of an annotated video from the UAV Collection (AV000119). The full video clip can be watched online: \url{https://youtu.be/BA52d7X8ihw}}
    \label{fig:videoclip}
\end{figure}

Bounding boxes establishing object regions were manually annotated using the VATIC Video Annotation Tool~\cite{Vatic:2013}. we provide the VATIC annotation files for every annotated video in the dataset. As shown in Fig.~\ref{fig:videoclip}, a single frame might contain multiple annotated objects of different shapes and sizes.

\begin{figure}[h]
\centering
    \includegraphics[width=0.55\textwidth]{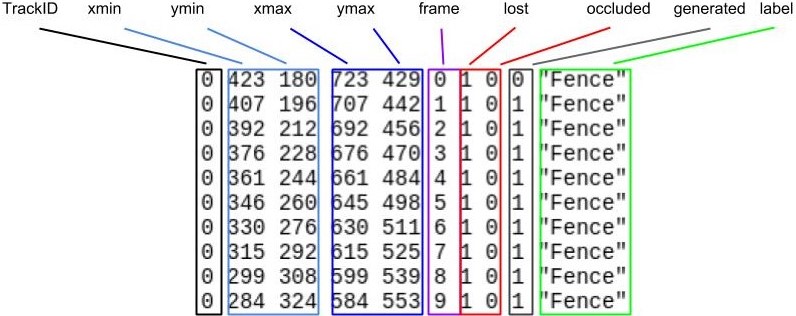}
    \caption{Example annotation file from the UG$^2$ dataset.}
    \label{fig:annotationfile}
\end{figure}

Each annotation file (see Fig.~\ref{fig:annotationfile}) follows the annotation structure provided by VATIC. Each line contains one object annotation, which is defined by 10 columns. The definition of each column is as follows:

\begin{itemize}
    \item \textbf{Track ID}. All rows with the same ID belong to the same path of the same object through different video frames.
    \item \textbf{xmin}. The top left x-coordinate of the bounding box.
    \item \textbf{ymin}. The top left y-coordinate of the bounding box. 
    \item \textbf{xmax}. The bottom right x-coordinate of the bounding box.
    \item \textbf{ymax}. The bottom right y-coordinate of the bounding box.
    \item \textbf{frame}. The frame that this annotation represents.
    \item \textbf{lost}. If 1, the annotation is outside of the view screen. In this case we did not extract any cropped region.
    \item \textbf{occluded}. If 1, the annotation is occluded. In this case we did not extract any cropped region.
    \item \textbf{generated}. If 1, the annotation was automatically interpolated. 
    \item \textbf{label}. The class for this annotation, enclosed in quotation marks.
\end{itemize}

\bibliographystyle{IEEEtran}
% argument is your BibTeX string definitions and bibliography database(s)
\bibliography{main}

\newpage

\begin{table}[h]
\centering
\begin{center}
\begin{tabular}{|c|c|c|}
\hline
\textbf{Filters} & \textbf{Kernel Shape} & \textbf{Activation} \\
\hline\hline

128 & 5x5 & LeakyRelu \\
128 & 1x1 & LeakyRelu \\
128 & 1x1 & LeakyRelu \\
128 & 1x1 & LeakyRelu \\
4 & 1x1 & Relu \\

\hline
\end{tabular}
\end{center}
\caption{The SSR network layers.}
\label{tab:HopkinsNetwork}
\vspace{-50mm}
\end{table}

\begin{figure}
    \centering
    \includegraphics[width=0.26\textwidth]{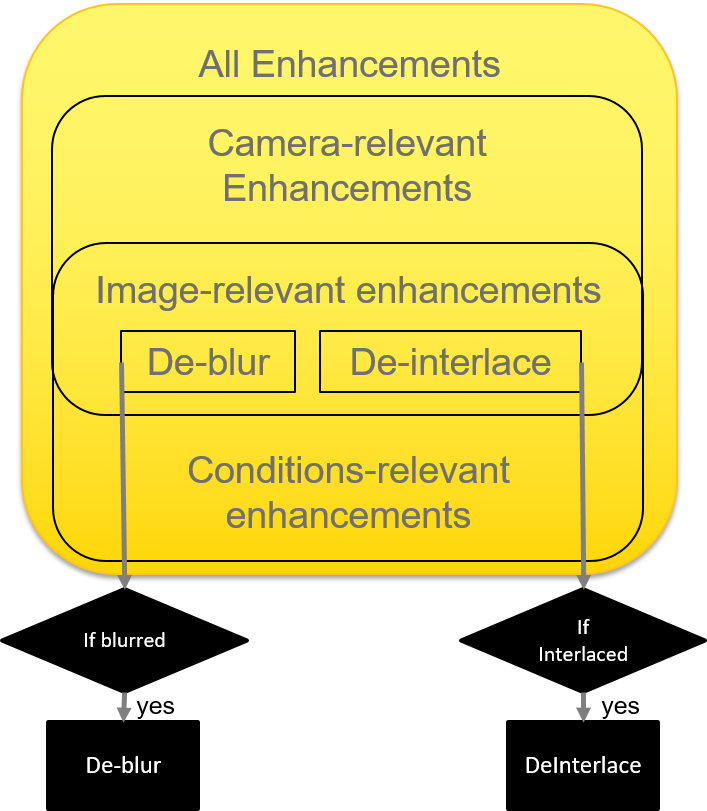}
    \caption{The CCRE approach conditionally selects enhancements.}
    \label{fig:HONsystem}
\end{figure}

\newpage

\begin{figure}
    \centering
    \includegraphics[width=0.30\textwidth]{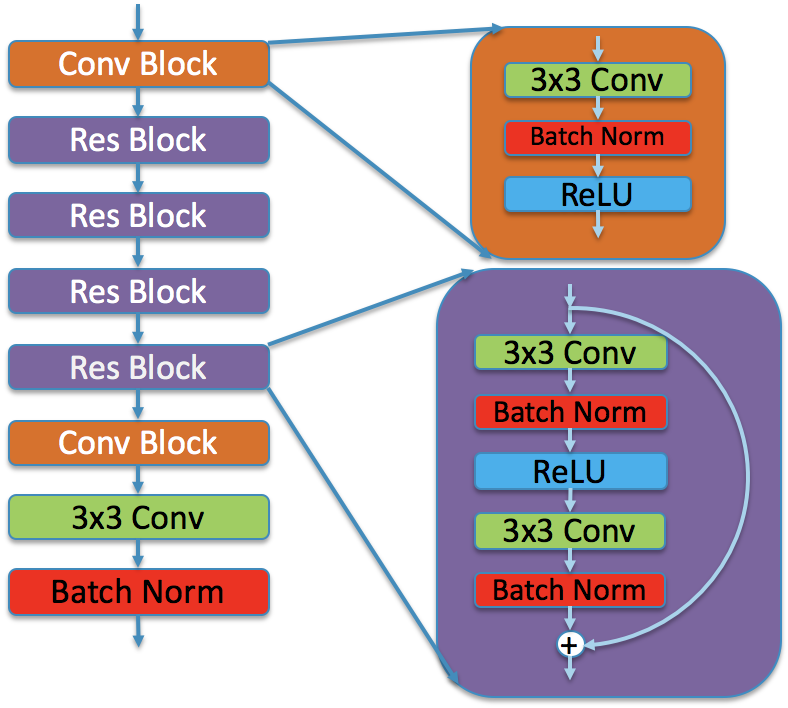}
    \caption{Network architecture used for MA-CNN image enhancement.}
    \label{fig:NorthwesternNet}
\end{figure}

\begin{figure}[t]
    \centering
    %left bottom right top
    \includegraphics[clip, trim=4cm 4.3cm 4cm 4.3cm, width=0.45\textwidth]{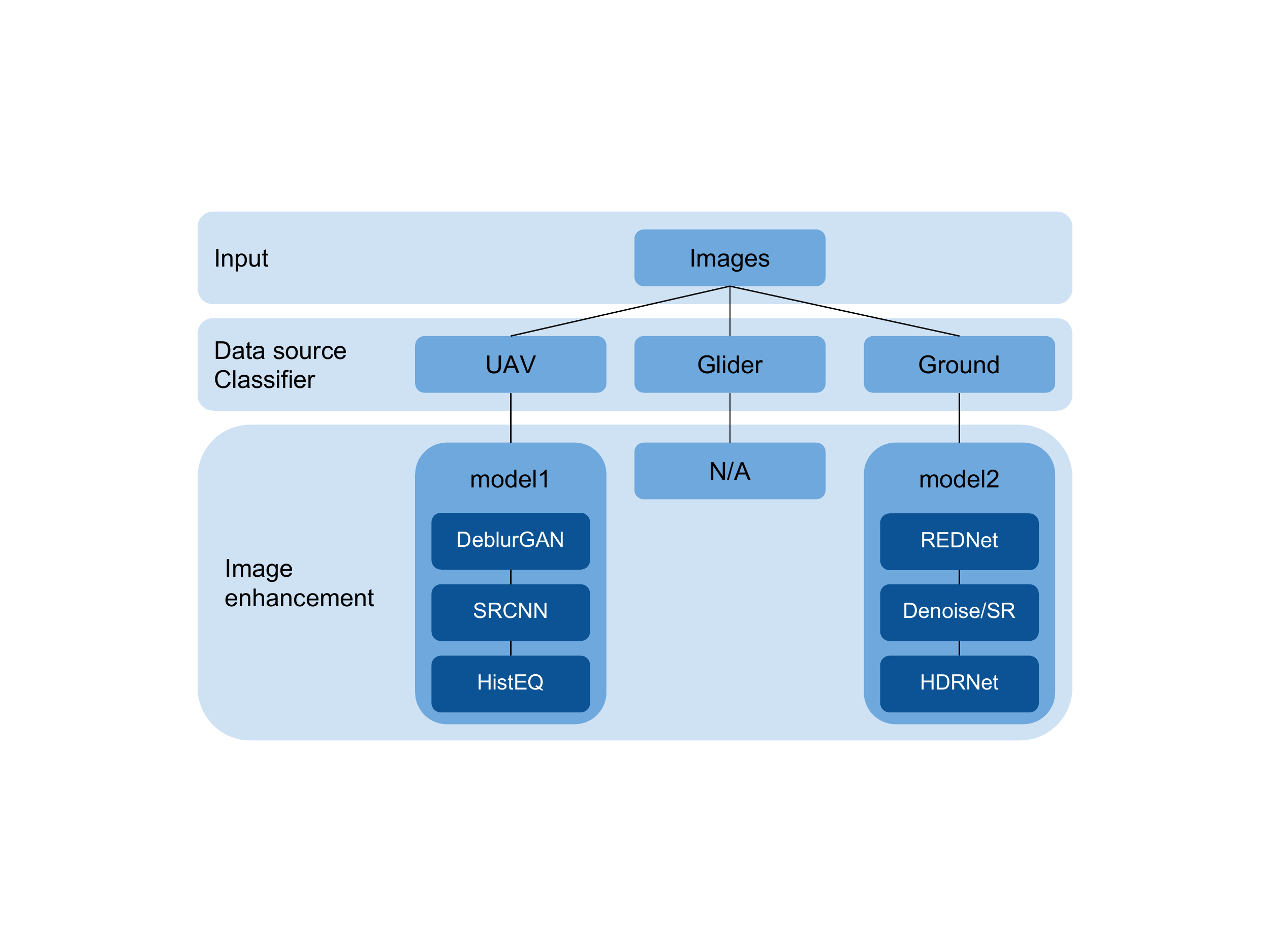}
    \caption{The CDRM enhancement pipleline. If the glider set is detected, no action is taken (recognition is deemed to be good enough by default).}
    \label{fig:TAM}
    \vspace{-2em}
\end{figure}

\begin{table}
\centering
%testing scores
\begin{center}
\begin{tabular}{|c|c|c|c|}
\hline
\textbf{Network} & \textbf{UAV} & \textbf{Glider}  & \textbf{Ground} \\
\hline\hline

VGG16-M1 & 24.33\% & 48.40\% & 47.35\% \\
VGG16-M2 & 5.17\% & 7.10\% & 38.78\% \\
VGG19-M1 & 24.90\% & 47.93\% & 54.50\% \\
VGG19-M2 & 5.47\% & 6.50\% & 45.21\% \\
Inception-M1 & 31.00\% & 55.87\% & 55.26\% \\
Inception-M2 & 7.23\% & 6.17\% & 49.67\% \\
ResNet-M1 & 28.77\% & 51.33\% & 65.23\% \\
ResNet-M2 & 5.87\% & 6.63\% & 52.06\% \\
MobileNetV2-M1 & 1.9\% & 2.7\% & 6.53\% \\
MobileNetV2-M2 & 0.07\% & 0.07\% & 6.53\% \\
NASNetMobile-M1 & 1.63\% & 1.53\% & 0\% \\
NASNetMobile-M2 & 0\% & 0\% & 0\% \\

\hline
\end{tabular}
\end{center}
\caption{Classification scores for the un-altered images of the testing dataset. The evaluated algorithms were expected to improve these scores.}
\label{tab:testing_dataset_basescores}
\vspace{-6mm}
\end{table}

\begin{table}
\centering
\begin{center}
\begin{tabular}{|c|c|c|c|}
\hline
\textbf{Metric} & \textbf{UAV} & \textbf{Glider}  & \textbf{Ground} \\
\hline\hline

VGG16-M1 & CDRM (26.43\%) & CCRE (\textcolor{red}{48.37\%})  & CCRE (\textcolor{red}{46.35\%}) \\ \hline
VGG16-M2 & CDRM (\textcolor{red}{4.87\%}) & MA-CNN (9.33\%) & TM-DIP (42.42\%) \\ \hline
VGG19-M1 & CDRM (26.40\%) & CCRE (\textcolor{red}{47.93\%}) & CCRE (54.53\%) \\ \hline
VGG19-M2 & CCRE (\textcolor{red}{5.40\%}) & MA-CNN (9.60\%) & CDRM (48.16\%) \\ \hline
Inception-M1 & CCRE (\textcolor{red}{30.27\%}) & CCRE (\textcolor{red}{55.87\%}) & CDRM (60.56\%) \\ \hline
Inception-M2 & CCRE (\textcolor{red}{6.43\%}) & MA-CNN (6.23\%) & CDRM (54.88\%) \\ \hline
ResNet-M1 & CDRM (\textcolor{red}{26.30\%}) & CCRE (51.53\%) & CCRE (65.45\%) \\ \hline
ResNet-M2 & CCRE (6.07\%) & CCRE (6.63\%) & CCRE (53.78\%) \\

\hline
\end{tabular}
\end{center}
\caption{Highest classification rates for the evaluated participant enhancement algorithms that improved the classification performance of the un-altered test images. Improvement was not achieved in all cases (deterioration denoted by red entries).}
\label{tab:testing_dataset_scores}
\vspace{-8mm}
\end{table}

\begin{figure*}[!ht]
\centering
\subfloat[\label{fig:graph:C2Baselines:UAV}]
{
    \pgfplotscreateplotcyclelist{my colors}{
        % color for the legend
        black!50\\
        % color for the "real" plots
        red\\
        cyan\\
        pink\\
        blue\\
        green\\
    }
    \pgfplotsset{
        compat=newest,
        cycle list name=my colors,
        legend cell align=left,
        width=0.33\textwidth,
    }

\begin{tikzpicture}
    \begin{axis}[
        title={UAV Collection},
        xlabel={M1 classification rate [\%]},
        ylabel={M2 classification rate [\%]},
        ymin=0.0,ymax=10,
        minor y tick num=2,
        ymajorgrids=true,
        xmajorgrids=true,
        yminorgrids=true,
        minor grid style=loosely dotted,
        only marks,
        scatter,
        mark size=2.5pt,
        scatter src=explicit symbolic,
        table/meta=Method,
        scatter/classes={
            VGG16={mark=diamond*},
            VGG19={mark=halfdiamond*},
            Inception={mark=square*},
            ResNet={mark=triangle*},
            MobileNet={mark=star},
            NASNetMobile={mark=otimes}
        },
        % this legend only shows the defined scatter classes
        % (as it is stated in the manual)
        % legend entries={
        %     VGG16,
        %     VGG19,
        %     Inception,
        %     ResNet%
        % },
        % %legend pos= north west,
        % %legend pos=outer north east,
        % legend style={  at={(0.5,-0.2)},
        %                 anchor=north,legend columns=-1}
    ]

        % ---------------------------------------------------------------------
        % dummy plot for the legend
        % (make sure the expression values are outside the visible axis limits
        %  if this `\addplot wouldn't be present. This requires at least
        %  setting one of the limits explicitly, i.e. in this case `xmin')
        \addplot table [
            x expr=-10,
            y expr=-10,
        ] {figures/C2_BUAV.dat};
        % ---------------------------------------------------------------------

        % simplified the call for the data
        \foreach \i in {
            Baseline,
            SRCNN,
            Bilinear,
            Honeywell,
            TexasAMPeking%
        }{
            \addplot table [
                x=\i-M1,
                y=\i-M2,
            ] {figures/C2_BUAV.dat};
        }

    \end{axis}

    % this is a dummy `axis' environment only to create the second legend
    % \begin{axis}[
    %     % set some axis limits and plot the coordinates outside that box
    %     % so they don't show up
    %     xmin=1,
    %     xmax=2,
    %     ymin=1,
    %     ymax=2,
    %     % of course we also don't want to show this axis
    %     hide axis,
    %     % we need only marks
    %     only marks,
    %     % state the legend entries for the second legend
    %     % (here we don't have scatter classes, so each `\addplot' gets its
    %     %  own entry in the legend)
    %     legend entries={
    %         ,       % the dummy plot should not show up in the legend
    %         Baseline,
    %         Honeywell,
    %         TexasAMPeking,
    %         Northwestern,
    %         TsingHua%
    %     },
    %     % place it below the other legend
    %     % therefore we have to shift it down (manually)
    %     %legend pos=outer north east,
    %     legend style={  at={(0.5,-0.2)},
    %                     anchor=north,
    %                     legend columns=3,
    %                     yshift=-20pt}
    %     %legend style={
    %      %   yshift=-60pt,
    %     %},
    % ]
    %     % just add some dummy plots to create the legend
    %     \foreach \i in {0,...,5} {
    %         \addplot+ [mark=*] coordinates { (0,0) };
    %     }
    % \end{axis}
\end{tikzpicture}
}
\subfloat[\label{fig:graph:C2Baselines:Glider}]
{
    \pgfplotscreateplotcyclelist{my colors}{
        % color for the legend
        black!50\\
        % color for the "real" plots
        red\\
        cyan\\
        pink\\
        blue\\
        green\\
    }
    \pgfplotsset{
        compat=newest,
        cycle list name=my colors,
        legend cell align=left,
        width=0.33\textwidth,
    }

\begin{tikzpicture}
    \begin{axis}[
        title={Glider Collection},
        xlabel={M1 classification rate [\%]},
        ylabel={M2 classification rate [\%]},
        ymin=0.0,ymax=10,
        minor y tick num=2,
        ymajorgrids=true,
        xmajorgrids=true,
        yminorgrids=true,
        minor grid style=loosely dotted,
        only marks,
        scatter,
        mark size=2.5pt,
        scatter src=explicit symbolic,
        table/meta=Method,
        scatter/classes={
            VGG16={mark=diamond*},
            VGG19={mark=halfdiamond*},
            Inception={mark=square*},
            ResNet={mark=triangle*},
            MobileNet={mark=star},
            NASNetMobile={mark=otimes}
        },
        % this legend only shows the defined scatter classes
        % (as it is stated in the manual)
        % legend entries={
        %     VGG16,
        %     VGG19,
        %     Inception,
        %     ResNet%
        % },
        % %legend pos= north west,
        % %legend pos=outer north east,
        % legend style={  at={(0.5,-0.2)},
        %                 anchor=north,legend columns=-1}
    ]

        % ---------------------------------------------------------------------
        % dummy plot for the legend
        % (make sure the expression values are outside the visible axis limits
        %  if this `\addplot wouldn't be present. This requires at least
        %  setting one of the limits explicitly, i.e. in this case `xmin')
        \addplot table [
            x expr=-10,
            y expr=-10,
        ] {figures/C2_BGlider.dat};
        % ---------------------------------------------------------------------

        % simplified the call for the data
        \foreach \i in {
            Baseline,
            SRCNN,
            Bilinear,
            Honeywell,
            TexasAMPeking%
        }{
            \addplot table [
                x=\i-M1,
                y=\i-M2,
            ] {figures/C2_BGlider.dat};
        }

    \end{axis}

    % this is a dummy `axis' environment only to create the second legend
    % \begin{axis}[
    %     % set some axis limits and plot the coordinates outside that box
    %     % so they don't show up
    %     xmin=1,
    %     xmax=2,
    %     ymin=1,
    %     ymax=2,
    %     % of course we also don't want to show this axis
    %     hide axis,
    %     % we need only marks
    %     only marks,
    %     % state the legend entries for the second legend
    %     % (here we don't have scatter classes, so each `\addplot' gets its
    %     %  own entry in the legend)
    %     legend entries={
    %         ,       % the dummy plot should not show up in the legend
    %         Baseline,
    %         Honeywell,
    %         TexasAMPeking,
    %         Northwestern,
    %         TsingHua%
    %     },
    %     % place it below the other legend
    %     % therefore we have to shift it down (manually)
    %     %legend pos=outer north east,
    %     legend style={  at={(0.5,-0.2)},
    %                     anchor=north,
    %                     legend columns=3,
    %                     yshift=-20pt}
    %     %legend style={
    %      %   yshift=-60pt,
    %     %},
    % ]
    %     % just add some dummy plots to create the legend
    %     \foreach \i in {0,...,5} {
    %         \addplot+ [mark=*] coordinates { (0,0) };
    %     }
    % \end{axis}
\end{tikzpicture}
}
\subfloat[\label{fig:graph:C2Baselines:Ground}]
{
    \pgfplotscreateplotcyclelist{my colors}{
        % color for the legend
        black!50\\
        % color for the "real" plots
        red\\
        cyan\\
        pink\\
        blue\\
        green\\
    }
    \pgfplotsset{
        compat=newest,
        cycle list name=my colors,
        legend cell align=left,
        width=0.33\textwidth,
    }

\begin{tikzpicture}
    \begin{axis}[
        title={Ground Collection},
        xlabel={M1 classification rate [\%]},
        ylabel={M2 classification rate [\%]},
        ymin=0,ymax=60,
        minor y tick num=2,
        ymajorgrids=true,
        xmajorgrids=true,
        yminorgrids=true,
        minor grid style=loosely dotted,
        only marks,
        scatter,
        mark size=2.5pt,
        scatter src=explicit symbolic,
        table/meta=Method,
        scatter/classes={
            VGG16={mark=diamond*},
            VGG19={mark=halfdiamond*},
            Inception={mark=square*},
            ResNet={mark=triangle*},
            MobileNet={mark=star},
            NASNetMobile={mark=otimes}
        },
        % this legend only shows the defined scatter classes
        % (as it is stated in the manual)
        % legend entries={
        %     VGG16,
        %     VGG19,
        %     Inception,
        %     ResNet%
        % },
        % %legend pos= north west,
        % %legend pos=outer north east,
        % legend style={  at={(0.5,-0.2)},
        %                 anchor=north,legend columns=-1}
    ]

        % ---------------------------------------------------------------------
        % dummy plot for the legend
        % (make sure the expression values are outside the visible axis limits
        %  if this `\addplot wouldn't be present. This requires at least
        %  setting one of the limits explicitly, i.e. in this case `xmin')
        \addplot table [
            x expr=-10,
            y expr=-10,
        ] {figures/C2_BGround.dat};
        % ---------------------------------------------------------------------

        % simplified the call for the data
        \foreach \i in {
            Baseline,
            SRCNN,
            Bilinear,
            Honeywell,
            TexasAMPeking%
        }{
            \addplot table [
                x=\i-M1,
                y=\i-M2,
            ] {figures/C2_BGround.dat};
        }

    \end{axis}

    % this is a dummy `axis' environment only to create the second legend
    % \begin{axis}[
    %     % set some axis limits and plot the coordinates outside that box
    %     % so they don't show up
    %     xmin=1,
    %     xmax=2,
    %     ymin=1,
    %     ymax=2,
    %     % of course we also don't want to show this axis
    %     hide axis,
    %     % we need only marks
    %     only marks,
    %     % state the legend entries for the second legend
    %     % (here we don't have scatter classes, so each `\addplot' gets its
    %     %  own entry in the legend)
    %     legend entries={
    %         ,       % the dummy plot should not show up in the legend
    %         Baseline,
    %         Honeywell,
    %         TexasAMPeking,
    %         Northwestern,
    %         TsingHua%
    %     },
    %     % place it below the other legend
    %     % therefore we have to shift it down (manually)
    %     %legend pos=outer north east,
    %     legend style={  at={(0.5,-0.2)},
    %                     anchor=north,
    %                     legend columns=3,
    %                     yshift=-20pt}
    %     %legend style={
    %      %   yshift=-60pt,
    %     %},
    % ]
    %     % just add some dummy plots to create the legend
    %     \foreach \i in {0,...,5} {
    %         \addplot+ [mark=*] coordinates { (0,0) };
    %     }
    % \end{axis}
\end{tikzpicture}
}
\vfill \vspace{0.5em}
\begin{tikzpicture}
    \begin{axis}[%
        hide axis,
        xmin=10,
        xmax=50,
        ymin=0,
        ymax=0.4,
        only marks,
        mark size=3.5pt,
        legend style={
            draw=white!15!black,
            legend cell align=left, 
            legend columns=6,
        }
    ]
    \addlegendimage{black!50,mark=diamond*}
    \addlegendentry{VGG16};
    \addlegendimage{black!50,mark=halfdiamond*}
    \addlegendentry{VGG19};
    \addlegendimage{black!50,mark=square*}
    \addlegendentry{Inception};
    \addlegendimage{black!50,mark=triangle*}
    \addlegendentry{ResNet};
    \addlegendimage{black!50,mark=star}
    \addlegendentry{MobileNet};
    \addlegendimage{black!50,mark=otimes}
    \addlegendentry{NASNetMobile};
    
    \addlegendimage{red,mark=*}
    \addlegendentry{Baseline};
    
    \addlegendimage{cyan,mark=*}
    \addlegendentry{SRCNN};
    
    \addlegendimage{pink,mark=*}
    \addlegendentry{Bilinear};
    
    \addlegendimage{blue,mark=*}
    \addlegendentry{CCRE};
    
    \addlegendimage{green,mark=*}
    \addlegendentry{CDRM};

    \end{axis}
\end{tikzpicture}
\caption{Performance comparison between the two best performing participant algorithms and two best baseline enhancement algorithms.}
\label{fig:graph:C2Baselines}
\end{figure*}
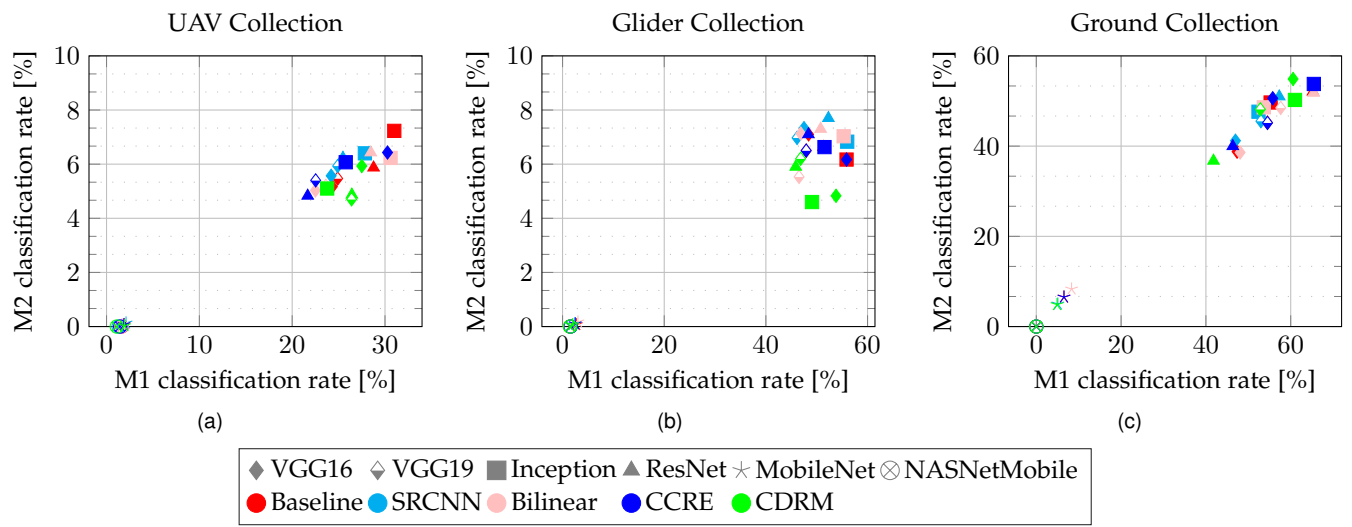